\documentclass[10pt,twocolumn,letterpaper]{article}

\usepackage[pagenumbers]{iccv} 
\usepackage[dvipsnames]{xcolor}

\newcommand{\name}{\textsc{SuperDec}}

\usepackage{multirow}
\usepackage{overpic}
\usepackage{amsmath}
\usepackage[export]{adjustbox}

\usepackage{caption} 
\definecolor{pastelyellow}{rgb}{1.0, 1.0, 0.6}
\definecolor{pastelgreen}{rgb}{0.7, 1.0, 0.7}
\definecolor{pastelblue}{rgb}{0.7, 0.85, 1.0}
\definecolor{pastelred}{rgb}{1.0, 0.7, 0.7}
\definecolor{pastelpurple}{rgb}{0.85, 0.7, 1.0}

\input{macros}

\definecolor{iccvblue}{rgb}{0.21,0.49,0.74}
\usepackage[pagebackref,breaklinks,colorlinks,allcolors=iccvblue]{hyperref}

\def\paperID{4571} 
\def\confName{ICCV}
\def\confYear{2025}

\title{\name{}: 3D Scene Decomposition with Superquadric Primitives}

\author{
\vspace{-25px}\\
{Elisabetta Fedele}$^{1,2}$ \hspace{10px}
{Boyang Sun}$^{1}$ \hspace{10px} 
{Leonidas Guibas}${^2}$  \hspace{10px}%
{Marc Pollefeys}${^{1,3}}$ \hspace{10px}
{Francis Engelmann}${^{2}}$
\vspace{10px}\\
{\small
$^1$ETH Zurich \hspace{15px}
$^2$Stanford University \hspace{15px}
$^3$Microsoft
}
}

\begin{document}
\maketitle
\begin{abstract}
We present \name{}, an approach for creating compact 3D scene representations via decomposition into superquadric primitives.
While most recent methods use geometric primitives to obtain photorealistic 3D reconstructions,
we instead leverage them to obtain a compact yet expressive representation.
To this end, we design a novel architecture that efficiently decomposes point clouds of arbitrary objects into a compact set of superquadrics.
We train our model on ShapeNet and demonstrate its generalization capabilities on object instances
from ScanNet++ as well as on full Replica scenes.
Finally, we show that our compact superquadric-based representation supports a wide range of downstream applications, including robotic manipulation and controllable visual content generation.
Project page: \url{https://super-dec.github.io}.
\end{abstract}    
\section{Introduction}
\label{sec:intro}

\begin{figure}[t]
    \centering
    \includegraphics[width=0.495\linewidth]{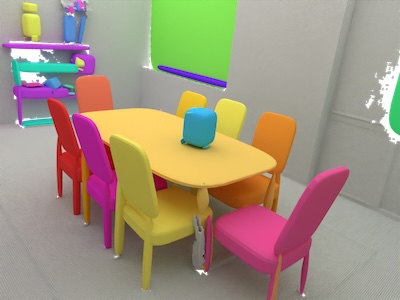}\hspace{1px}%
    \includegraphics[width=0.495\linewidth]{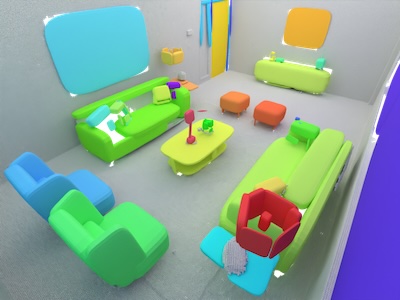}\\
    \includegraphics[width=0.495\linewidth,,clip]{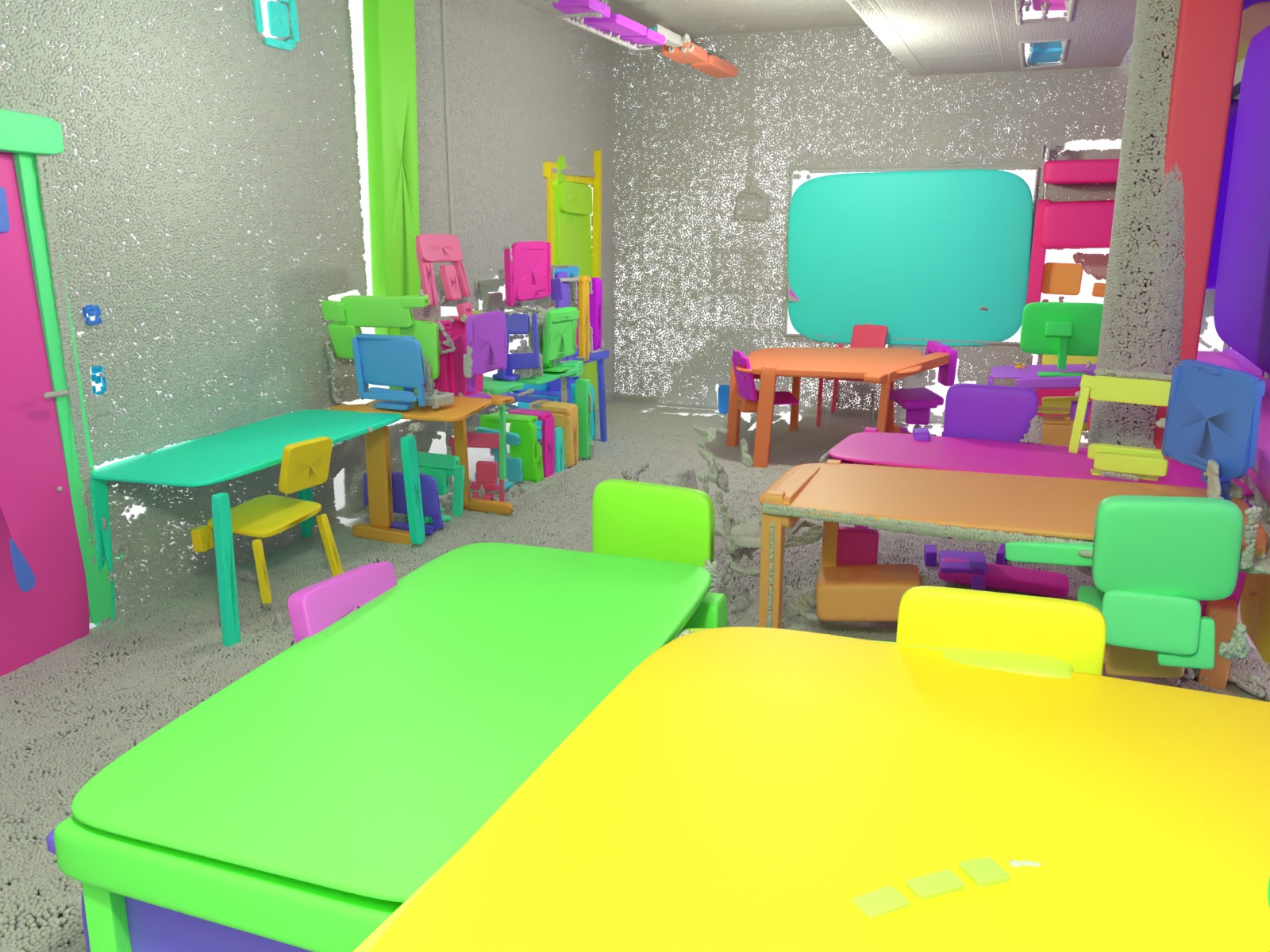}\hspace{1px}%
    \includegraphics[width=0.495\linewidth,,clip]{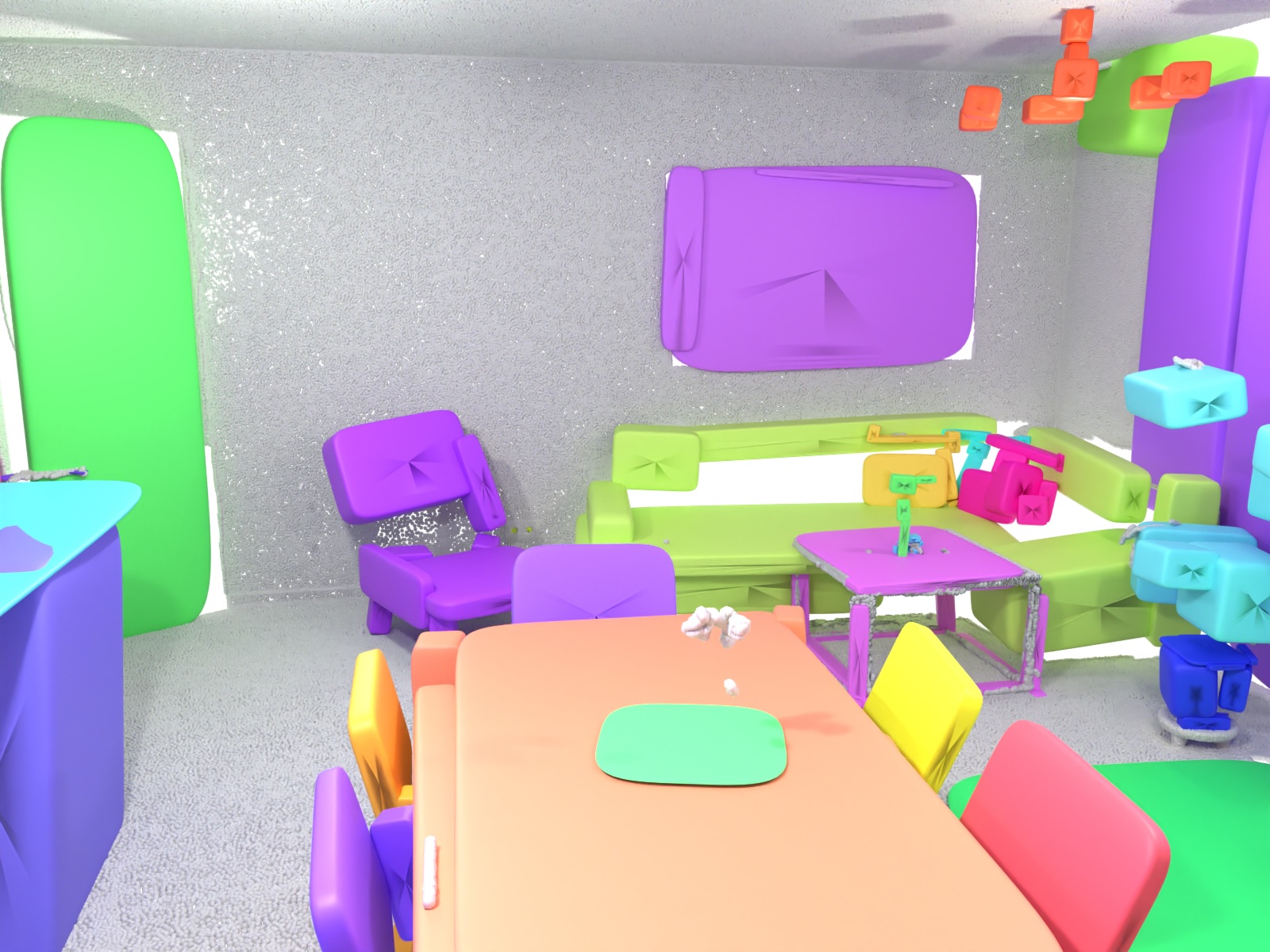}%
    \caption{\textbf{3D Scene Decomposition with Superquadrics.}
    Given a 3D point cloud of an indoor scene, \name{} decomposes all scene objects into a compact set of superquadric primitives. Different object instances are visualized with different colors.}
    \label{fig:teaser}
\end{figure}

3D scene representations play a central role in computer vision and robotics,
enabling tasks such as 3D scene understanding~\cite{openscene, takmaz2023openmask3d, takmaz2025search3d}, scene generation~\cite{shriram2024realmdreamer, schult2024controlroom3d, huang2025vipscene},
functional reasoning~\cite{zhang2025open, koch2024open3dsg},
and scene interaction~\cite{ray2024task, gu2024conceptgraphs, delitzas2024scenefun3d, li2024genzi}.
Recent work~\cite{kerbl3Dgaussians} utilizes 3D Gaussians as geometric primitives to produce high-quality, photorealistic reconstructions.
However, such representations are often memory-intensive.
In contrast, we propose a more lightweight yet geometrically faithful 3D scene representation by decomposing the input point cloud into a compact set of explicit primitives,
namely superquadrics (Fig.~\ref{fig:teaser}).

\noindent Representations for 3D scenes include well-established formats such as point clouds, meshes, signed distance functions, and voxel grids, each offering different trade-offs among geometric detail, computational cost, resolution, performance, interpretability, and editability.
Recently, multi-view approaches such as Neural Radiance Fields (NeRF)~\cite{mildenhall2020nerf} and Gaussian Splatting (GS)~\cite{kerbl3Dgaussians} have gained popularity as 3D scene representations.
These methods optimize photometric losses to ensure that their underlying representations (implicit in the case of NeRF, and explicit in the case of GS) are consistent with observed images.
While these approaches excel at achieving photorealism, they lack explicit control over compactness, often resulting in large, non-modular scene encodings that are unsuitable for tasks requiring explicit spatial reasoning.

\noindent While optimizing compactness on a scene level remains a challenging task, prior work has shown that geometric primitives such as cuboids~\cite{Tulsiani2016LearningSA, Yang2021UnsupervisedLF} or superquadrics~\cite{Paschalidou2019CVPR, alaniz2023iterative, Monnier2023DifferentiableBW} enable compact and interpretable decompositions of \textit{individual objects}.
These methods are generally either learning-based ~\cite{Paschalidou2019CVPR, Yang2021UnsupervisedLF}, favoring speed at the cost of geometric accuracy, or optimization-based~\cite{liu2022robust, alaniz2023iterative, Monnier2023DifferentiableBW}, offering improved accuracy  at the expense of higher computational cost.
Although both types of approaches can be effective for certain object classes, they often struggle to generalize across datasets with diverse object geometries. Learning-based methods typically require category-specific training, while optimization-based ones rely on hand-crafted heuristics, limiting their scalability in open-world settings.

\noindent Motivated by the abstraction capabilities of geometric primitives for individual object categories, we propose to represent complex 3D scenes using a compact set of superquadrics.
To this end, we learn \emph{class-agnostic} object-level shape priors to optimize compactness and leverage an off-the-shelf 3D instance segmentation method (Mask3D~\cite{Schult2022Mask3DMT}) to scale our approach to full 3D scenes.
 
\noindent We choose superquadrics as building blocks for our representation, as they offer more accurate shape modeling than cuboids while incurring minimal additional parameter overhead (9\,\emph{v.s.}\,11, including 6-DoF pose parameters).
To develop a model capable of generalizing across diverse object types, we draw inspiration from supervised segmentation~\cite{Schult2022Mask3DMT, Cheng2021MaskedattentionMT, Cheng2021PerPixelCI}, and approach the problem from the perspective of unsupervised geometric-based segmentation, using local point-based features to iteratively refine predicted geometric primitives.
Our model is trained on ShapeNet~\cite{shapenet2015} and evaluated on three challenging and diverse 3D datasets: ShapeNet~\cite{shapenet2015}, ScanNet++~\cite{Yeshwanth2023ScanNetAH}, and Replica~\cite{Straub2019TheRD}.
On the 3D object dataset ShapeNet, our approach achieves an L2 error six times smaller than prior state-of-the-art methods~\cite{Paschalidou2019CVPR}, while requiring only half the number of primitives.
On ScanNet++ and Replica, we demonstrate that our method generalizes well to real-world scene-level settings, despite being trained solely on ShapeNet.
Finally, we show the practical utility of our method as a scene representation for robotic tasks, including path planning and object grasping, as well as an editable 3D scene representation for controllable image generation.

\noindent In summary, our contributions are the following:\\
{1)} We introduce \name{}, a novel method for decomposing 3D scenes using superquadric primitives.\\
{2)} \name{} achieves state-of-the-art object decomposition scores on ShapeNet trained jointly on multiple classes.\\
{3)} We demonstrate the effectiveness of 3D superquadric scene representations for robotic tasks and controllable generative content creation.
\section{Related Work}
\label{sec:related-works}
\vspace{-5px}
\paragraph{Learning-based methods}
\label{sec:obj-level-learning}
have shown that neural networks, when equipped with suitable reconstruction losses, can directly predict geometric primitive parameters to decompose point clouds into a minimal set of primitives for specific object categories.
Tulsiani~\cite{Tulsiani2016LearningSA} introduced a CNN-based method for cuboid decomposition, which was later extended to more expressive primitives such as superquadrics by Paschalidou \etal~\cite{Paschalidou2019CVPR}.
CSA~\cite{Yang2021UnsupervisedLF} further enhanced interpretability by employing a stronger point encoder and jointly predicting cuboid parameters and part segmentations.
However, these methods remain constrained by their reliance on category-specific training.
We attribute this limitation to their model design, which encodes only global shape features, sufficient for intra-category generalization but ineffective for decomposing out-of-category objects.
 
\vspace{-10px}
\paragraph{Optimization-based methods}
\label{sec:obj-level-optimization}
largely originate from the literature on superquadric fitting.
EMS~\cite{liu2022robust} revisited this line of work by introducing a probabilistic formulation that enables the decomposition of arbitrary objects into multiple superquadrics.
Given an input point cloud, the method first fits a superquadric to the main structure and identifies unfitted outlier clusters, which are then recursively processed in a hierarchical fashion up to a predefined depth level.
However, as noted in their paper and confirmed by our experiments, this approach implicitly assumes that objects exhibit a hierarchical geometric structure, limiting its applicability to many real-world objects such as tables and chairs.
Other methods, such as Marching Primitives~\cite{liu2023marching}, require Signed Distance Functions (SDFs) as input, which are generally not easily available in real-world scenes.
More fundamentally, since these approaches optimize from scratch for each object, they cannot leverage generalizable point features or learned shape priors, both of which are critical for abstraction and robustness under partial observations, a common challenge in practical 3D capture scenarios.

\vspace{-10px}
\paragraph{Scene-level decomposition.}
\label{sec:scene-level}
With the emergence of 3DGS~\cite{kerbl3Dgaussians}, an increasing number of works have explored representing 3D scenes using various geometric primitives as generalized ellipsoids~\cite{Held20243DConvex} and convexes~\cite{hamdi2024ges}.
While heuristics can control the number of Gaussians, achieving truly compact representations remains challenging. 
DBW~\cite{Monnier2023DifferentiableBW} addresses this by fitting a small set of textured superquadrics to 3D scenes,
building on the principles of 3DGS.
Given a set of scene images, it performs test-time optimization with a photometric loss and renders the primitives using a differentiable rasterizer~\cite{Liu2019SoftRA}.
To model the environment, DBW adds a meshed ground plane and a meshed icosphere for the background.
However, it is restricted to scenes with fewer than 10 primitives and requires objects to be aligned to a ground plane.
Furthermore, the optimization is computationally expensive, taking around three hours even on simple DTU~\cite{Jensen2014LargeSM} scenes.
Our method differs significantly in terms of input requirements, generality of application, and computational efficiency. 

\vspace{-10px}
\paragraph{Superquadrics} are a parametric family of shapes introduced by Barr \etal~\cite{Barr1981SuperquadricsAA} in 1981 and have since been widely adopted in both computer vision and graphics~\cite{pentland1986parts, Chevalier2003SegmentationAS, Solina1990RecoveryOP}.
Their popularity stems from their ability to represent a diverse range of shapes with a highly compact parameterization.
A superquadric in its canonical pose is defined by just five parameters:
$(\scalex, \scaley, \scalez)$ for the scales along the three principal semi-axes and $(\shape_1, \shape_2)$ for the shape-defining exponents. Given those parameters, their surface is described by the implicit equation:
\begin{equation}
    f(\vx)= \left(  \left(\frac{x}{\scalex}\right)^{\frac{2}{\epsilon_2}} +\left(\frac{y}{\scaley}\right)^{\frac{2}{\epsilon_2}}\right)^\frac{\epsilon_2}{\epsilon_1}  +\left(\frac{z}{\scalez}\right)^{\frac{2}{\epsilon_1}}=1\ .\label{eq:implicit}
\end{equation}
Extending this representation to a global coordinate system requires 6 additional parameters (3 for translation and 3 for rotation), resulting in a total of 11 parameters per superquadric.
Another key property of superquadrics is the ability to compute the \emph{radial distance} from any point in 3D space to the superquadric surface,
\ie{}, the distance between a point and the superquadric's surface along the line connecting that point to the center of the superquadric.
Specifically, given a point $\vx \in \R^3$, its radial distance to the surface of a canonically oriented superquadric is defined as:
\begin{equation}
\label{eq:rad-dist}
    \drad=|\vx|\cdot|1-f(\vx)^{-\epsilon_1/2}|\ ,
\end{equation}
where $f(\vx)$ is given in Eq.~\ref{eq:implicit}.
We refer the reader to \cite{Dop1998FittingUS} for the derivation of Eq.~\ref{eq:rad-dist} and to~\cite{jaklic2000segmentation} for a more comprehensive overview on superquadrics.
\begin{figure*}
    \centering
    \includegraphics[width=1.0\textwidth]{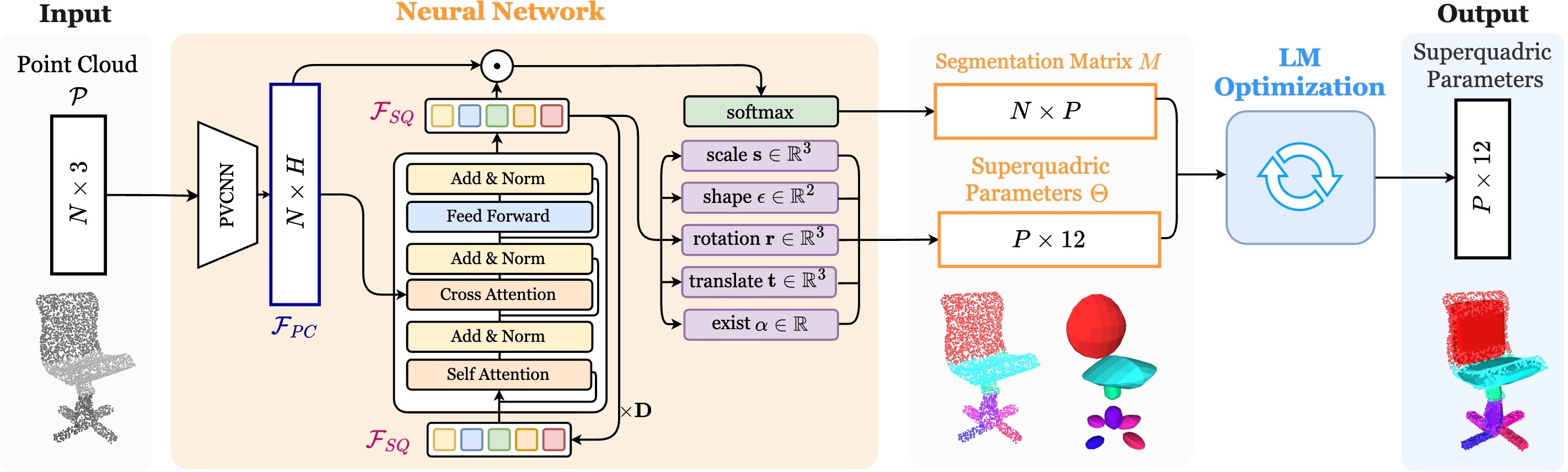}
    \caption{\textbf{Illustration of the \name{} Model.} Given a point cloud of an object with $N$ points, a Transformer-based neural network predicts parameters for $P$ superquadrics, as well as a soft segmentation matrix that assigns points to superquadrics. The predicted parameters include the 11 superquadric parameters and an objectness score. These predictions provide an effective initialization for the subsequent Levenberg–Marquardt (LM) optimization, which refines the superquadrics. 
    }
    \label{fig:model}
\end{figure*}

\section{Method}
Our ultimate goal is a 3D scene decomposition using superquadric primitives.
To this end, we first focus on single-object decomposition and then show how our method, combined with 3D instance segmentation~\cite{Schult2022Mask3DMT}, can be applied to full 3D scenes.
We detail the single-object approach in Sec.~\ref{sec:object-level-method} and its extension to full scenes in Sec.~\ref{sec:method-3d-scenes}.

\subsection{Single Object Decomposition}
\label{sec:object-level-method}

Fig.~\ref{fig:model} illustrates our model for single-object decomposition.
It consists of two main components:
a self-supervised feed-forward neural network that jointly predicts superquadric parameters and a segmentation matrix associating points to superquadrics,
followed by a lightweight Levenberg–Marquardt (LM) {optimization}~\cite{Levenberg1944AMF, Marquardt1963AnAF}.

\subsubsection{Feed-forward Neural Network}
\label{sec:nn}

Our deep learning model draws inspiration from recent fully-supervised Transformer-based~\cite{Vaswani2017AttentionIA} segmentation models~\cite{Schult2022Mask3DMT, Cheng2021MaskedattentionMT, Cheng2021PerPixelCI}.
These models iteratively decode a sequence of queries, each representing a segmentation mask, by cross-attending to input pixels or points.
In our case, the queries represent superquadrics.
Next, we show how such an architecture can be adapted to \textit{unsupervisedly} segment superquadrics, instead of \textit{supervisedly} segment objects.

\vspace{-10px}
\paragraph{Model Details.}
Given an input  point cloud $\pc\in\R^{\NumPoints\times 3}$, where each of the $\NumPoints$ points has a 3D coordinate, we first extract rich point features $\PointFeatures\in\R^{\NumPoints\times\HiddenDim}$ using the PVCNN~\cite{liu2019pvcnn} point encoder.
At the same time, we initialize $P$ superquadrics features $\SQFeatures\in\R^{\NumPrim\times\HiddenDim}$ with sinusoidal positional encodings. We feed these features in a Transformer decoder~\cite{Vaswani2017AttentionIA} which leverages self-attention, cross-attention to the point features, and feed-forward layers to refine them.

\noindent Once refined, the superquadric features $\SQFeatures$ and the point features $\PointFeatures$
are fed into two prediction heads:
The \textit{segmentation head} takes as input $\SQFeatures$ and $\PointFeatures$ and predicts a soft assignment matrix $\AssignMatrix\in\R^{N\times P}$ associating points to superquadrics and whose elements are defined as:
\begin{equation}
    \AssignMatrixEl_{ij} =\sigma \bigl(\phi(\PointFeatures) \cdot \SQFeatures\bigr) \ ,
\end{equation}
where $\phi(\PointFeatures)\in\R^{\NumPoints\times\HiddenDim}$ is a learned projection of the point features to match the dimensionality of the superquadric features, and $\sigma$ is the softmax function.
The second head, the \textit{superquadric head}, takes the superquadric features $\SQFeatures$ as input and predicts 12 parameters for each superquadric: 11 encoding its 5-DoF shape and 6-DoF pose, and one modeling its existence probability $\exist$, enabling a variable number of superquadrics per object.

\vspace{-10px}
\paragraph{Losses.}
We train our model in a self-supervised manner, without requiring any ground truth annotation. Specifically, the total loss is defined as:
\begin{equation}\label{eq:loss}
   \Loss = \ReconstructionLoss + \wpar\ParsimonyLoss \ + \wexist\ExistLoss \ ,
\end{equation}
where $\ReconstructionLoss$ is the reconstruction loss aligning the predicted superquadrics to the input point cloud $\pc$, $\ParsimonyLoss$ is the parsimony loss encouraging a small number of primitives, $\ExistLoss$ is the existence loss, and $\wpar$, $\wexist$ are weighting coefficients.
The reconstruction loss $\ReconstructionLoss$ consists of three terms:
\begin{align}
    \ReconstructionLoss = \PCSQLoss + \SQPCLoss + \NorLoss \ .\label{eq:rec-loss}
\end{align}
The first two terms correspond to the bi-directional Chamfer distance between the input point cloud and the superquadric surfaces, while the third term serves as a regularizer incorporating normal information to improve convergence during training. To compute the Chamfer distance, we approximate each superquadric surface by uniformly sampling $\NumSampledPoints$ points, following the method of Pilu \etal~\cite{pilu1995equal}.
Denoting by $d(\vx_i,\vx_{js}')$ the euclidean distance between the $i$-th point in the input point cloud and the $s$-th point sampled on the surface of the $j$-th superquadric, we define $\PCSQLoss$ as:
\begin{equation}
    \PCSQLoss =\frac{1}{\NumPoints}\sum_{i=1}^\NumPoints\sum_{j=1}^\NumPrim \AssignMatrixEl_{ij} \min_{s\in[S]} d(\vx_i, \vx'_{js})\ , 
\end{equation} 
and 
$\SQPCLoss$ as:
\begin{equation}
    \SQPCLoss = \frac{1}{\NumSampledPoints\sum_{j=1}^\NumPrim \exist_j} \sum_{j=1}^\NumPrim \exist_j\sum_{s=1}^\NumSampledPoints \min_{i\in[N]}  d(\vx_i, \vx'_{js})\ .
\end{equation}
The last term of Eq.~\ref{eq:rec-loss}, \ie{}, $\NorLoss$ is defined as the reconstruction loss from Yang \etal~\cite{Yang2021UnsupervisedLF},
and is used to incorporate normal information during training which leads to accelerated convergence. Additionally, since we seek not only accuracy but also compactness, we introduce a parsimony loss to encourage the use of fewer primitives. To do that, we optimize the $0.5$-norm of $m_{j}\coloneqq\sum_{i=1}^\NumPoints\frac{m_{ij}}{\NumPoints}$ and define the parsimony loss as:
\begin{equation}
    \ParsimonyLoss = \left(\frac{1}{\NumPrim}\sum_{j=1}^\NumPrim\frac{\sqrt{m_{j}}}{\NumPrim}\right)^2 \ .
\end{equation}
Lastly, we employ an existence loss $\ExistLoss$ which uses the predicted segmentation as a teacher for the linear head in charge of predicting the existence probability. More specifically, given a threshold $\epsilon_{exist}$, we define the ground-truth existence of the $j$th superquadric as $\hat{\exist}_j\defeq m_j>\epsilon_{exist}$ and define $\ExistLoss$ as:
\begin{equation}
    \ExistLoss = \sum_{j=1}^\NumPrim \frac{BCE(\exist_j,\hat{\exist}_j)}{\NumPrim} \ ,
\end{equation}
where BCE is the binary cross entropy and $\exist_j$ is the predicted existence probability for the $j$th superquadric.
\subsubsection{Optimization}
\label{sec:optimization-module}
Our optimization module takes as input the predicted soft segmentation matrix $M$ as well as the superquadric parameters $\Theta$, and further refines the superquadric parameters using the Levenberg-Marquardt (LM)~\cite{Levenberg1944AMF, Marquardt1963AnAF} algorithm.
Specifically, given a point cloud of $N$ points, it iteratively refines the parameters $\Theta_j$ of the $j$th superquadric, by computing two sets of residuals:
The first set of residuals $r_{ij}$ with $i\in[1,N]$ and $j\in[1,P]$ is defined as:
\begin{equation}
    r_{ij}=\AssignMatrixEl_{ij}\tilde{d}_j(\vx_i)\ ,
\end{equation}
where $\tilde{d}_j(\vx_i)$ denotes the radial distance of point $\vx_i$ from the $j$th superquadric, computed according to Eq.~\ref{eq:rad-dist}.
The second set of residuals is used for normalization and is obtained by sampling a set of $K$ points $\mathbf{p_1},\dots,\mathbf{p_K}$ on the surface of the given superquadric and then computing the distance of each of them from the point cloud.
Specifically, for $i\in[N+1,N+K]$ and $j\in[1,P]$ we compute $r_{ij}$ as: 
\begin{equation}
    r_{ij}=\min_k||\mathbf{p_{i-N}}-\Pi_j(\mathbf{x_k})||_2, \quad \text{with }k\in[1,N] \ .
\end{equation}
\begin{table*}[ht]
\centering
\small
\setlength{\tabcolsep}{7pt}
\begin{tabular}{llc|ccc|ccc}
\toprule
& & & \multicolumn{3}{c}{\textit{In-category}} & \multicolumn{3}{c}{\textit{Out-of-category}}  \\
\textbf{Model} & \textbf{Primitive Type}& \textbf{Segmentation}&
\textbf{L1} $\downarrow$& \textbf{L2} $\downarrow$&  \textbf{\#\,Prim.}$\downarrow$ &
\textbf{L1} $\downarrow$& \textbf{L2} $\downarrow$&  \textbf{\#\,Prim.}$\downarrow$ \\
\midrule
EMS (Liu \etal)~\cite{liu2022robust} & Superquadrics&  \xmark & $5.771$& $1.345$ & $5.68$  & $5.410$& $1.211$ & $5.68$\\
CSA (Yang \etal)~\cite{Yang2021UnsupervisedLF} & Cuboids&  \cmark & $5.157$& $0.527$ & $9.21$ &$4.897$ & $0.427$ & $11.75$\\
SQ (Paschalidou \etal)~\cite{Paschalidou2019CVPR} & Superquadrics&  \xmark & $3.668$& $0.279$ & $10$ & $4.193$& $0.354$ & $9$ \\
\name{} (Ours) & Superquadrics & \cmark & $\mathbf{1.698}$& $\mathbf{0.047}$ & $\mathbf{5.8}$ & $\mathbf{1.847}$& $\mathbf{0.061}$ & $\mathbf{5.26}$\\ 
\bottomrule
\end{tabular}
\caption{\textbf{Quantitative Results on ShapeNet~\cite{shapenet2015}.}
We show scores for in-category and out-of-category experiments and are scaled by $10^2$.} 
\label{tab:eval-shapenet}
\end{table*}

\vspace{-10px}
\subsection{Decomposition of Full 3D Scenes}
\label{sec:method-3d-scenes}
After training on single objects, extending \name{} to full 3D scenes is straightforward.
Given a scene-level point cloud, we extract 3D object instance masks using Mask3D~\cite{Schult2022Mask3DMT}.
Each object is centered and uniformly rescaled to a sphere of radius $0.5$.
We then predict the superquadric primitives for each object individually using our model.
We found our model trained on ShapeNet~\cite{shapenet2015} to generalize well on real-world 3D scenes from ScanNet++~\cite{Yeshwanth2023ScanNetAH} and Replica~\cite{Straub2019TheRD} without additional fine-tuning.
\section{Experiments}
We first compare our \name{} with previous state-of-the-art methods on individual objects and full 3D scenes (Sec.~\ref{sec:eval}).
We then demonstrate the usefulness of our representation on down-stream applications for robotics and controllable image generation (Sec.~\ref{sec:applications}).
Finally, in Sec.~\ref{sec:ablation}, we present additional analyses on part segmentation and the implicit learning of shape categories, followed by a study of the compactness–accuracy trade-off and runtime.

\begin{figure*}
\centering
\begin{small}
\vspace{-12px}
\begin{tabular}{cccc}
\hspace{5mm} & \hspace{10cm} & \hspace{1mm} & \hspace{5cm} \\
 & \emph{In-category} && \textit{Out-of-category} \\
\cmidrule(lr){2-2}\cmidrule(lr){4-4} 
\end{tabular}
\\
\vspace{-3px}
\rotatebox{90}{\hspace{10px}Point Cloud}
\hspace{-4px}
\includegraphics[width=0.16\linewidth,trim={50px 70px 50px 70px},clip]
{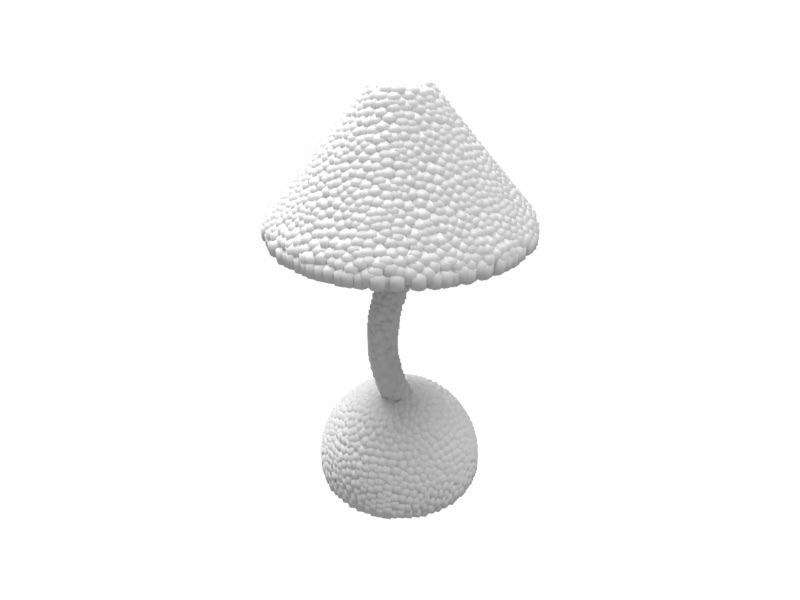}\hspace{-9px}
\includegraphics[width=0.16\linewidth,trim={50px 70px 50px 70px},clip]{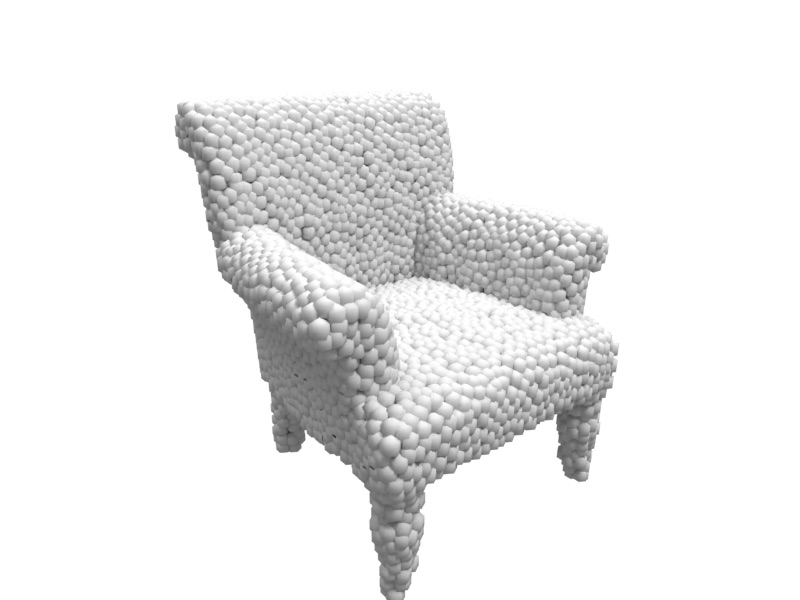}
\includegraphics[width=0.16\linewidth,trim={50px 70px 50px 70px},clip]{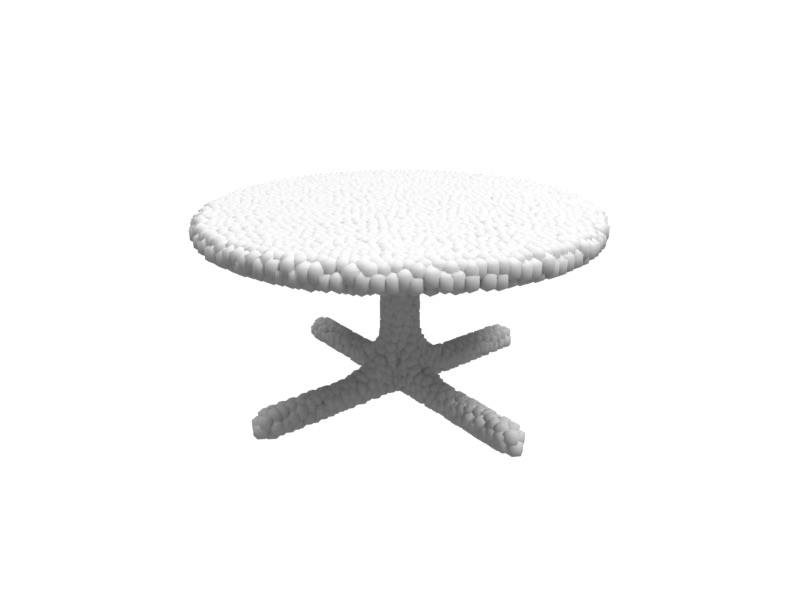}
\includegraphics[width=0.16\linewidth,trim={50px 70px 50px 70px},clip]{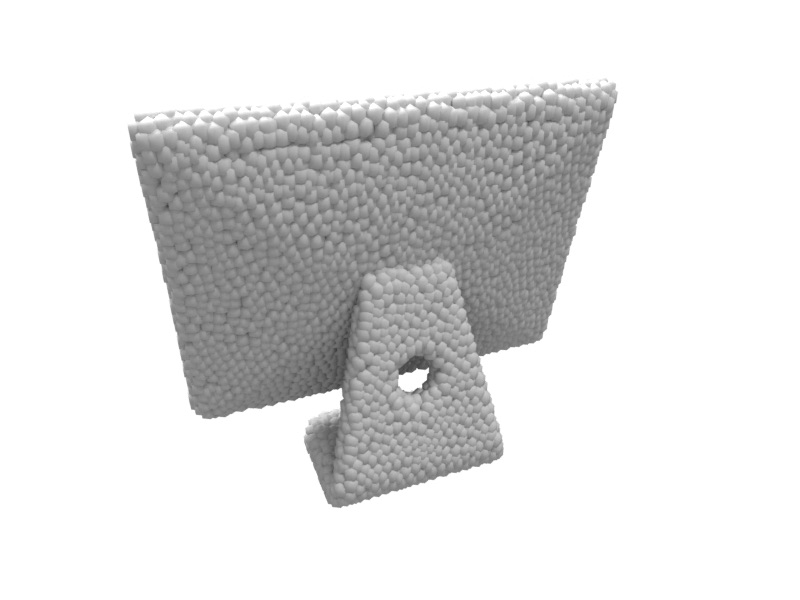}
\includegraphics[width=0.16\linewidth,trim={100px 70px 100px 100px},clip]{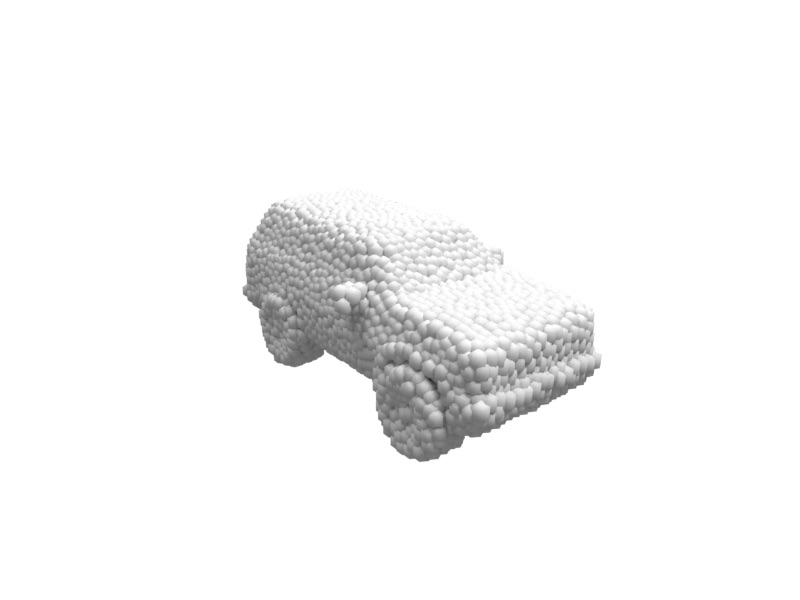}
\includegraphics[width=0.16\linewidth,trim={100px 70px 100px 100px},clip]{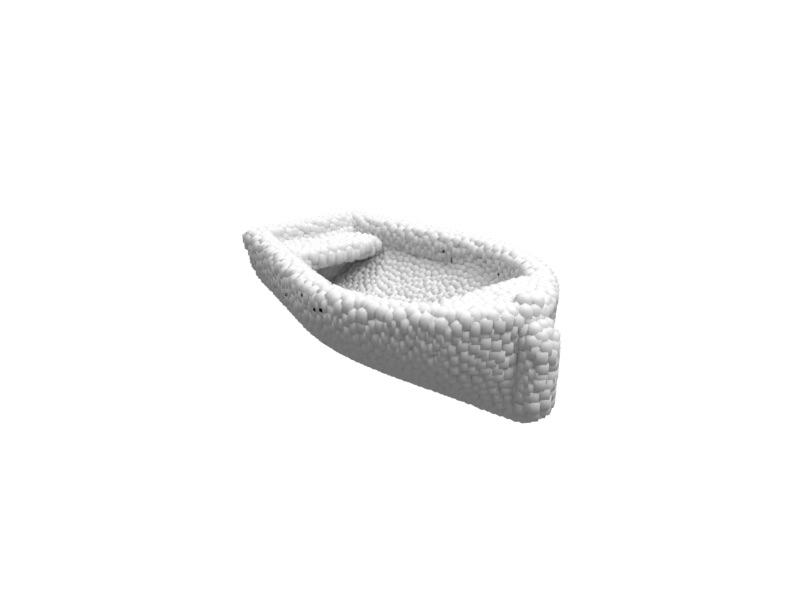}\\
\vspace{-1px}
\rotatebox{90}{\hspace{15px}EMS~\cite{liu2022robust}}
\hspace{-5px}
\includegraphics[width=0.16\linewidth,trim={50px 70px 50px 70px},clip]{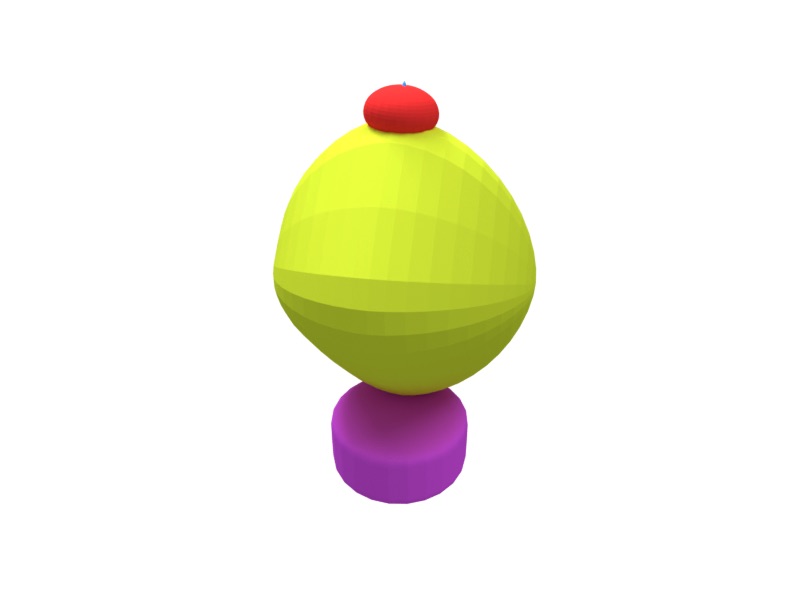}\hspace{-9px}
\includegraphics[width=0.16\linewidth,trim={50px 70px 50px 70px},clip]{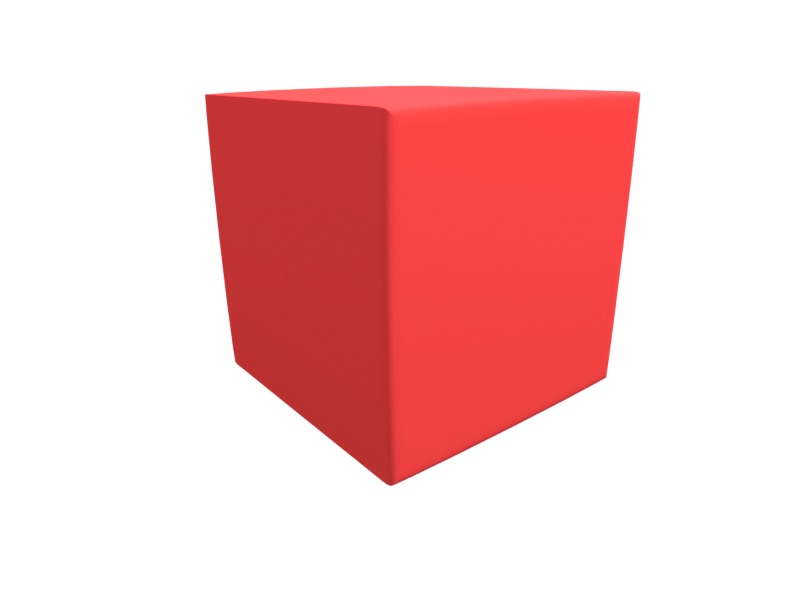}
\includegraphics[width=0.16\linewidth,trim={50px 70px 50px 70px},clip]{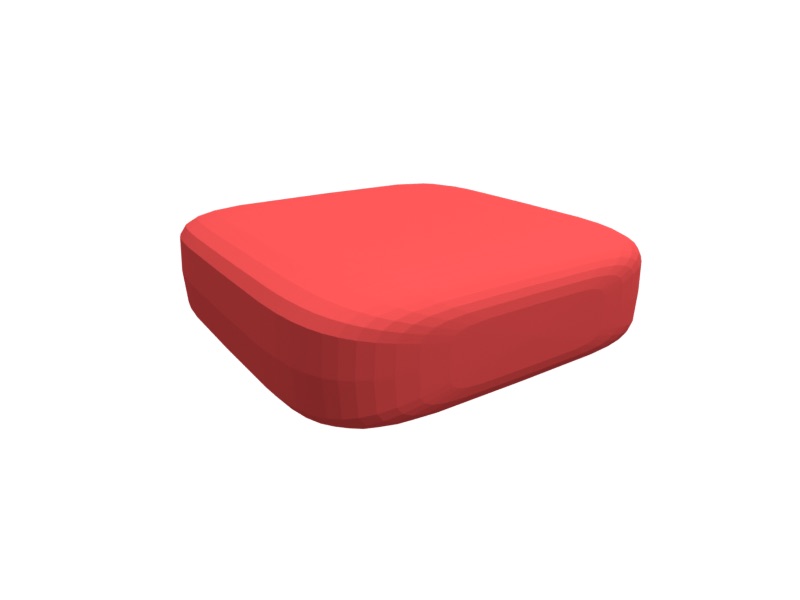}
\includegraphics[width=0.16\linewidth,trim={50px 70px 50px 70px},clip]{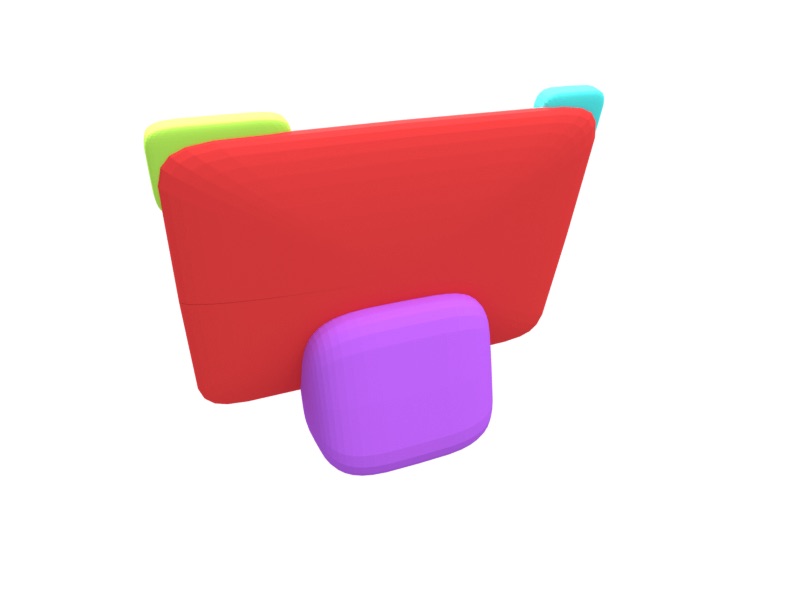}
\includegraphics[width=0.16\linewidth,trim={100px 100px 100px 100px},clip]{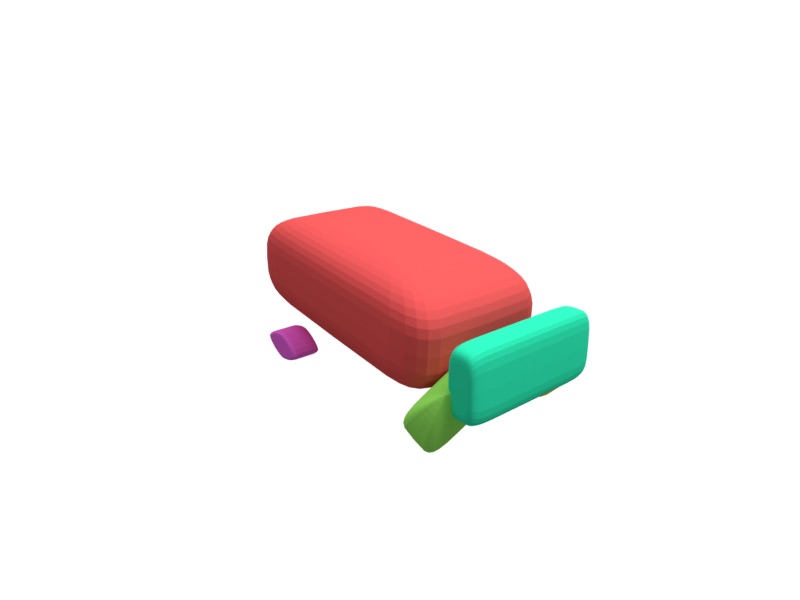}
\includegraphics[width=0.16\linewidth,trim={100px 100px 100px 100px},clip]{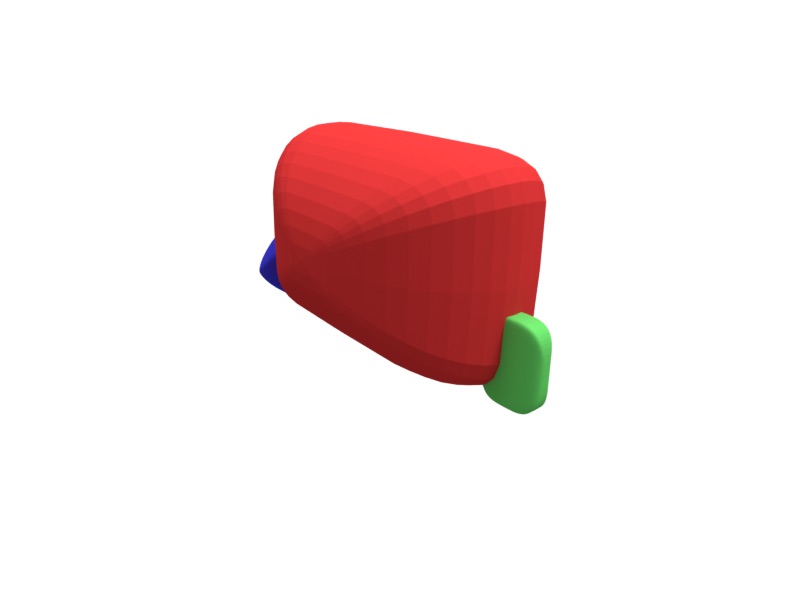}\\
\vspace{-5px}
\rotatebox{90}{\hspace{15px}CSA~\cite{Yang2021UnsupervisedLF}}
\hspace{-5px}
\includegraphics[width=0.16\linewidth,trim={50px 70px 50px 70px},clip]{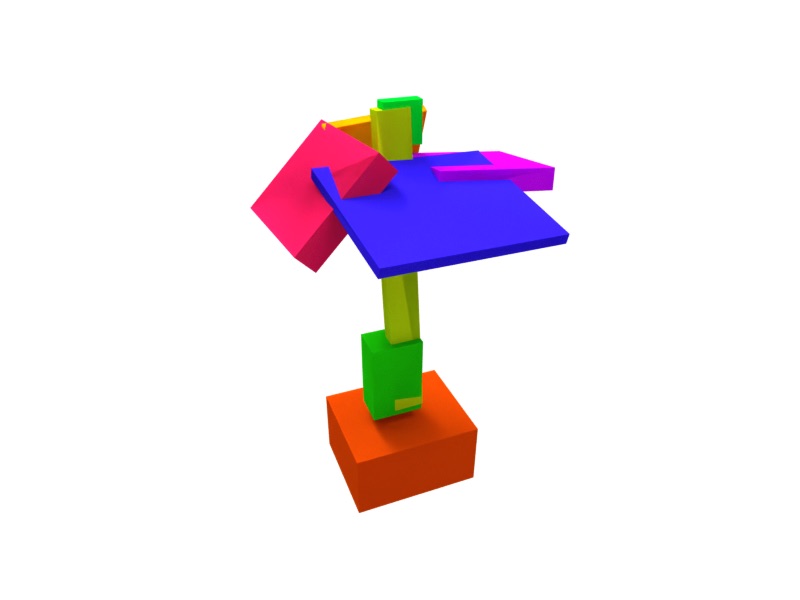}\hspace{-9px}
\includegraphics[width=0.16\linewidth,trim={50px 70px 50px 70px},clip]{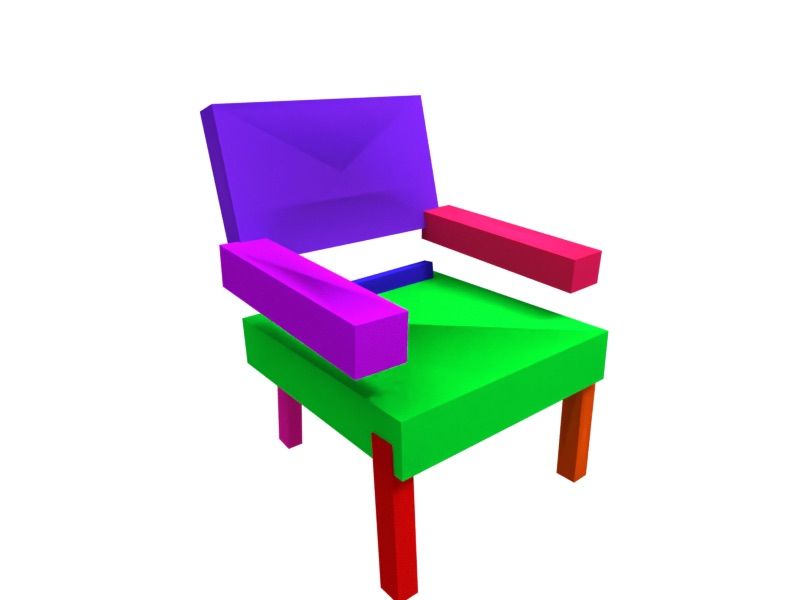}
\includegraphics[width=0.16\linewidth,trim={50px 70px 50px 70px},clip]{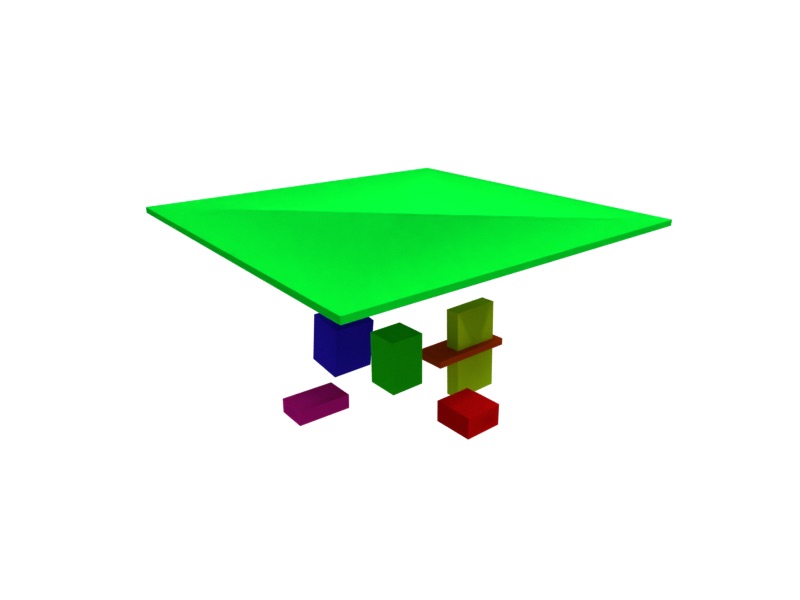}
\includegraphics[width=0.16\linewidth,trim={50px 70px 50px 70px},clip]{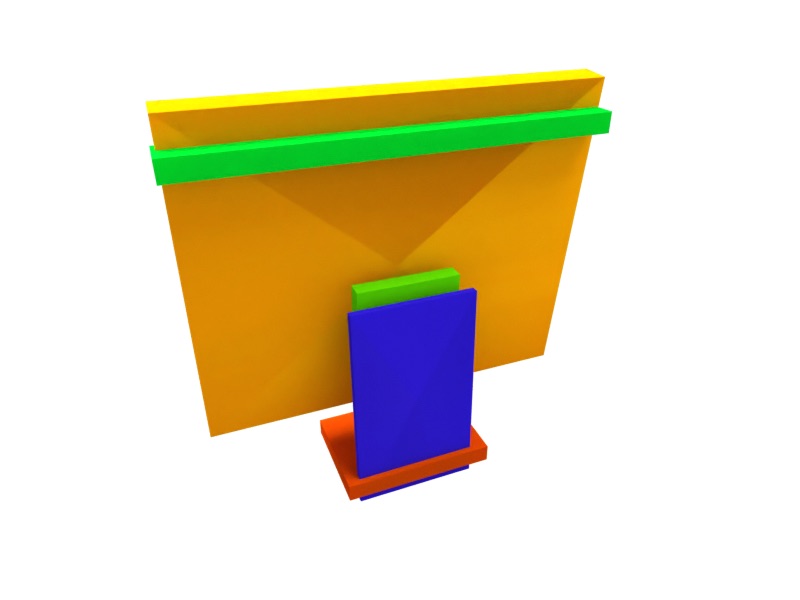}
\includegraphics[width=0.16\linewidth,trim={100px 100px 100px 100px},clip]{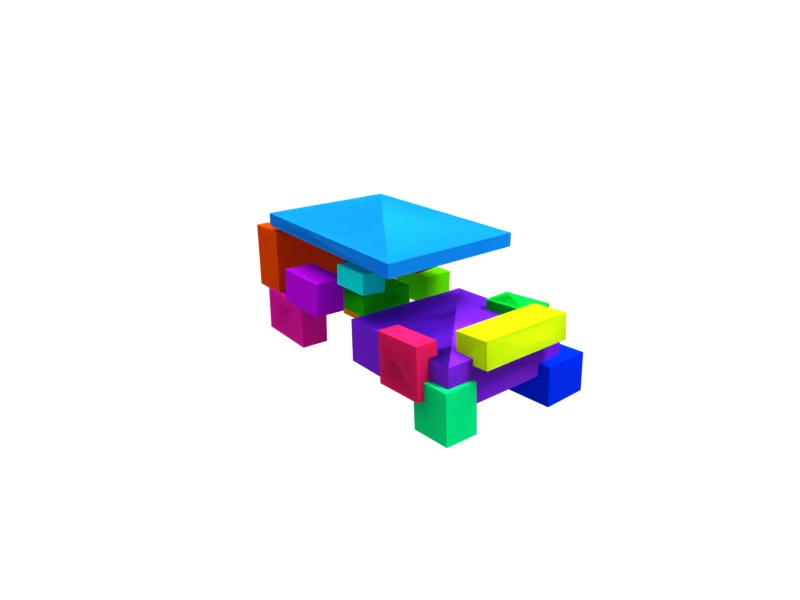}
\includegraphics[width=0.16\linewidth,trim={100px 100px 100px 100px},clip]{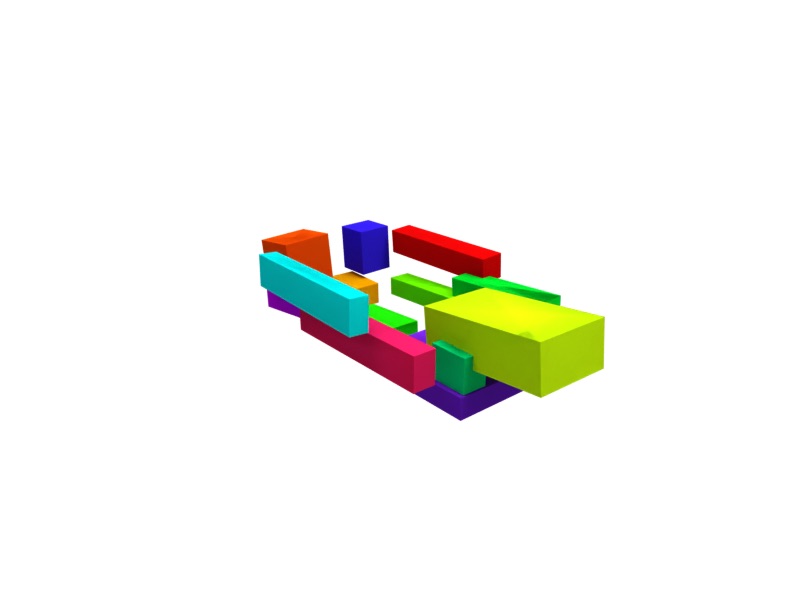}\\
\vspace{-0px}
\rotatebox{90}{\hspace{20px}SQ \cite{Paschalidou2019CVPR}}
\hspace{-5px}
\includegraphics[width=0.16\linewidth,trim={50px 70px 50px 70px},clip]{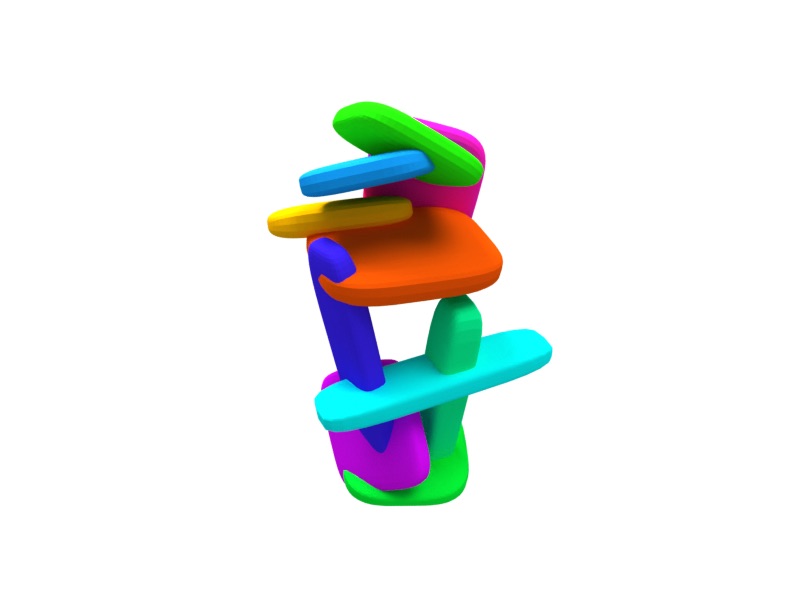}\hspace{-9px}
\includegraphics[width=0.16\linewidth,trim={50px 70px 50px 70px},clip]{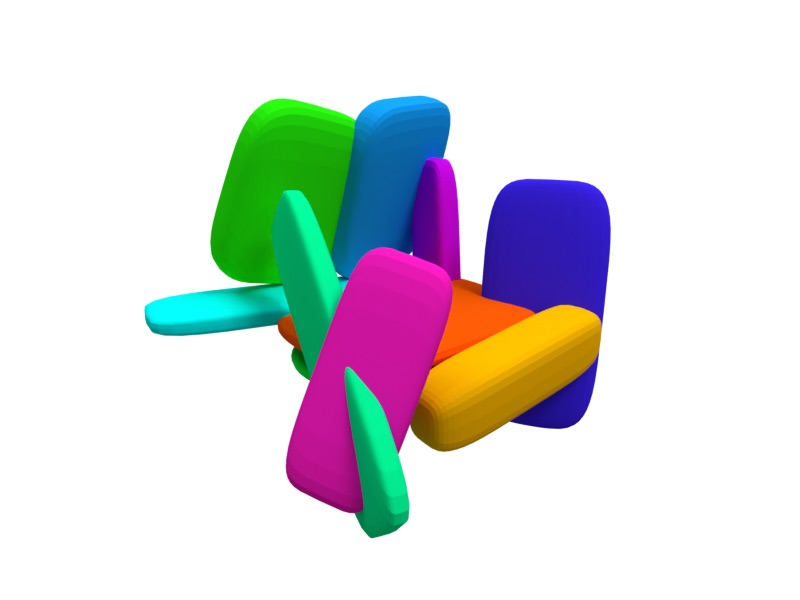}
\includegraphics[width=0.16\linewidth,trim={50px 70px 50px 70px},clip]{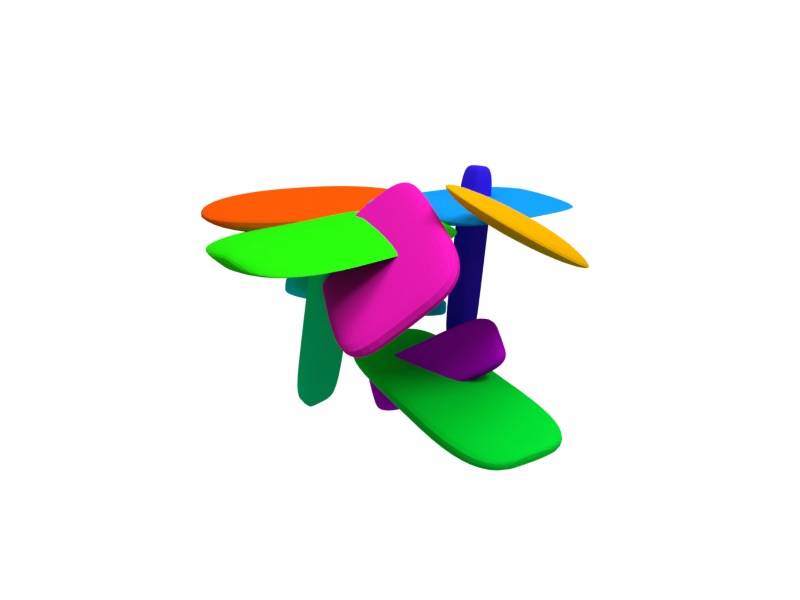}
\includegraphics[width=0.16\linewidth,trim={50px 70px 50px 70px},clip]{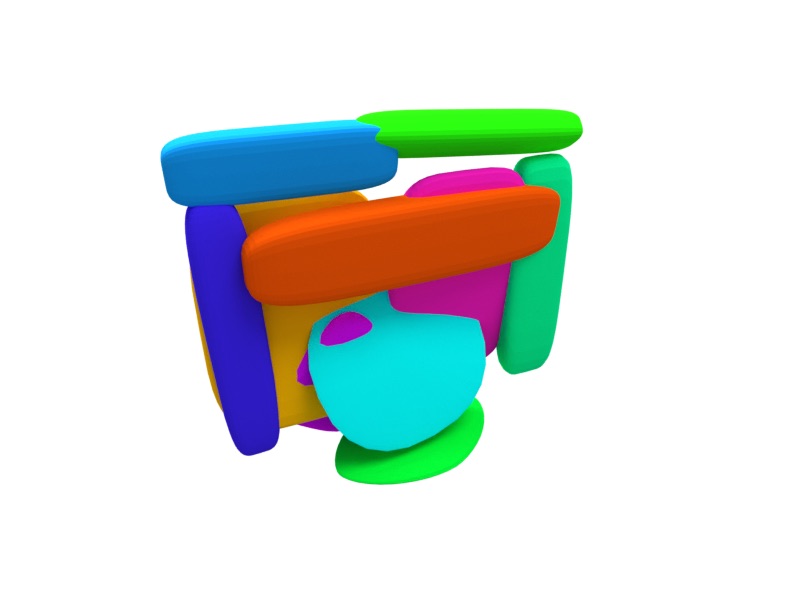}
\includegraphics[width=0.16\linewidth,trim={100px 100px 100px 100px},clip]{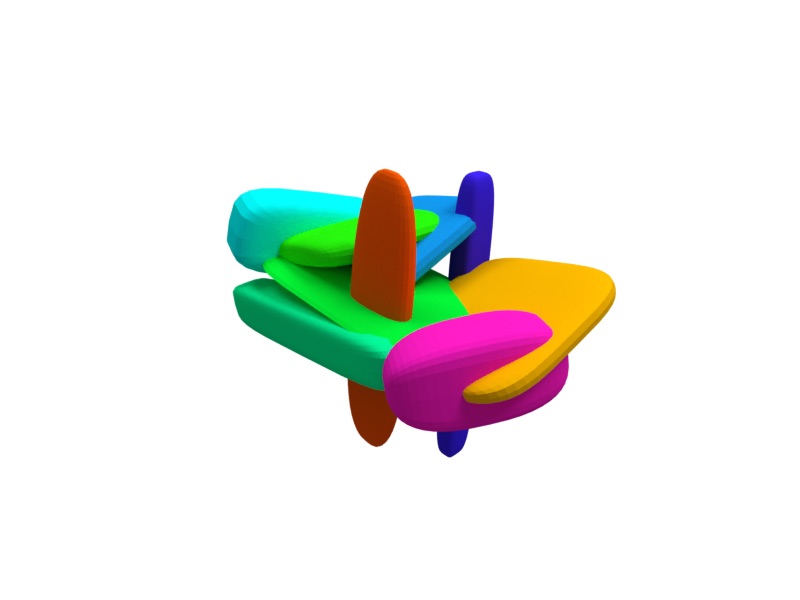}
\includegraphics[width=0.16\linewidth,trim={100px 100px 100px 100px},clip]{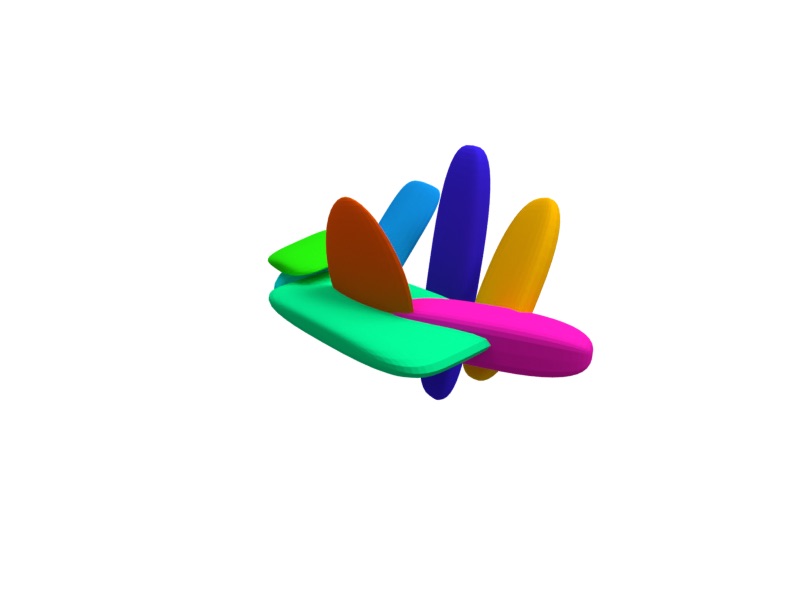}\\
\vspace{-12px}
\rotatebox{90}{\hspace{5px}\name{} (Ours)}
\hspace{-5px}
\includegraphics[width=0.16\linewidth,trim={50px 70px 50px 70px},clip]{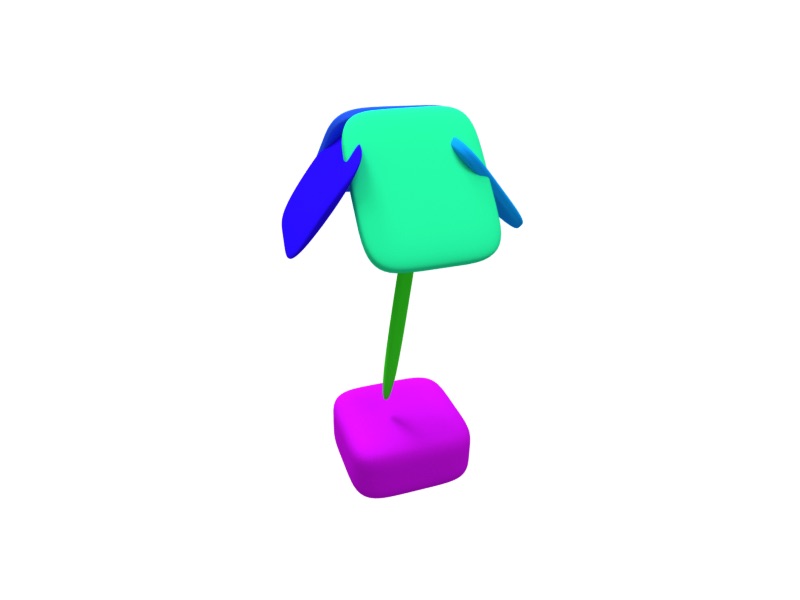}\hspace{-9px}
\includegraphics[width=0.16\linewidth,trim={50px 70px 50px 70px},clip]{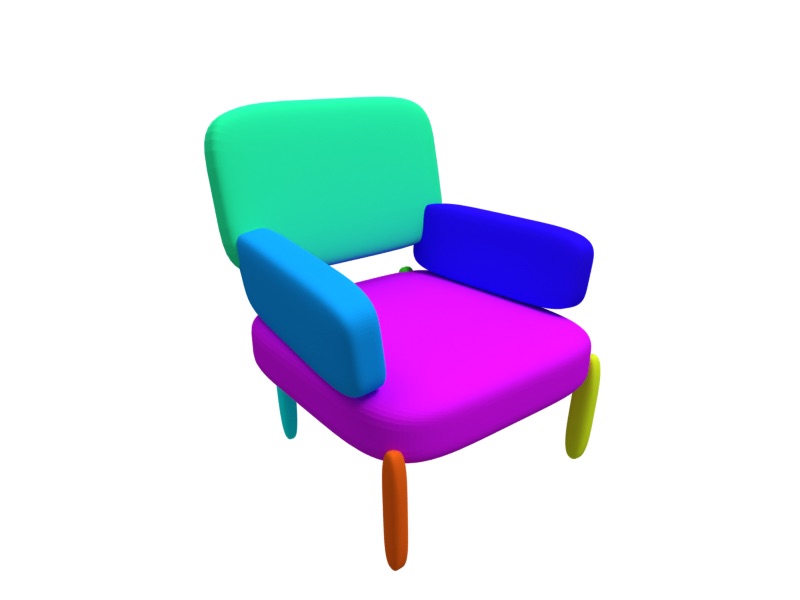}
\includegraphics[width=0.16\linewidth,trim={50px 70px 50px 70px},clip]{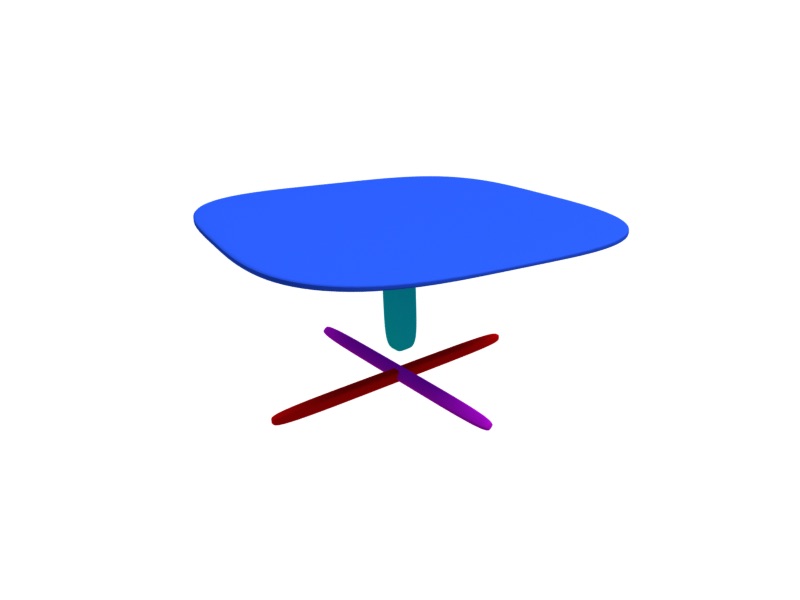}
\includegraphics[width=0.16\linewidth,trim={50px 70px 50px 70px},clip]{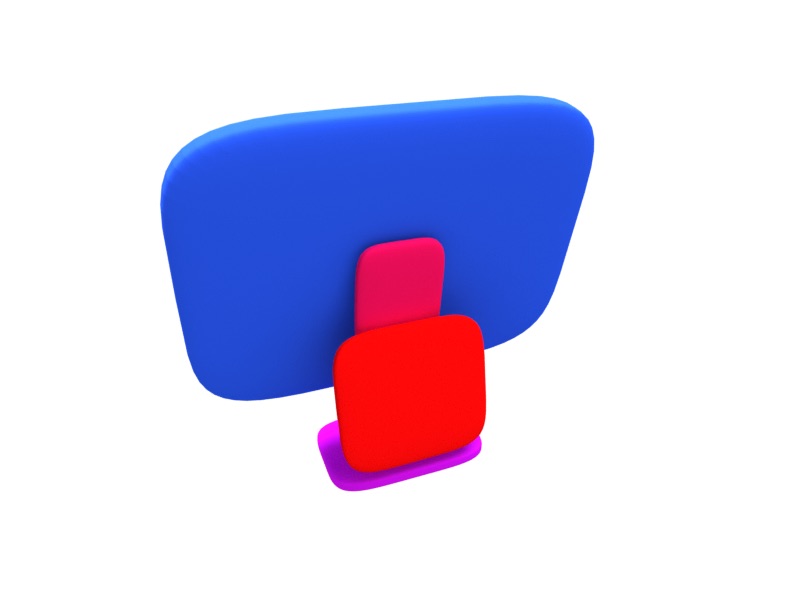}
\includegraphics[width=0.16\linewidth,trim={100px 100px 100px 100px},clip]{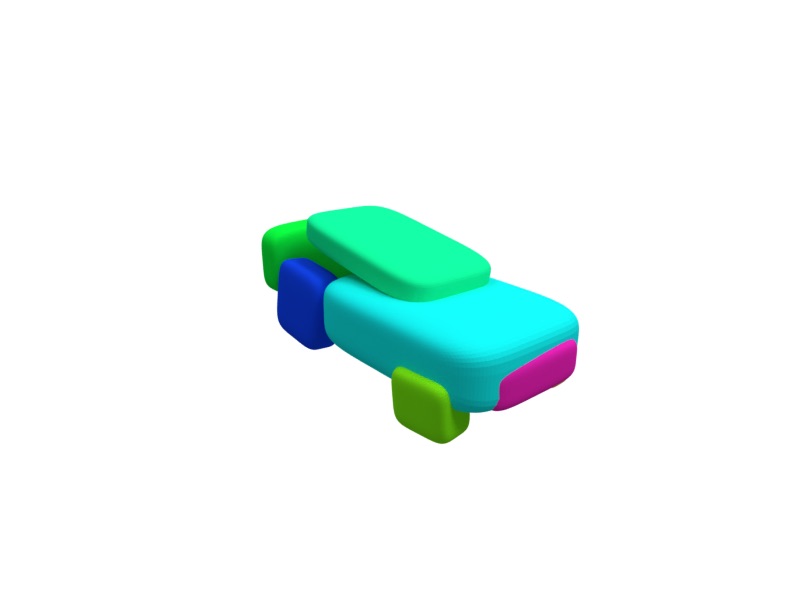}
\includegraphics[width=0.16\linewidth,trim={100px 100px 100px 100px},clip]{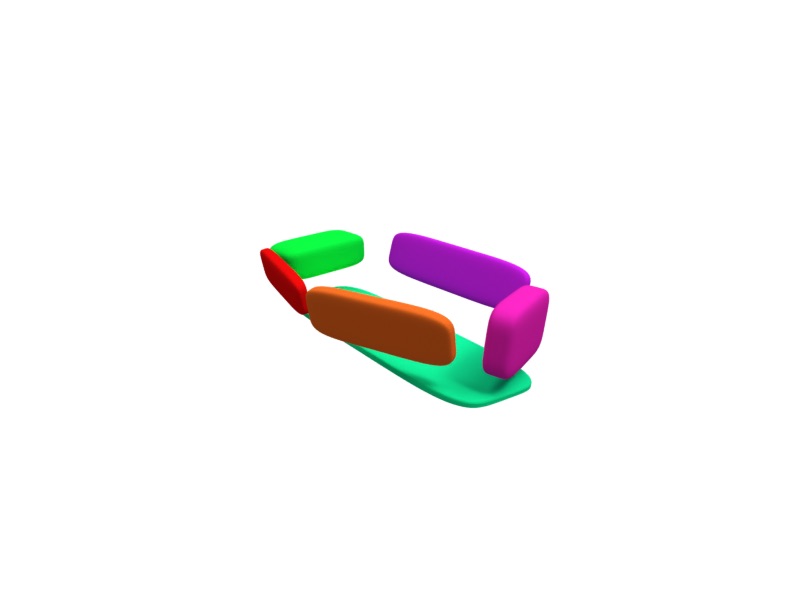}\\
\end{small}
\caption{\textbf{Qualitative Results on ShapeNet~\cite{shapenet2015}.}
We show results on test samples for in-category \emph{(four first columns)} classes and out-of-category classes \emph{(two last columns)}.
The latter were not seen during training and illustrate how well models generalize to novel classes.}
\label{fig:quali-shapenet}
\end{figure*}

\subsection{Comparing with State-of-the-art Methods}
\label{sec:eval}

\paragraph{Datasets.} We compare on three different datasets:
\noindent\textit{ShapeNet}~\cite{shapenet2015}:
We use the $13$ classes of the ShapeNet subset and train-val-test splits as defined in Choy \etal~\cite{choy20163d}. All objects are pre-aligned in a canonical orientation. For each object we sample $4096$ points using Farthest Point Sampling (FPS)~\cite{qi2017pointnetplusplus}. ShapeNet is a widely used dataset and is well-suited for comparison with existing baselines.\\ 
\noindent\textit{ScanNet++} ~\cite{Yeshwanth2023ScanNetAH}:
We further evaluate our model on real-world object scans from the ScanNet++ validation set.
Each object is extracted using ground truth mask annotations, and $4096$ points per object are sampled with FPS.
In contrast to ShapeNet, these object point clouds are noisier, partially observed, and subject to random orientation and translation, providing a more realistic and challenging evaluation setting for our method.\\
\noindent\textit{Replica}~\cite{Straub2019TheRD}:
Finally, we present both quantitative and qualitative results on full 3D scenes from Replica.
Object instances are extracted using either ground truth instance annotations or the pre-trained 3D instance segmentation model Mask3D~\cite{Schult2022Mask3DMT}, enabling us to evaluate our approach in a fully realistic setting, including scenarios where no ground truth annotations are available.

\noindent\textbf{Methods in comparison.}
We compare to learning- and optimization-based prior works using both cuboids and superquadrics as geometric primitives.
\textit{SQ}~\cite{Paschalidou2019CVPR} is a learning-based approach for object-level decomposition using superquadrics; it takes a voxel grid as input and predicts superquadric primitives via a CNN.
\textit{CSA}~\cite{Yang2021UnsupervisedLF} is another learning-based method but uses cuboids as geometric primitives. It takes a point cloud as input and predicts cuboid parameters from a global latent code.
Lastly, \textit{EMS}~\cite{liu2022robust} is an optimization-based approach that decomposes objects by hierarchically fitting superquadrics to parts of a point cloud.
\vspace{-10px}
\paragraph{Training Details.}
Our goal is to develop a general-purpose, class-agnostic model capable of representing arbitrary objects as superquadrics.
Existing methods typically train separate models for each object class, assuming that all classes are known in advance and that sufficient training data is available for each.
These assumptions, however, often fail in real-world scenarios.
To address this, we jointly train a single model on all 13 ShapeNet classes using the publicly available code of prior methods,
moving towards a more realistic \emph{class-agnostic} solution.
In our model we set the following hyper-parameters
$P$\,=\,$16$, $S$\,=\,$4096$, $K$\,=\,$25$, $H$\,=\,$128$, $D$\,=\,$3$, $\epsilon_{exist}$\,=\,$24$, $\wexist$\,=\,$0.01$, $\wpar$\,=\,$0.06$. We refer to the supplementary for additional details.

\vspace{-10px}
\paragraph{Metrics.}
\noindent We assess \textit{reconstruction accuracy} using L1 and L2 Chamfer distances and \textit{compactness} by the average number of geometric primitives.

\subsubsection{Results on ShapeNet}
\label{sec:shapenet}
We show scores in Tab.~\ref{tab:eval-shapenet} and qualitative results in Fig.~\ref{fig:quali-shapenet}.
To evaluate both accuracy and generalization, we conduct two experiments: \textit{in-category} and \textit{out-of-category}.
In the \emph{in-category} setting, all learning-based methods are jointly trained on the 13 classes of the ShapeNet training set and evaluated on the corresponding test set.
In the \emph{out-of-category} setting, models are trained on half of the categories (\textit{airplane}, \textit{bench}, \textit{chair}, \textit{lamp}, \textit{rifle}, \textit{table}) and tested on the remaining ones (\textit{car}, \textit{sofa}, \textit{loudspeaker}, \textit{cabinet}, \textit{display}, \textit{telephone}, \textit{watercraft}).
Our \name{} model significantly outperforms both learned and non-learned baselines.
Compared to learned baselines, we reduce the L2 loss by a factor of six while using nearly half the number of primitives,
supporting our hypothesis that leveraging local point features as opposed to a single global descriptor improves 3D decomposition in both accuracy and compactness.
Compared to the non-learned baseline, we predict a similar number of primitives but achieve an L2 loss approximately 20 times smaller,
validating the benefit of learning shape priors to avoid local minima that often hinder purely optimization-based approaches.

\subsubsection{Results on 3D Scenes}\label{sec:scenes}
In this section, models are evaluated on real-world, out-of-category objects, which appear in arbitrary orientations and often exhibit incomplete point clouds due to reconstruction artifacts and occlusions. Tab.~\ref{tab:eval-scannetpp} shows the quantitative results.
Despite never being trained on real-world objects, our method outperforms both the optimization- and the learning-based baselines by a large margin. Lastly, we qualitatively evaluate our pipeline on full 3D scenes from Replica, where our object-level model is applied on top of class-agnostic instance segmentation predictions from Mask3D~\cite{Schult2022Mask3DMT}. As shown in Fig.~\ref{fig:teaser}, our method effectively reconstructs object shapes, even under noisy segmentation masks and geometries that differ substantially from those seen during training. We refer to the supplementary for additional qualitative results.
\begin{table}[t]
\centering

\setlength{\tabcolsep}{7pt}
\resizebox{\columnwidth}{!}{
\begin{tabular}{l ccc|ccc }
\toprule
&  \multicolumn{3}{c}{\textbf{ScanNet++}~\cite{Yeshwanth2023ScanNetAH}} & \multicolumn{3}{c}{\textbf{Replica}~\cite{Straub2019TheRD}} \\
\midrule 
\textbf{Method}& \textbf{L1} $\downarrow$ &
\textbf{L2} $\downarrow$ & 
\textbf{\#\,Prim.}$\downarrow$ 
&\textbf{L1} $\downarrow$ &
\textbf{L2} $\downarrow$ & 
\textbf{\#\,Prim.}$\downarrow$\\
\midrule
SQ~\cite{Paschalidou2019CVPR}  & $10.45$ & $4.26$ & $10.0$& $11.84$ & $5.91$ & $10$\\
EMS~\cite{liu2022robust}  & $5.51$& $2.11$ & $\mathbf{4.25}$ & $5.40$& $2.12$ & $\mathbf{3.61}$ \\
CSA~\cite{Yang2021UnsupervisedLF}  & $2.91$& $0.41$ & $11.64$ & $3.68$& $0.70$ & $9.63$\\
\name{} (Ours)  & $\mathbf{1.70}$&  $\mathbf{0.11}$ & $5.18$ &$\mathbf{1.79}$& $\mathbf{0.19}$ & $6.58$\\ 
\bottomrule
\end{tabular}
}
\caption{\textbf{Quantitative Results on Scenes.} Evaluation on objects from 3D scene datasets~\cite{Straub2019TheRD,Yeshwanth2023ScanNetAH}. Scores are scaled by $10^2$.} 
\label{tab:eval-scannetpp}
\end{table}




\subsection{Down-stream Applications}
\label{sec:applications}
Next, we show the versatility of the \name{} representation for downstream applications,
including robotics tasks such as path planning and object grasping (Sec.~\ref{sec:app-robotics}), and controllable image generation (Sec.~\ref{sec:app-generation-editing}).

\subsubsection{Robotics}\label{sec:app-robotics}
\paragraph{Path planning} seeks to compute a collision-free shortest path between a given start and end point in 3D space, enabling efficient robot navigation. Although essential for traversing large environments, it typically demands storing large-scale 3D representations. Here, we assess whether our compact representation can perform this task effectively while reducing memory requirements.
We conduct experiments on 15 ScanNet++~\cite{Yeshwanth2023ScanNetAH} scenes, comparing \name{} to common 3D representations, including dense occupancy grids, point clouds, voxel grids, and cuboids~\cite{ramamonjisoa2022monteboxfinder}. As shown in Tab.~\ref{tab:exp_robot}, \name{} not only reduces memory consumption compared to traditional representations but also achieves a higher success rate than dense point clouds. Further details about experiment setup, metrics and analysis are provided in the Appendix.
\begin{table}[t]
\centering
\small
\setlength{\tabcolsep}{7pt}
\begin{tabular}{lccc}
\toprule
\textbf{Method} & \textbf{Time (ms)} & \textbf{Suc. (\%)}  & \textbf{Mem. (MB)} \\
\midrule
Occupancy  & 0.056  & \textbf{100.00}    & 0.873 \\
PointCloud  & 0.063  & 89.57    & 19.286 \\
Voxels  & \textbf{0.030}  & 98.78   & 0.101 \\
Cuboids~\cite{ramamonjisoa2022monteboxfinder}  & 0.120  & 61.23    & \textbf{0.024} \\
\name{} & 0.150  &91.71   & 0.042 \\
\bottomrule
\end{tabular}
\caption{\textbf{Path Planning Scores.} Mean on 15 ScanNet++ scenes.}
\vspace{-5px}
\label{tab:exp_robot_summary}
\end{table}

\begin{figure}[t]
    \centering
    \includegraphics[width=0.59\linewidth, trim={18px 0px 110px 100px},clip]{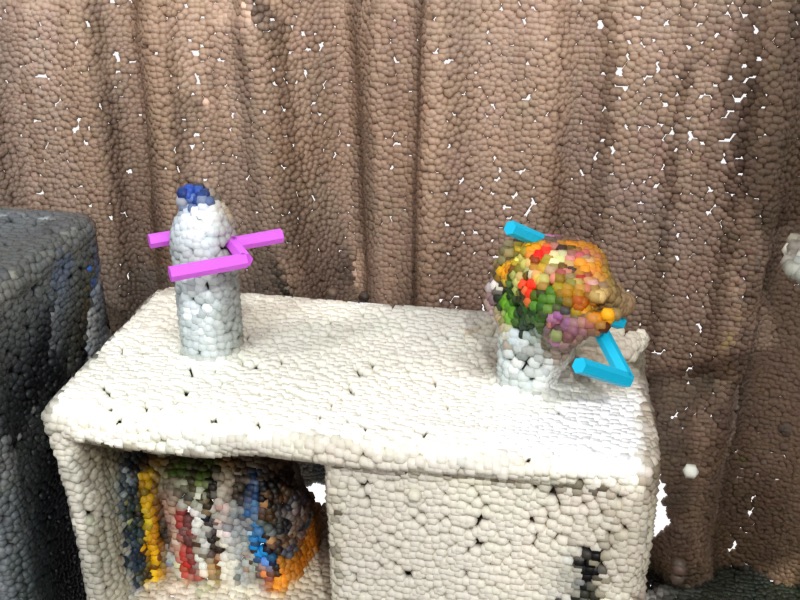}
    \includegraphics[width=0.39\linewidth, trim={10px 0px 280px 20px},clip]{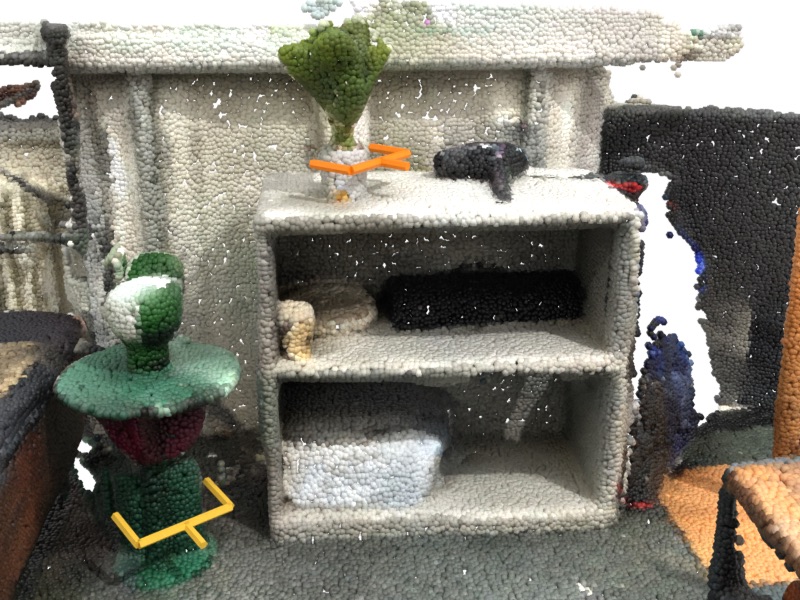}
    \caption{\textbf{Grasping Result.} Visualization of computed grasp poses for a {\color{magenta}milk bottle}, some {\color{Cerulean}flowers}, a {\color{Dandelion}side table}, and a {\color{BurntOrange}plant}. }
    \label{fig:real_grasp}
    \vspace{-1.\baselineskip}
\end{figure}
\vspace{-10px}

\paragraph{Object Grasping} enables robots to grasp real-world objects by computing suitable grasping poses. Existing methods fall into two categories, each with complementary limitations. Geometry-based approaches~\cite{cai2022volumetric,1241860,131614} require precise 3D object models, which are often unavailable in real-world scenarios. Learning-based approaches~\cite{wang2021graspness,mahler2019learning,sundermeyer2021contact} operate directly on raw sensor data but tend to be biased towards training data, which typically consists of tabletop scenes with small, convex, or low-genus objects~\cite{fang2023anygrasp,sudry2023hierarchical}. To overcome these limitations, SuperQ-GRASP~\cite{tu2024superq} explored decomposing objects into explicit primitives. However, its reliance on Marching Primitives~\cite{liu2023marching} to obtain superquadrics from the object's Signed Distance Function (SDF) makes it unsuitable for most real-world cases where only point clouds are available. In contrast, our approach directly processes point clouds of entire scenes and, when combined with the class-agnostic segmentations from Mask3D~\cite{Schult2022Mask3DMT}, extracts superquadrics for all objects. Given the superquadric parameters, we employ a superquadric-based geometric method~\cite{vezzani2017grasping} to compute grasping poses for selected objects. Fig.~\ref{fig:real_grasp} shows predicted grasping poses on objects from a real-world 3D scan of a room. In practice, our method eliminates the need for data-driven grasping models while remaining adaptable to diverse object shapes and producing high-quality grasping poses.

\begin{figure}
\centering
\begin{tabular}{cc}
\vspace{-15px}
\hspace{0.44\linewidth} & \hspace{0.45\linewidth} \vspace{-0.1cm}\\
\footnotesize{\emph{Input Point Cloud}}& \footnotesize{\textit{Superquadrics}}
\end{tabular}
    \includegraphics[width=0.495\linewidth, trim={0px 0px 0px 150px}, clip]{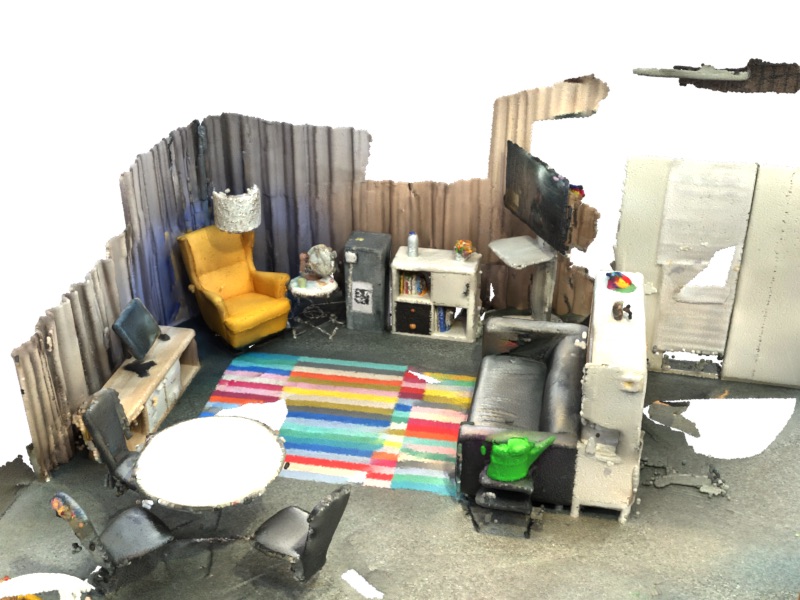}\hspace{0.01\linewidth}\includegraphics[width=0.495\linewidth, trim={0px 120px 0px 30px}, clip]{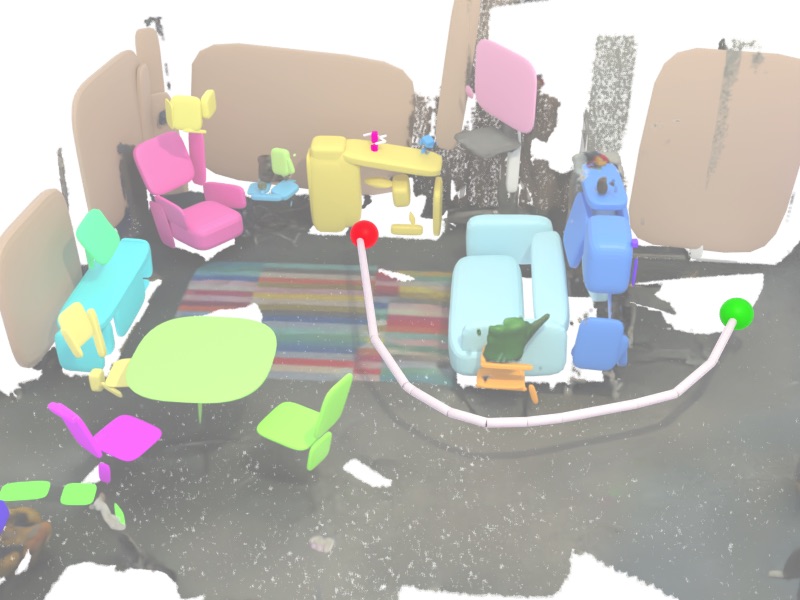}\\\includegraphics[width=\linewidth, trim={0px 100px 0px 120px},clip]{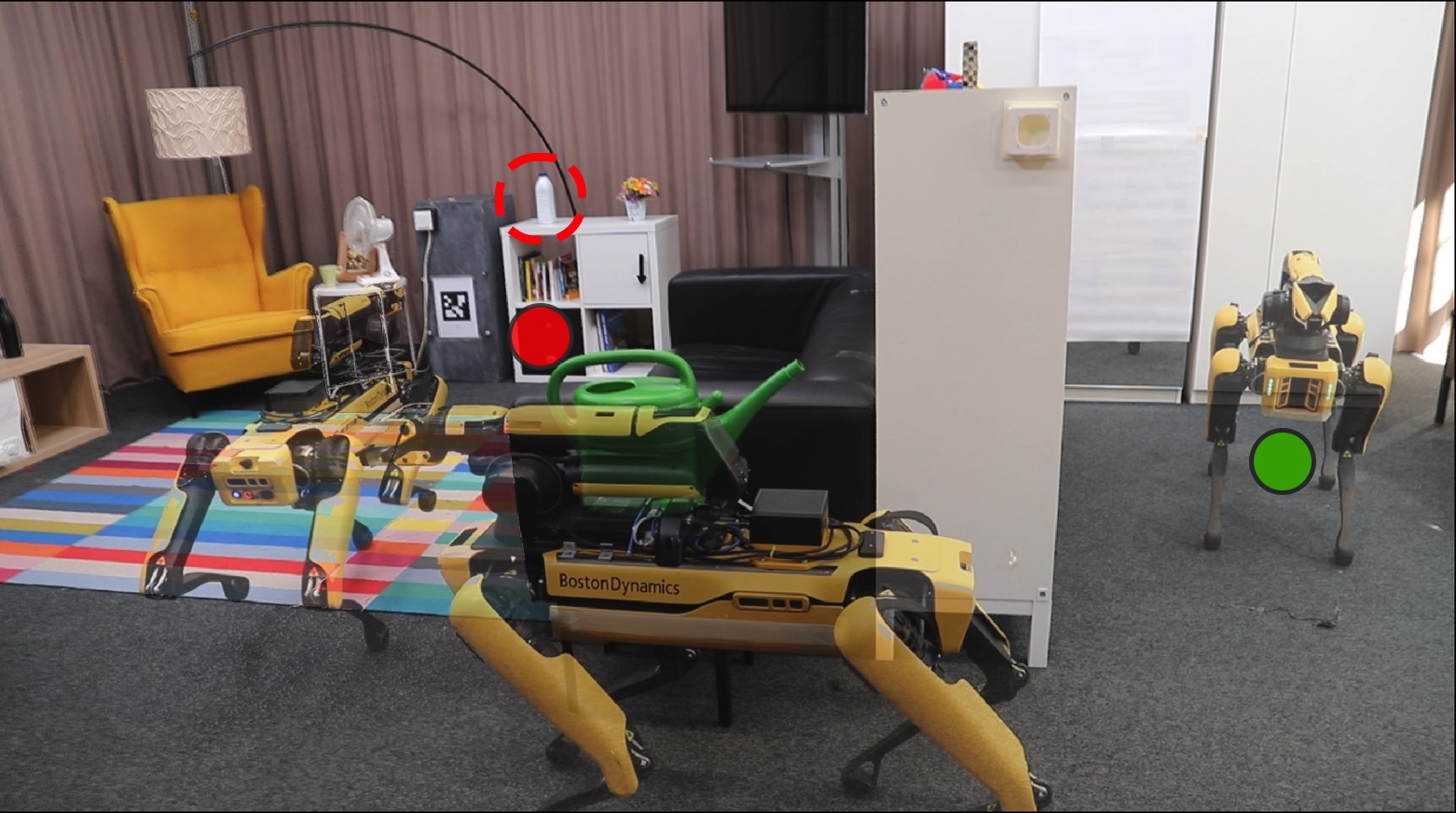}
    \caption{\textbf{Real-world robot experiment.} The top row shows the input scan (\textit{left}) and the representation from \name{} with the computed path and grasping pose (\textit{right}). The bottom row illustrates the robot following the planned path. We denote the starting point of the path with a green sphere, and the target location with a red sphere. The target object (a milk bottle) is circled in red.\vspace{-10px}}
    \label{fig:real_world}
\end{figure}
\vspace{-10px}

\paragraph{Real-world Experiment.}
Finally, we demonstrate the real-world applicability of our superquadric-based representation by deploying it on a legged robot (Boston Dynamics Spot) equipped with an arm, supporting both motion planning and object grasping in an indoor environment.
We scan a scene using a 3D scanning application on an iPad, extract a dense point cloud, and run \name{} on it. Given the robot’s starting position and a specified target object (a milk bottle), we compute both the path and the grasping pose as described earlier (see Fig.~\ref{fig:real_grasp}), enabling the robot to approach and successfully grasp the object. Fig.~\ref{fig:real_world} shows the computed representation, the planned trajectory, the grasping pose, and a frame from the real-world demonstration. This experiment suggests that integrating \name{} with open-vocabulary segmentation methods such as OpenMask3D~\cite{takmaz2023openmask3d} could allow robots to navigate to and grasp arbitrary objects specified via natural-language prompts.

\subsubsection{Controllable Generation and Editing}
\label{sec:app-generation-editing}
\begin{figure}[t]
\centering
\begin{small}
\vspace{-15px}
\begin{tabular}{cccc}
\hspace{0.2\linewidth} & \hspace{0.2\linewidth} & \hspace{0.2\linewidth} & \hspace{0.2\linewidth} \\
 \emph{Original} & \emph{Editing} &\emph{Addition}& \textit{Deletion} 
\end{tabular}
\\
\includegraphics[width=0.25\linewidth,trim={170px 0px 100px 0px},clip]{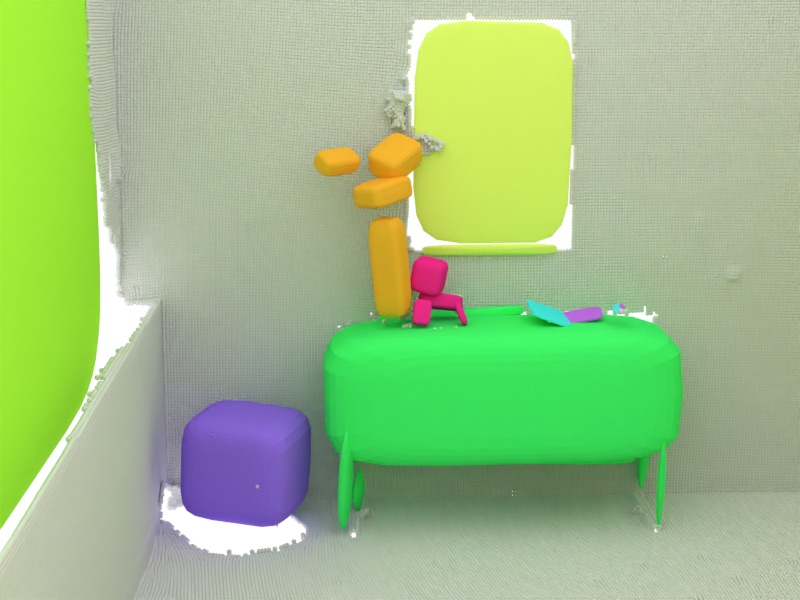}
\includegraphics[width=0.25\linewidth,trim={170px 0px 100px 0px},clip]{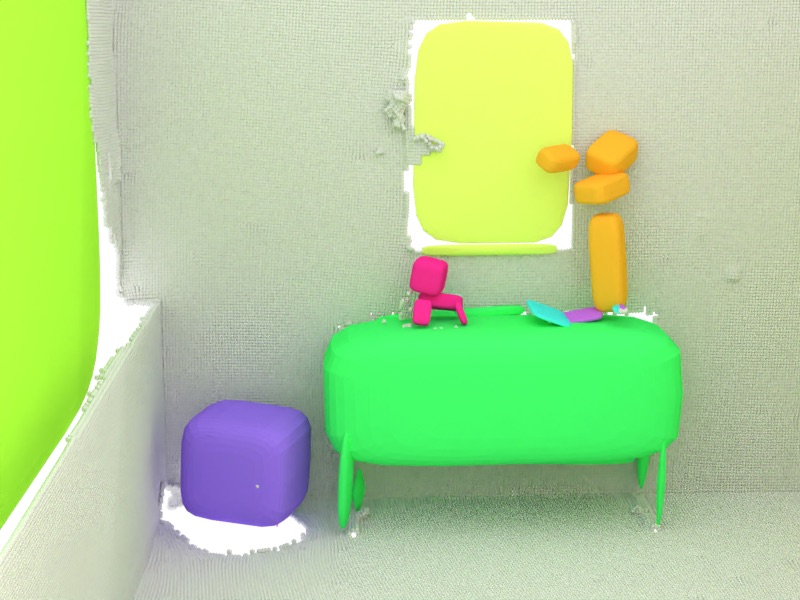}%
\includegraphics[width=0.25\linewidth,trim={170px 0px 100px 0px},clip]{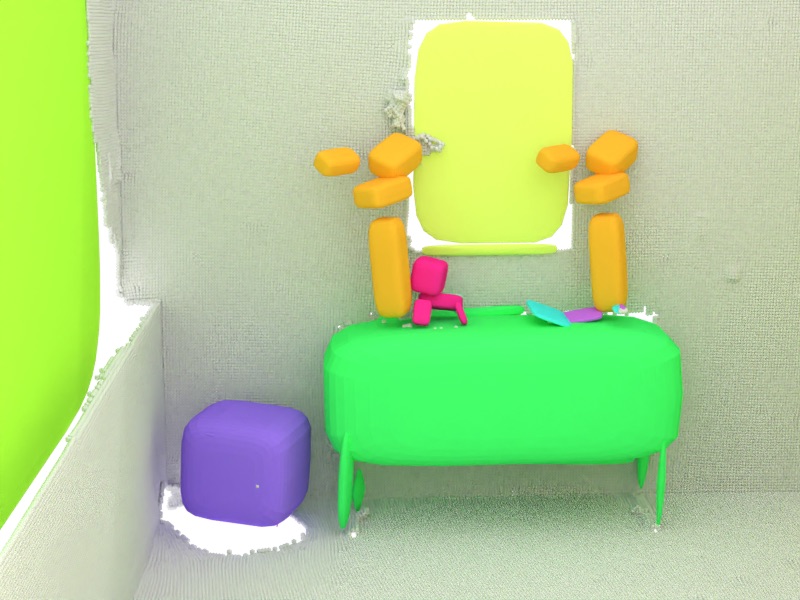}%
\includegraphics[width=0.25\linewidth,trim={170px 0px 100px 0px},clip]{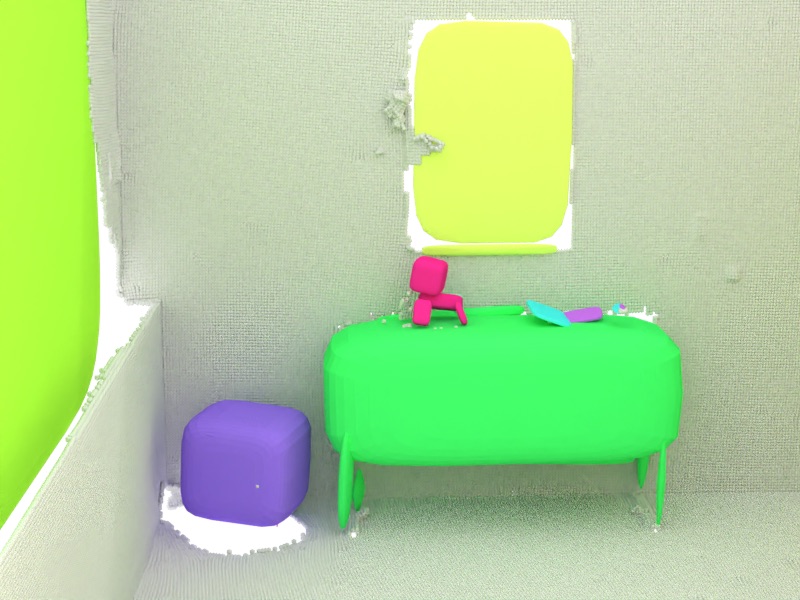}\\
\vspace{-1.0px}
\includegraphics[width=0.25\linewidth,trim={170px 0px 100px 0px},clip]{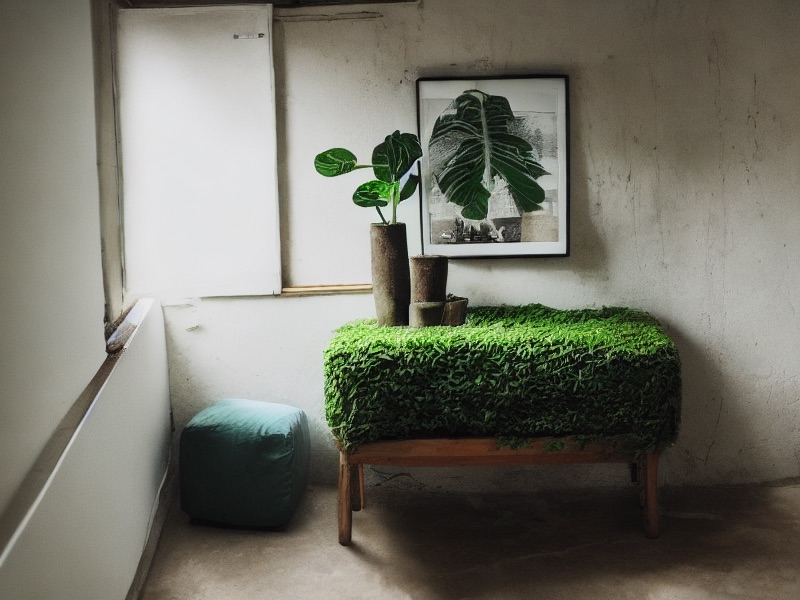}%
\includegraphics[width=0.25\linewidth,trim={170px 0px 100px 0px},clip]{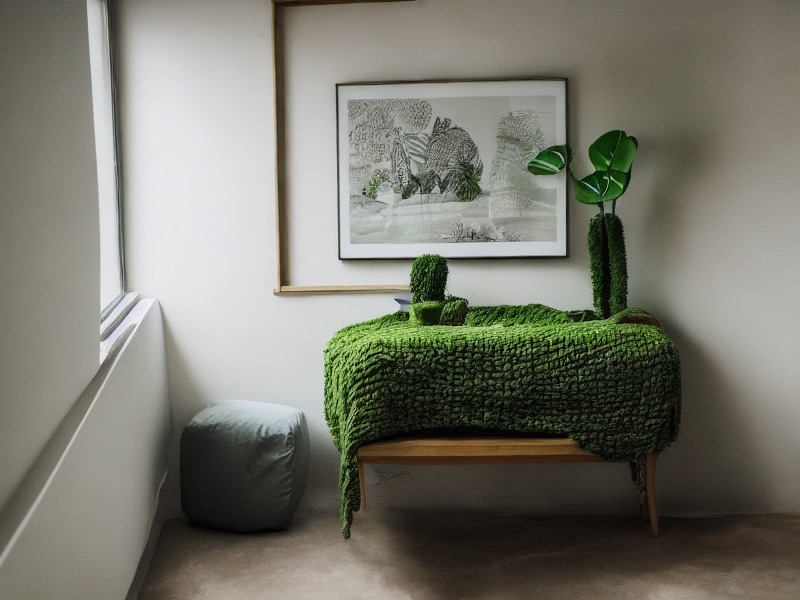}%
\includegraphics[width=0.25\linewidth,trim={170px 0px 100px 0px},clip]{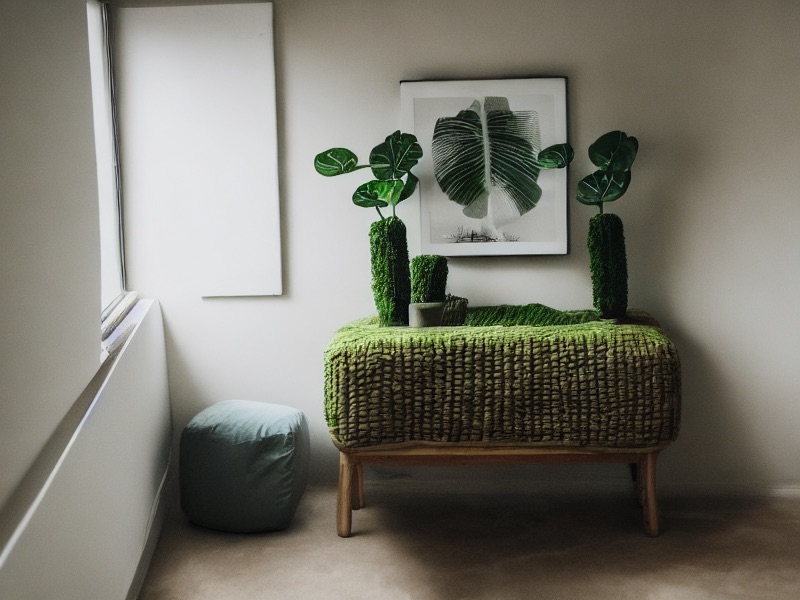}%
\includegraphics[width=0.25\linewidth,trim={170px 0px 100px 0px},clip]{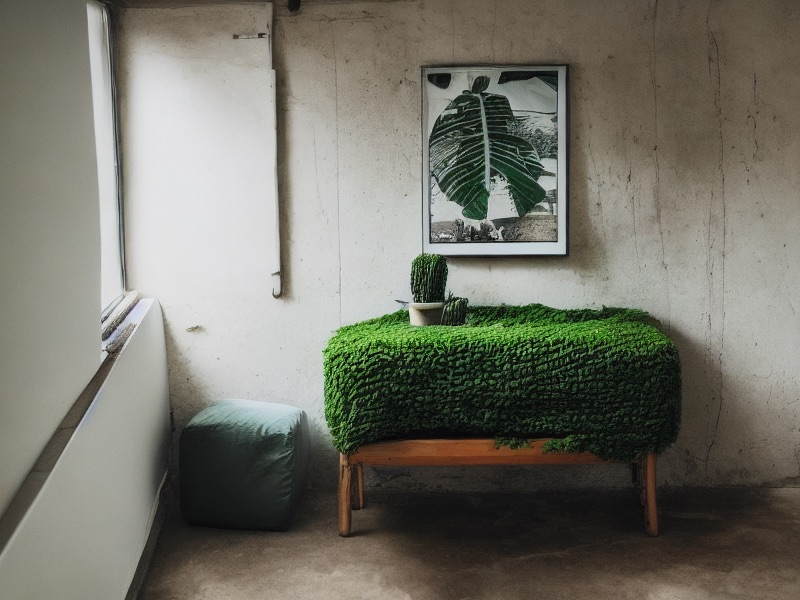}\\
\end{small}
\vspace{-1px}
\caption{\textbf{\textit{Spatial} control using \name{}.} Top row shows superquadrics generated by \name{}, bottom row shows generated images using the prompt \textit{'A corner of a room with a plant'}.}
\label{fig:3d-generation}
\vspace{-10px}
\end{figure}

\begin{figure}
\centering
\begin{small}
\begin{tabular}{cc}
\hspace{0.44\linewidth} & \hspace{0.44\linewidth} \vspace{-0.1cm}\\
\emph{Superquadrics} & \emph{Depth Prompt}
\end{tabular}
\\
\vspace{2px}
\includegraphics[width=0.49\linewidth,trim={50px 70px 50px 70px},clip]{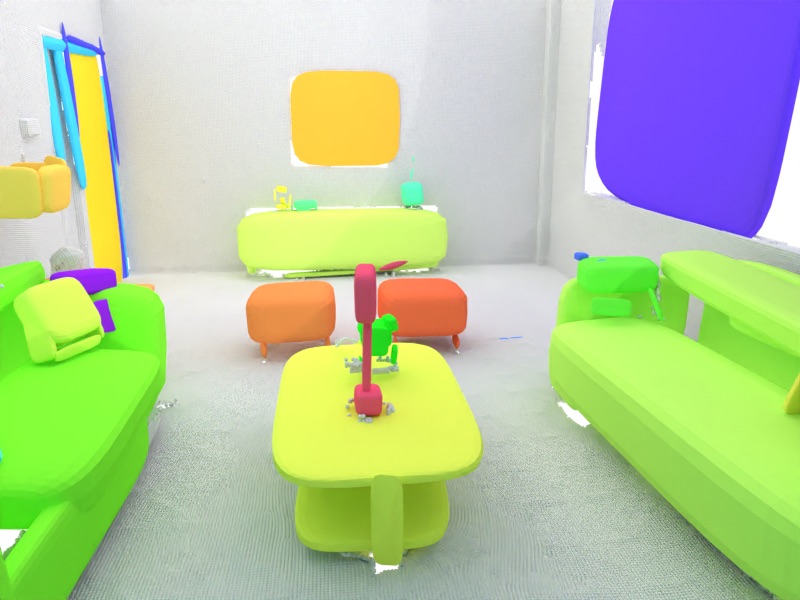}%
\hspace{0.1cm}\includegraphics[width=0.49\linewidth,trim={50px 70px 50px 70px},clip]{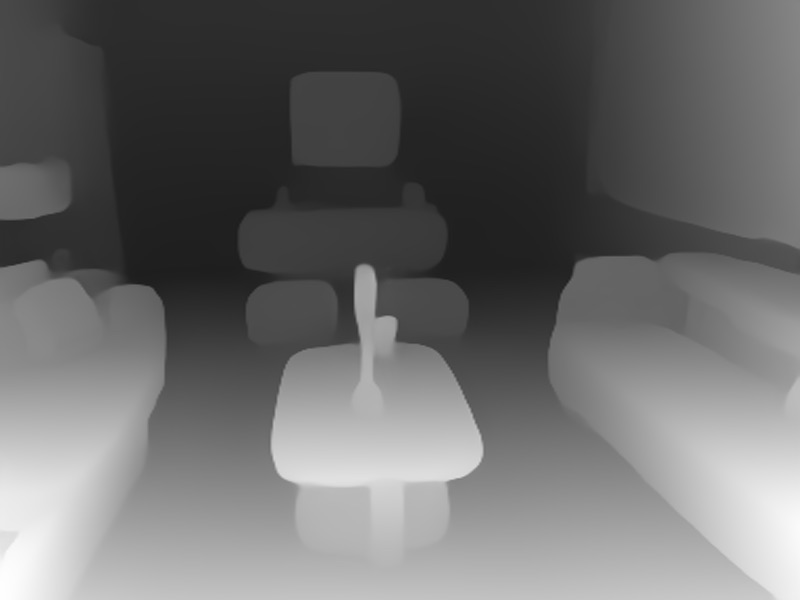}%
\\
\vspace{-0.3cm}
\begin{tabular}{cc}
\hspace{0.43\linewidth} & \hspace{0.43\linewidth} \vspace{-0.1cm}\\
\emph{``\textbf{Pink} living room"}& \textit{``\textbf{Modern} living room"}
\end{tabular}
\\
\includegraphics[width=0.49\linewidth,trim={50px 70px 50px 70px},clip]{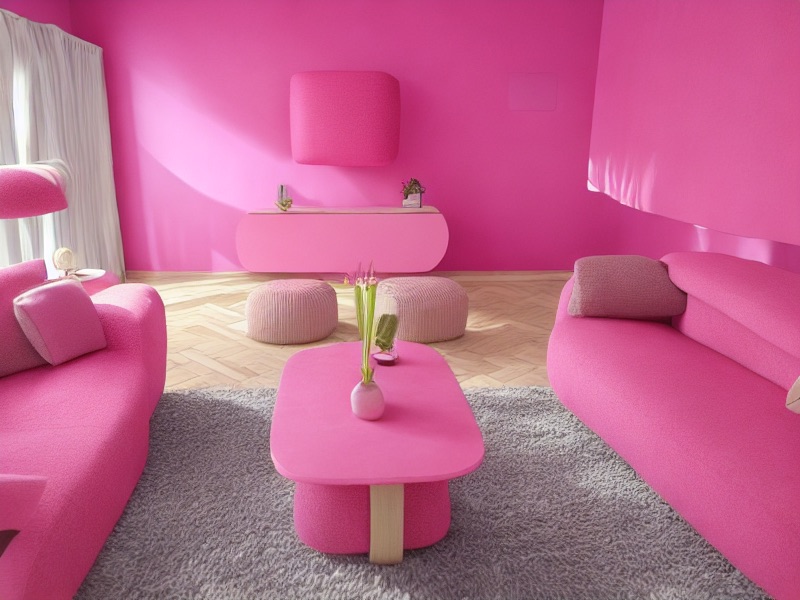}%
\hspace{0.1cm}\includegraphics[width=0.49\linewidth,trim={50px 70px 50px 70px},clip]{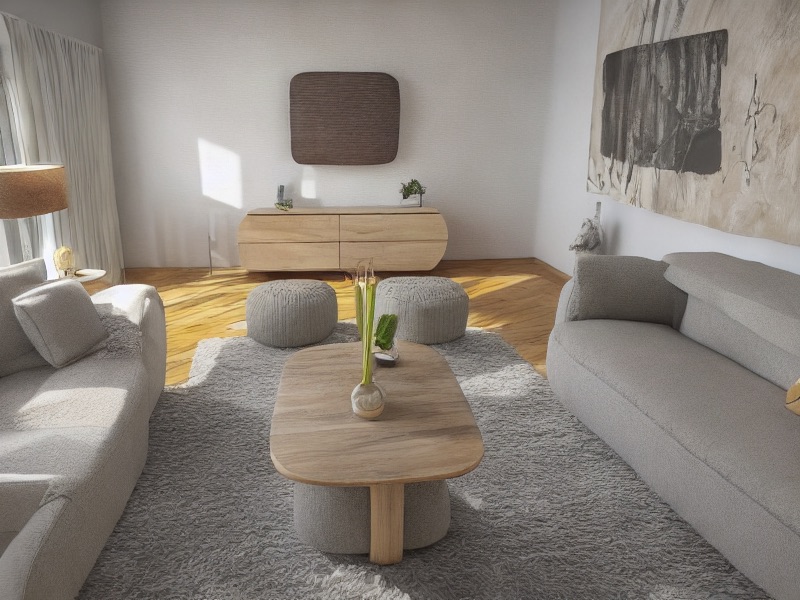}\\
\end{small}
\caption{\textbf{\textit{Semantic} control using \name{}.}
\emph{Top row}: superquadrics created by our \name{}, and depth map to prompt the generation of text-to-image diffusion model. \emph{Bottom row}: generations with two different textual prompts. \vspace{-10px}}
\label{fig:3d-generation-room0}
\end{figure}

We investigate how the \name{} representation can be used to introduce joint spatial and semantic control in text-to-image diffusion models~\cite{rombach2021highresolution}. Specifically, we generate images by conditioning ControlNet~\cite{Zhang2023AddingCC} on depth maps rendered from the superquadrics extracted from Replica~\cite{Straub2019TheRD} scenes. Qualitative results are shown in Fig.~\ref{fig:3d-generation} and Fig.~\ref{fig:3d-generation-room0}. Fig.~\ref{fig:3d-generation} demonstrates spatial control: by \textit{moving}, \textit{duplicating}, or \textit{removing} superquadrics corresponding to a plant, we coherently influence the generated images. Fig.~\ref{fig:3d-generation-room0} highlights semantic control: we can vary the room's style while preserving its semantic and geometric structure, and observe that object semantics naturally emerge from the spatial arrangement of superquadrics without explicit conditioning, \eg{}, pillows appear on couches, and a plant is placed on a central table, reflecting plausible real-world arrangements.

\subsection{Analysis Experiments}
\label{sec:ablation}

\begin{figure}
\centering
\begin{small}
\begin{tabular}{cc}
 \hspace{4.8cm} & \hspace{2.55cm} \\
 \emph{In-category} & \textit{Out-of-category} \\
\cmidrule(lr){1-1}\cmidrule(lr){2-2} 
\end{tabular}
\\
\vspace{-2px}
\hspace{-12px}
\includegraphics[width=0.22\linewidth,trim={150px 70px 50px 70px},clip]{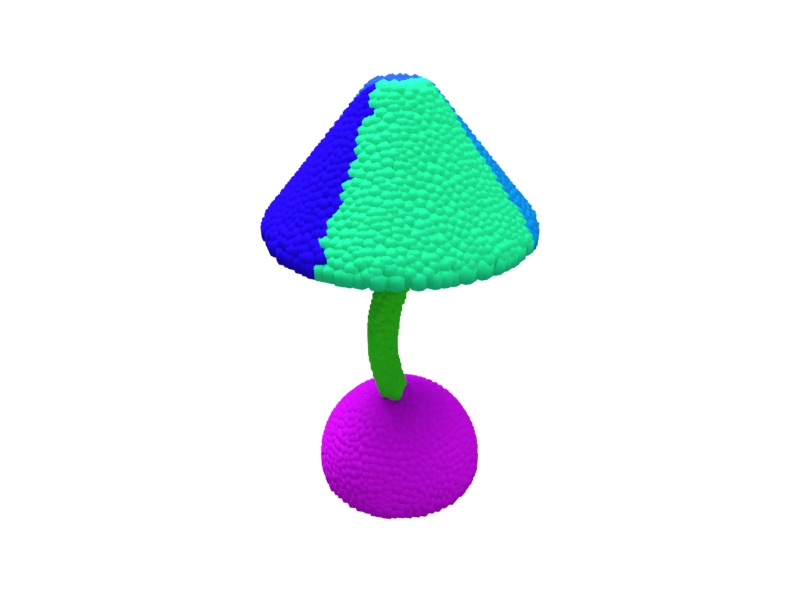}\hspace{-18px}
\includegraphics[width=0.22\linewidth,trim={150px 70px 40px 70px},clip]{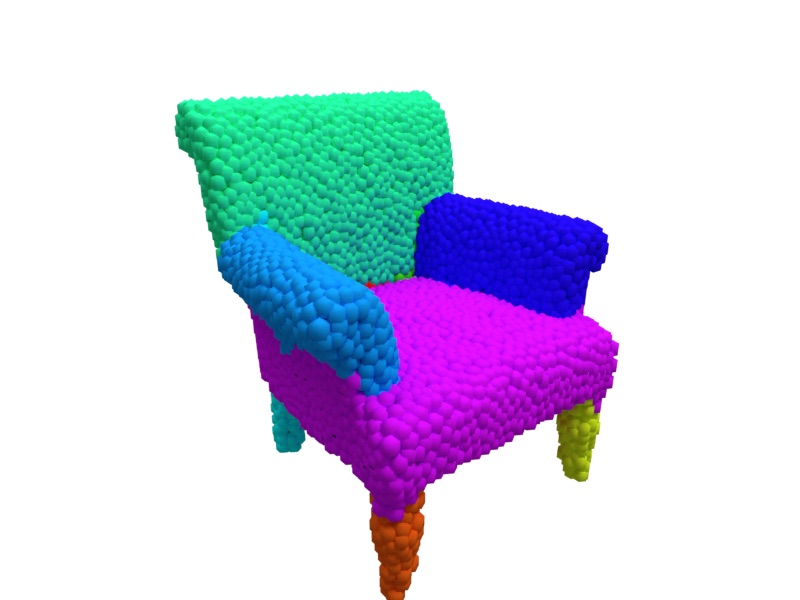}\hspace{-15px} 
\includegraphics[width=0.22\linewidth,trim={150px 70px 50px 70px},clip]{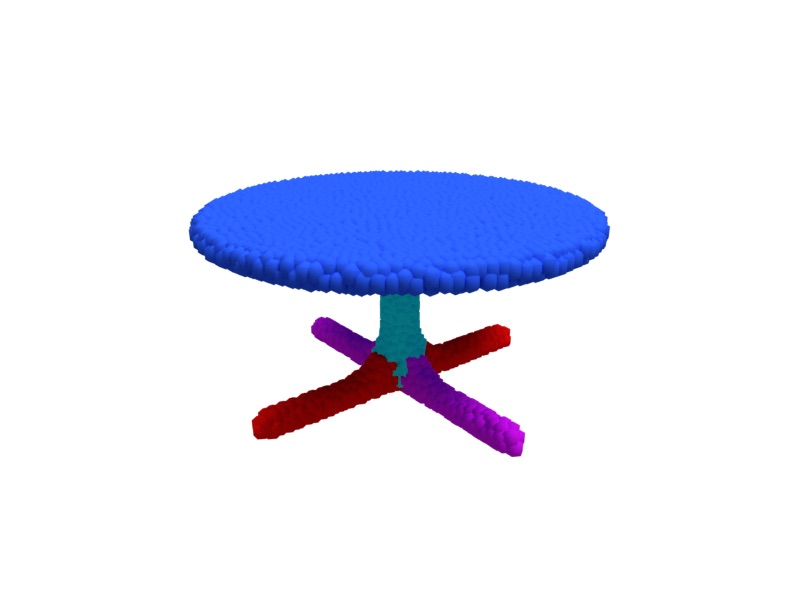}\hspace{-10px}
\includegraphics[width=0.22\linewidth,trim={130px 70px 5px 70px},clip]{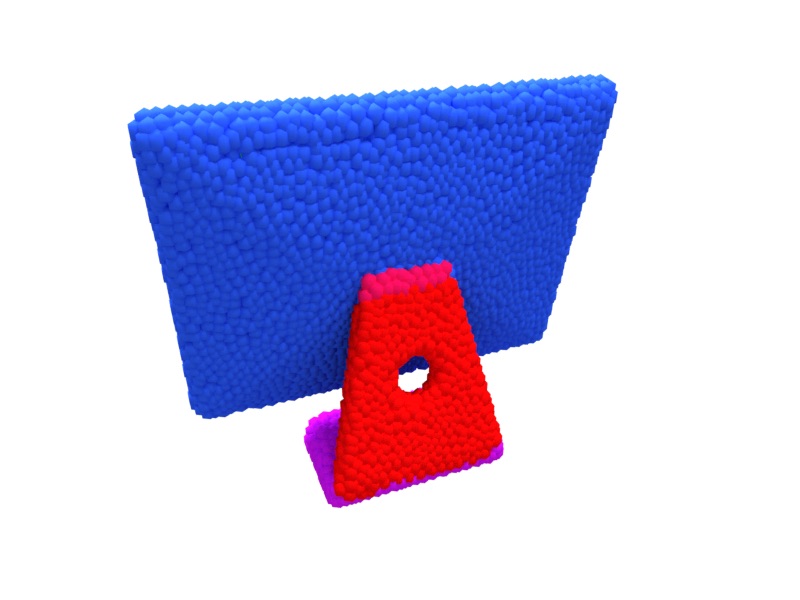}\hspace{-15px}
\includegraphics[width=0.22\linewidth,trim={180px 100px 100px 100px},clip]{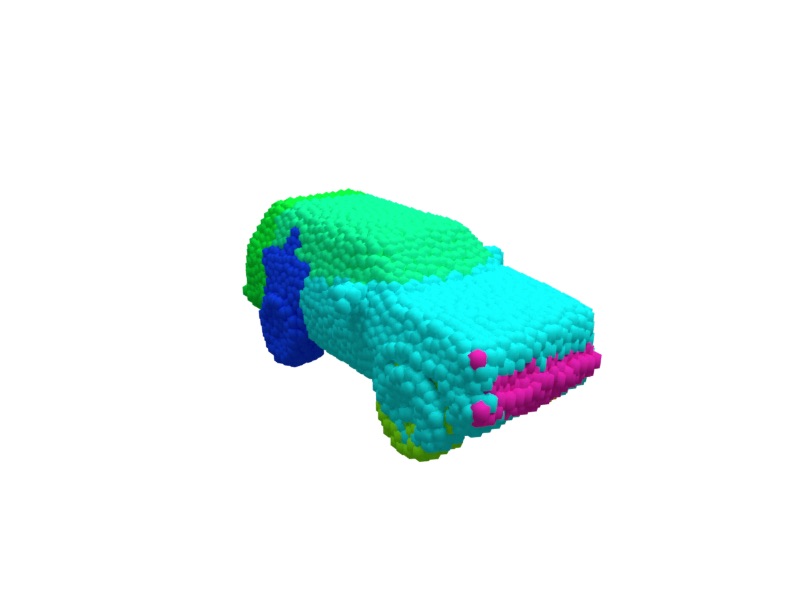}\hspace{-12px}
\includegraphics[width=0.18\linewidth,trim={240px 100px 150px 100px},clip]{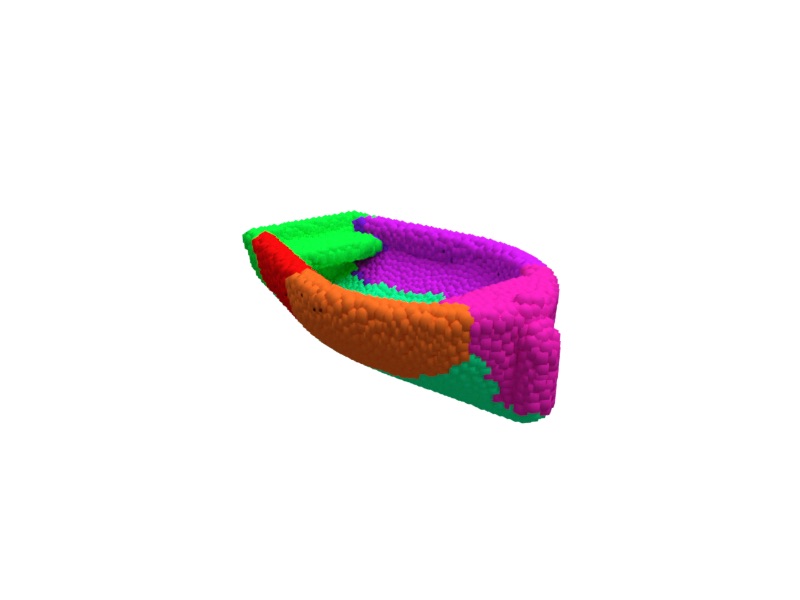}\\
\end{small}
    \caption{\textbf{Qualitative Results on ShapeNet~\cite{shapenet2015} segmentation.}
    We show the resulting segmentation matrices on test samples for in-category \emph{(four first columns)} classes and out-of-category classes \emph{(two last columns)}. The latter were not seen during training.}
    \label{fig:quali-shapenet-seg}
\end{figure}

\paragraph{Unsupervised part segmentation.}
Besides superquadric parameters, our method also predicts a segmentation matrix which segments the initial point cloud into parts that are fitted to the predicted superquadrics.
In Fig.~\ref{fig:quali-shapenet-seg}, we visualize the predicted segmentation masks for the same examples shown in Fig.~\ref{fig:quali-shapenet}. We observe that segmentation masks, especially in the \textit{in-category} experiments, appear very sharp. This suggests that our method, especially if trained at a larger scale, can be leveraged for different applications as geometry-based part segmentation or as pretraining for supervised semantic part segmentation.

\paragraph{Can we do hierarchical decomposition?}
We run \name{} in a hierarchical fashion on a whole simple scene, without using any instance segmentation method to segment the scene into different objects. We observed that by applying \name{} on the whole point cloud it can coarsely segment the scene into components such as grass, table, and chairs. By applying the model again on the points segmented during the first prediction, we obtained a more accurate decomposition of the individual object instances. Given that \name{} was trained only on single object instances from ShapeNet~\cite{shapenet2015} we found this results very interesting. We see future potential in the use of our model to obtain unsupervised hierarchical decompositions.

\begin{figure}
    \begin{overpic}[width=\columnwidth]{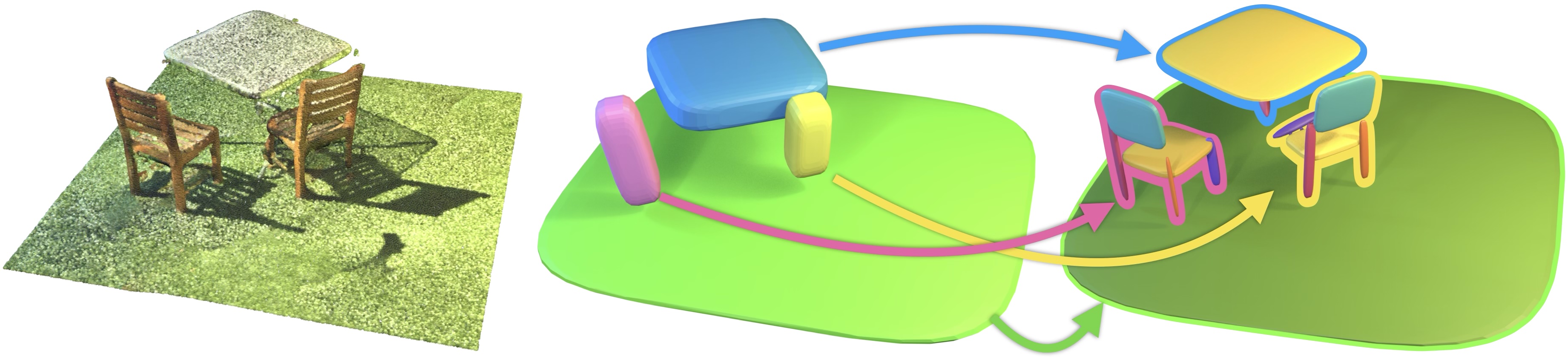}
  \put(4,-4){{\footnotesize \colorbox{white}{Input 3D Scan}}}
  \put(37,-4){{\footnotesize \colorbox{white}{1\textsuperscript{st} Hierarchy}}}
  \put(71,-4){{\footnotesize \colorbox{white}{2\textsuperscript{nd} Hierarchy}}}
\end{overpic}
\vspace{0.1cm}
\caption{\textbf{Hierarchical decomposition with \name{}.} From left to right we visualize: the input point cpngloud, the superquadrics predicted in the first hierarchy, the superquadrics predicted in the second hierarchy.}
\end{figure}

\begin{figure}[t]
    \centering
    \vspace{-10px}
    \includegraphics[width=0.8\linewidth, trim={60px 40px 10px 35px},clip]{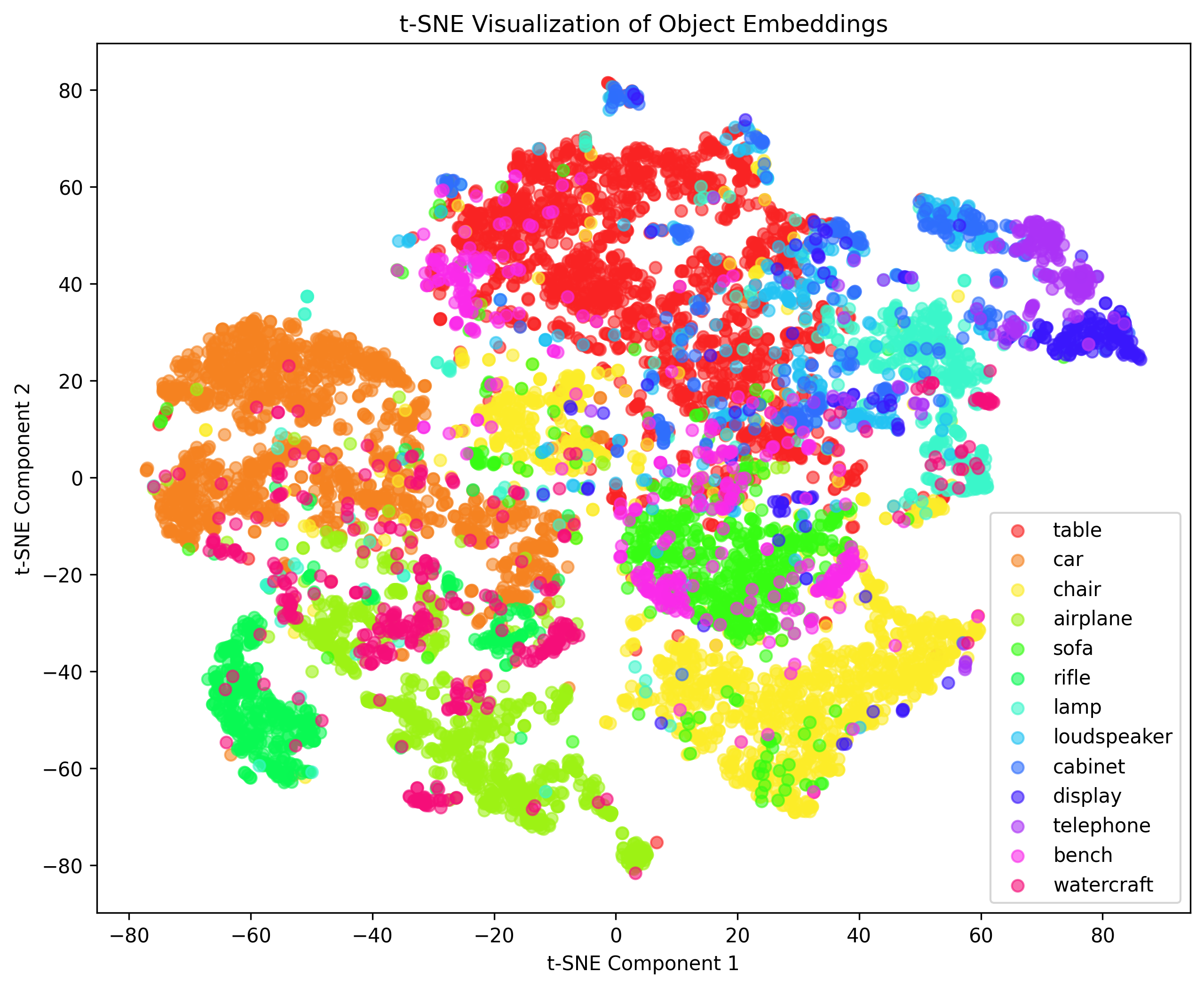}
    \caption{\textbf{t-SNE Visualization of Primitive Embeddings} across different ShapeNet classes.}
    \label{fig:tsne-embs}
\end{figure}

\vspace{-10px}
\paragraph{What does our network learn?}
Since our network performs unsupervised part segmentation, we analyze the features learned by the Transformer decoder across object classes. Inspired by BERT~\cite{devlin2019bert}'s \verb|[CLS]| token, we append a learnable embedding to the sequence of embedded superquadrics; although never explicitly decoded, this embedding is refined through self- and cross-attention. After training, we extract and visualize these embeddings using t-SNE~\cite{van2008visualizing} for ShapeNet~\cite{shapenet2015} categories (Fig.~\ref{fig:tsne-embs}). We observe that categories with consistent shapes, such as \textit{chairs}, \textit{airplanes}, and \textit{cars}, form clear clusters, while categories with high intra-class variability, such as \textit{watercraft}, are more dispersed. This indicates that our model organizes objects by geometric structure without requiring class annotations.

\vspace{-15px}
\paragraph{How fast is our method?}
Our model is highly parallelizable, allowing multiple objects to be batched and processed simultaneously in a single forward pass.
On an RTX 4090 (24 GB), we can process up to 256 objects in parallel.
On average, the forward pass takes 0.13 s for a complete Replica~\cite{Straub2019TheRD} scene, 3D instance segmentation with Mask3D~\cite{Schult2022Mask3DMT} requires $0.3$ s, and each LM optimization step takes less than $1$ s, see supplementary for more details.


\vspace{-6px}
\section{Conclusion}
\vspace{-6px}
We proposed \name{}, a method for deriving compact yet expressive 3D scene representations based on simple geometric primitives -- specifically, superquadrics.
Our model outperforms prior primitive-based methods and generalizes well to out-of-category classes.
We further demonstrated the potential of the resulting 3D scene representation for various applications in robotics, and as a geometric prompt for diffusion-based image generation.
While this is only a first step towards more compact, geometry-aware 3D scene representations, we anticipate broader applications and expect to see further research in this direction.

\vspace{-12px}
\begin{small}
\paragraph{Acknowledgments.}
Elisabetta Fedele is supported by the ETH AI Center doctoral fellowship, by the Swiss National Science Foundation (SNSF) Advanced Grant 216260 (\textit{Beyond Frozen Worlds: Capturing Functional 3D Digital Twins from the Real World}), and an SNSF Mobility Grant. Francis Engelmann is supported by an SNSF PostDoc.mobilty grant. We also gratefully
acknowledge a GPU grant from NVIDIA.
\end{small}

{
    \small
    \bibliographystyle{ieeenat_fullname}
    \bibliography{main}

@String(BMVC    = {British Machine Vision Conference (BMVC)})

@String(CVPR    = {International Conference on Computer Vision and Pattern Recognition (CVPR)})

@String(ECCV    = {European Conference on Computer Vision (ECCV)})

@String(ICCV    = {International Conference on Computer Vision (ICCV)})

@String(NEURIPS = {International Conference on Neural Information Processing Systems (NeurIPS)})

@String(ICLR    = {International Conference on Learning Representations (ICLR)})

@String(ICRA    = {International Conference on Robotics and Automation (ICRA)})

@String(CORL    = {Conference on Robot Learning (CoRL)})

@String(TOG     = {ACM Transactions On Graphics (TOG)})

@String(AAAI    = {Association for the Advancement of Artificial Intelligence (AAAI)})

@String(PAMI    = {Transactions on Pattern Analysis and Machine Intelligence (PAMI)})

@String(RAL     = {IEEE Robotics and Automation Letters (RA-L)})

@inproceedings{schult2024controlroom3d,
  title={Controlroom3d: Room generation using semantic proxy rooms},
  author={Schult, Jonas and Tsai, Sam and H{\"o}llein, Lukas and Wu, Bichen and Wang, Jialiang and Ma, Chih-Yao and Li, Kunpeng and Wang, Xiaofang and Wimbauer, Felix and He, Zijian and others},
  booktitle=cvpr,
  year={2024}
}

@inproceedings{gu2024conceptgraphs,
  title={Conceptgraphs: Open-vocabulary 3d scene graphs for perception and planning},
  author={Gu, Qiao and Kuwajerwala, Ali and Morin, Sacha and Jatavallabhula, Krishna Murthy and Sen, Bipasha and Agarwal, Aditya and Rivera, Corban and Paul, William and Ellis, Kirsty and Chellappa, Rama and others},
  booktitle=icra,
  year={2024},
}

@article{ray2024task,
  title={Task and motion planning in hierarchical 3d scene graphs},
  author={Ray, Aaron and Bradley, Christopher and Carlone, Luca and Roy, Nicholas},
  journal={arXiv preprint arXiv:2403.08094},
  year={2024}
}

@inproceedings{li2024genzi,
  title={Genzi: Zero-shot 3d human-scene interaction generation},
  author={Li, Lei and Dai, Angela},
  booktitle=cvpr,
  year={2024}
}

@inproceedings{delitzas2024scenefun3d, 
  title = {{SceneFun3D: Fine-Grained Functionality and Affordance Understanding in 3D Scenes}}, 
  author = {Delitzas, Alexandros and Takmaz, Ayca and Tombari, Federico and Sumner, Robert and Pollefeys, Marc and Engelmann, Francis}, 
  booktitle = cvpr, 
  year = {2024}
}

@inproceedings{Paschalidou2019CVPR,
  title = {Superquadrics Revisited: Learning 3D Shape Parsing beyond Cuboids},
  author = {Paschalidou, Despoina and Ulusoy, Ali Osman and Geiger, Andreas},
  booktitle = cvpr,
  year = {2019},
}

@article{shriram2024realmdreamer,
  title={Realmdreamer: Text-driven 3d scene generation with inpainting and depth diffusion},
  author={Shriram, Jaidev and Trevithick, Alex and Liu, Lingjie and Ramamoorthi, Ravi},
  journal={arXiv preprint arXiv:2404.07199},
  year={2024}
}

@article{Yang2021UnsupervisedLF,
  title={Unsupervised learning for cuboid shape abstraction via joint segmentation from point clouds},
  author={Kaizhi Yang and Xuejin Chen},
  journal=tog,
  year={2021}
}

@article{Barr1981SuperquadricsAA,
  title={Superquadrics and Angle-Preserving Transformations},
  author={Alan H. Barr},
  journal={IEEE Computer Graphics and Applications},
  year={1981}
}

@inproceedings{alaniz2023iterative,
  title={Iterative superquadric recomposition of 3d objects from multiple viewsd},
  author={Alaniz, Stephan and Mancini, Massimiliano and Akata, Zeynep},
  booktitle=iccv,
  year={2023}
}

@article{Tulsiani2016LearningSA,
  title={Learning Shape Abstractions by Assembling Volumetric Primitives},
  author={Shubham Tulsiani and Hao Su and Leonidas J. Guibas and Alexei A. Efros and Jitendra Malik},
  journal=CVPR,
  year={2017}
}

@article{Solina1990RecoveryOP,
  title={Recovery of Parametric Models from Range Images: The Case for Superquadrics with Global Deformations},
  author={Franc Solina and Ruzena Bajcsy},
  journal=PAMI,
  year={1990}
}

@inproceedings{openscene,
  title={{OpenScene: 3D Scene Understanding with Open Vocabularies}},
  author={Peng, Songyou and Genova, Kyle and Jiang, Chiyu Max and Tagliasacchi, Andrea and Pollefeys, Marc and Funkhouser, Thomas},
  booktitle = CVPR,
  year={2023}
}

@inproceedings{takmaz2023openmask3d,
  title={{OpenMask3D: Open-Vocabulary 3D Instance Segmentation}},
  author={Takmaz, Ay{\c{c}}a and Fedele, Elisabetta and Sumner, Robert W. and Pollefeys, Marc and Tombari, Federico and Engelmann, Francis},
  booktitle=neurips,
year={2023}
}

@inproceedings{Cheng2021PerPixelCI,
  title={{Per-Pixel Classification is Not All You Need for Semantic Segmentation}},
  author={Bowen Cheng and Alexander G. Schwing and Alexander Kirillov},
  booktitle=neurips,
  year={2021},
}

@article{Cheng2021MaskedattentionMT,
  title={{Masked-attention Mask Transformer for Universal Image Segmentation}},
  author={Bowen Cheng and Ishan Misra and Alexander G. Schwing and Alexander Kirillov and Rohit Girdhar},
  journal=CVPR,
  year={2021}
}

@article{Schult2022Mask3DMT,
  title={Mask3D: Mask Transformer for 3D Semantic Instance Segmentation},
  author={Jonas Schult and Francis Engelmann and Alexander Hermans and Or Litany and Siyu Tang and Bastian Leibe},
  journal=ICRA,
  year={2022}
}

@article{kerbl3Dgaussians,
  title={3D Gaussian Splatting for Real-Time Radiance Field Rendering},
  author={Bernhard Kerbl and Georgios Kopanas and Thomas Leimkuehler and George Drettakis},
  journal=TOG,
  year={2023}
}

@article{Yeshwanth2023ScanNetAH,
  title={ScanNet++: A High-Fidelity Dataset of 3D Indoor Scenes},
  author={Chandan Yeshwanth and Yueh-Cheng Liu and Matthias Nie{\ss}ner and Angela Dai},
  journal=ICCV,
  year={2023}
}

@article{Dop1998FittingUS,
  title={Fitting undeformed superquadrics to range data: improving model recovery and classification},
  author={Erik Roeland van Dop and Paulus P.L. Regtien},
  journal=CVPR,
  year={1998}
}

@inproceedings{mildenhall2020nerf,
 title={NeRF: Representing Scenes as Neural Radiance Fields for View Synthesis},
 author={Ben Mildenhall and Pratul P. Srinivasan and Matthew Tancik and Jonathan T. Barron and Ravi Ramamoorthi and Ren Ng},
 year={2020},
 booktitle=ECCV,
}

@article{Monnier2023DifferentiableBW,
  title={Differentiable Blocks World: Qualitative 3D Decomposition by Rendering Primitives},
  author={Tom Monnier and Jake Austin and Angjoo Kanazawa and Alexei A. Efros and Mathieu Aubry},
  journal=neurips,
  year={2023}
}

@article{Liu2019SoftRA,
  title={Soft Rasterizer: A Differentiable Renderer for Image-Based 3D Reasoning},
  author={Shichen Liu and Tianye Li and Weikai Chen and Hao Li},
  journal=ICCV,
  year={2019}
}

@article{Jensen2014LargeSM,
  title={Large Scale Multi-view Stereopsis Evaluation},
  author={Rasmus Ramsb{\o}l Jensen and A. Dahl and George Vogiatzis and Engil Tola and Henrik Aan{\ae}s},
  journal=CVPR,
  year={2014}
}

@inproceedings{Chevalier2003SegmentationAS,
  title={Segmentation and Superquadric Modeling of 3D Objects},
  author={Laurent Chevalier and Fabrice Jaillet and Atilla Baskurt},
  booktitle={International Conference in Central Europe on Computer Graphics and Visualization},
  year={2003}
}

@article{Straub2019TheRD,
  title={The Replica Dataset: A Digital Replica of Indoor Spaces},
  author={Julian Straub and Thomas Whelan and Lingni Ma and Yufan Chen and Erik Wijmans and Simon Green and Jakob J. Engel and Raul Mur-Artal and Carl Yuheng Ren and Shobhit Verma and Anton Clarkson and Ming Yan and Brian Budge and Yajie Yan and Xiaqing Pan and June Yon and Yuyang Zou and Kimberly Leon and Nigel Carter and Jesus Briales and Tyler Gillingham and Elias Mueggler and Luis Pesqueira and Manolis Savva and Dhruv Batra and Hauke Malte Strasdat and Renzo De Nardi and Michael Goesele and S. Lovegrove and Richard A. Newcombe},
  journal={ArXiv},
  year={2019}
}

@techreport{shapenet2015,
  title       = {{ShapeNet: An Information-Rich 3D Model Repository}},
  author      = {Chang, Angel X. and Funkhouser, Thomas and Guibas, Leonidas and Hanrahan, Pat and Huang, Qixing and Li, Zimo and Savarese, Silvio and Savva, Manolis and Song, Shuran and Su, Hao and Xiao, Jianxiong and Yi, Li and Yu, Fisher},
    journal={ArXiv},
  year        = {2015}
}

@book{jaklic2000segmentation,
  title={Segmentation and recovery of superquadrics},
  author={Jaklic, Ales and Leonardis, Ales and Solina, Franc},
  year={2000},
  publisher={Springer Science \& Business Media}
}

@inproceedings{choy20163d,
  title={{3D-R2N2: A Unified Approach for Single and Multi-view 3D Object Reconstruction}},
  author={Choy, Christopher B and Xu, Danfei and Gwak, JunYoung and Chen, Kevin and Savarese, Silvio},
  booktitle = ECCV,
  year={2016}
}

@inproceedings{liu2022robust,
  title={{Robust and Accurate Superquadric Recovery: A Probabilistic Approach}},
  author={Liu, Weixiao and Wu, Yuwei and Ruan, Sipu and Chirikjian, Gregory S},
  booktitle=cvpr,
  year={2022}
}

@article{Held20243DConvex,
title = {{3D} Convex Splatting: Radiance Field Rendering with {3D} Smooth Convexes},
author = {Held, Jan and Vandeghen, Renaud and Hamdi, Abdullah and Deli{`e}ge, Adrien and Cioppa, Anthony and Giancola, Silvio and Vedaldi, Andrea and Ghanem, Bernard and Van Droogenbroeck, Marc},
journal = {ArXiv},
year = {2024}
}

@inproceedings{hamdi2024ges,
  title={{Ges: Generalized exponential splatting for efficient radiance field rendering}},
  author={Hamdi, Abdullah and Melas-Kyriazi, Luke and Mai, Jinjie and Qian, Guocheng and Liu, Ruoshi and Vondrick, Carl and Ghanem, Bernard and Vedaldi, Andrea},
  booktitle=CVPR,
  year={2024}
}

@inproceedings{ramamonjisoa2022monteboxfinder,
  title={{MonteBoxFinder: Detecting and Filtering Primitives to Fit a Noisy Point Cloud}},
  author={Ramamonjisoa, Micha{\"e}l and Stekovic, Sinisa and Lepetit, Vincent},
  booktitle=eccv,
  year={2022},
}

@inproceedings{liu2019pvcnn,
  title={{Point-Voxel CNN for Efficient 3D Deep Learning}},
  author={Liu, Zhijian and Tang, Haotian and Lin, Yujun and Han, Song},
  booktitle=neurips,
  year={2019}
}

@inproceedings{vezzani2017grasping,
  title={A grasping approach based on superquadric models},
  author={Vezzani, Giulia and Pattacini, Ugo and Natale, Lorenzo},
  booktitle={2017 IEEE International Conference on Robotics and Automation (ICRA)},
  pages={1579--1586},
  year={2017},
  organization={IEEE}
}

@inproceedings{Vaswani2017AttentionIA,
  title={{Attention is All you Need}},
  author={Ashish Vaswani and Noam M. Shazeer and Niki Parmar and Jakob Uszkoreit and Llion Jones and Aidan N. Gomez and Lukasz Kaiser and Illia Polosukhin},
  booktitle=neurips,
  year={2017}
}

@inproceedings{pilu1995equal,
  title={{Equal-distance sampling of superellipse models}},
  author={Pilu, Maurizio and Fisher, Robert B},
  booktitle=BMVC,
  year={1995}
}

@article{Levenberg1944AMF,
  title={{A method for the solution of certain non-linear problems in least squares.}},
  author={Kenneth Levenberg},
  journal={Quarterly of Applied Mathematics},
  year={1944}
}

@article{huang2025vipscene,
  title={Video Perception Models for 3D Scene Synthesis},
  author={Huang, Rui and Zhai, Guangyao and Bauer, Zuria and Pollefeys, Marc and Tombari, Federico and Guibas, Leonidas and Huang, Gao and Engelmann, Francis},
  journal={arXiv preprint arXiv:2506.20601},
  year={2025}
}

@article{Marquardt1963AnAF,
  title={An Algorithm for Least-Squares Estimation of Nonlinear Parameters},
  author={Donald W. Marquardt},
  journal={Journal of The Society for Industrial and Applied Mathematics},
  year={1963}
}

@article{loshchilov2017decoupled,
  title={Decoupled weight decay regularization},
  author={Loshchilov, Ilya and Hutter, Frank},
  journal=iclr,
  year=2019
}

@article{takmaz2025search3d,
  title={{Search3D: Hierarchical Open-Vocabulary 3D Segmentation}},
  author={Takmaz, Ayca and Delitzas, Alexandros and Sumner, Robert W. and Engelmann, Francis and Wald, Johanna and Tombari, Federico},
  journal=RAL,
  year={2025}
}

@article{qi2017pointnetplusplus,
  title     = {PointNet++: Deep Hierarchical Feature Learning on Point Sets in a Metric Space},
  author    = {Qi, Charles R and Yi, Li and Su, Hao and Guibas, Leonidas J},
  journal   = neurips,
  year      = {2017}
}

@article{Zhang2023AddingCC,
  title={Adding Conditional Control to Text-to-Image Diffusion Models},
  author={Lvmin Zhang and Anyi Rao and Maneesh Agrawala},
  journal=ICCV,
  year={2023}
}

@article{rombach2021highresolution,
      title={High-Resolution Image Synthesis with Latent Diffusion Models}, 
      author={Robin Rombach and Andreas Blattmann and Dominik Lorenz and Patrick Esser and Björn Ommer},
      year={2022},
      journal=CVPR,
}

@inproceedings{cai2022volumetric,
  title     = {Volumetric-based Contact Point Detection for 7-DoF Grasping},
  author    = {Cai, Junhao and Su, Jingcheng and Zhou, Zida and Cheng, Hui and Chen, Qifeng and Wang, Michael Yu},
  booktitle = corl,
  year      = {2022},
}

@inproceedings{1241860,
  title     = {{Automatic Grasp Planning Using Shape Primitives}}, 
  author    = {Miller, A.T. and Knoop, S. and Christensen, H.I. and Allen, P.K.},
  booktitle = icra, 
  year      = {2003}}

@inproceedings{131614,
  title     = {{On Computing Two-finger Force-closure Grasps of Curved 2D Objects}}, 
  author    = {Faverjon, B. and Ponce, J.},
  booktitle = icra, 
  year      = {1991}}

@inproceedings{wang2021graspness,
  title     = {{Graspness Discovery in Clutters for Fast and Accurate Grasp Detection}},
  author    = {Wang, Chenxi and Fang, Hao-Shu and Gou, Minghao and Fang, Hongjie and Gao, Jin and Lu, Cewu},
  booktitle = cvpr,
  year      = {2021}
}

@article{mahler2019learning,
  title     = {{Learning ambidextrous robot grasping policies}},
  author    = {Mahler, Jeffrey and Matl, Matthew and Satish, Vishal and Danielczuk, Michael and DeRose, Bill and McKinley, Stephen and Goldberg, Ken},
  journal   = {Science Robotics},
  year      = {2019},
}

@inproceedings{sundermeyer2021contact,
  title={{Contact-graspnet: Efficient 6-dof grasp generation in cluttered scenes}},
  author={Sundermeyer, Martin and Mousavian, Arsalan and Triebel, Rudolph and Fox, Dieter},
  booktitle=ICRA,
  year={2021}
}

@article{tu2024superq,
  title={SuperQ-GRASP: Superquadrics-based Grasp Pose Estimation on Larger Objects for Mobile-Manipulation},
  author={Tu, Xun and Desingh, Karthik},
  journal={arXiv},
  year={2024}
}

@inproceedings{liu2023marching,
  title={{Marching-primitives: Shape abstraction from signed distance function}},
  author={Liu, Weixiao and Wu, Yuwei and Ruan, Sipu and Chirikjian, Gregory S},
  booktitle=CVPR,
  year={2023}
}

@article{fang2023anygrasp,
  title={{Anygrasp: Robust and efficient grasp perception in spatial and temporal domains}},
  author={Fang, Hao-Shu and Wang, Chenxi and Fang, Hongjie and Gou, Minghao and Liu, Jirong and Yan, Hengxu and Liu, Wenhai and Xie, Yichen and Lu, Cewu},
  journal={IEEE Transactions on Robotics},
  year={2023}
}

@inproceedings{sudry2023hierarchical,
  title={Hierarchical planning for rope manipulation using knot theory and a learned inverse model},
  author={Sudry, Matan and Jurgenson, Tom and Tamar, Aviv and Karpas, Erez},
  booktitle={Conference on Robot Learning},
  year={2023}
}

@inproceedings{devlin2019bert,
  title={{Bert: Pre-training of deep bidirectional transformers for language understanding}},
  author={Devlin, Jacob and Chang, Ming-Wei and Lee, Kenton and Toutanova, Kristina},
  booktitle={{Proceedings of the 2019 conference of the North American chapter of the association for computational linguistics: human language technologies.}},
  year={2019}
}

@article{van2008visualizing,
  title={{Visualizing data using t-SNE.}},
  author={Van der Maaten, Laurens and Hinton, Geoffrey},
  journal={Journal of machine learning research},
  year={2008}
}

@inproceedings{pentland1986parts,
  title={Parts: structured descriptions of shape},
  author={Pentland, Alex P},
  booktitle={{Proceedings of the Fifth AAAI National Conference on Artificial Intelligence}},
  year={1986}
}

@inproceedings{zhang2025open,
  title={{Open-Vocabulary Functional 3D Scene Graphs for Real-World Indoor Spaces}},
  author={Zhang, Chenyangguang and Delitzas, Alexandros and Wang, Fangjinhua and Zhang, Ruida and Ji, Xiangyang and Pollefeys, Marc and Engelmann, Francis},
  booktitle=cvpr,
  year={2025}
}

@inproceedings{koch2024open3dsg,
  title={{Open3DSG: Open-vocabulary 3D Scene Graphs from Point Clouds with Queryable Objects and Open-set Relationships}},
  author={Koch, Sebastian and Vaskevicius, Narunas and Colosi, Mirco and Hermosilla, Pedro and Ropinski, Timo},
  booktitle=cvpr,
  year={2024}
}
}

\maketitlesupplementary
\setcounter{page}{1}
\setcounter{section}{0}
\section{Superquadrics}
While our architecture can be easily adapted to segment and predict in an unsupervised manner other types of geometric primitives - in \name{} we decided to use superquadrics. When looking for a suitable geometric primitive for our approach we were keeping in mind two main criteria. First, we wanted the primitive to be represented by a compact parameterization so that it can be described by only using a few parameters. Second, we wanted the representation to be expressive, in order to be able to describe real-world objects by only using a few primitives. Inspired by 3DGS~\cite{kerbl3Dgaussians}, the first parameterization we took into consideration were the \textit{ellipsoids}. Ellipsoids have a very compact parameterization as their shape can be represented using the following implicit equation:
\begin{align*}
    f(\vx)&=\left(\frac{x}{\scalex}\right)^{2}+\left(\frac{y}{\scaley}\right)^{2}+\left(\frac{z}{\scalez}\right)^{2}=1,
\end{align*}
where the only free variables are $\scalex$, $\scaley$, $\scalez$, which are the lengths of the three main semi-axis.
However, if we start thinking about which objects and object parts can be effectively fitted using a single ellipsoid, we realize that their representational capabilities are not enough. In order to obtain higher representational capabilities while still keeping a simple representation, a natural extension are \textit{generalized ellipsoids}. In this representation, we not only allow the length of the semi-axis to be variable, but their roundness controlled by the three exponents, which previously were fixed to $2$. In that way, we obtain the following implicit function:
\begin{align*}
    f(\vx)&=\left(\frac{|x|}{\scalex}\right)^{e_1}+\left(\frac{|y|}{\scaley}\right)^{e_2}+\left(\frac{|z|}{\scalez}\right)^{e_3}=1 \ .
\end{align*}
Using generalized ellipsoids with high exponents it becomes possible to also represent cuboidal shapes. While having suitable representational capabilities, these primitives do not allow to compute distance to their surface in a closed form, a property which can be extremely useful for various downstream applications. This drawback is overcome by \textit{superquadrics}, at the cost of one less degree of freedom, which however does not substantially impact expressivity. Unlike generalized ellipsoids, which assign a separate roundness parameter to each axis, superquadrics share the same roundness for the $x$ and $y$ axes while allowing a distinct parameter for the $z$ axis. Their shape is represented in implicit form by the equation:
\begin{equation*}
    f(\vx)= \left(  \left(\frac{x}{\scalex}\right)^{\frac{2}{\epsilon_2}} +\left(\frac{y}{\scaley}\right)^{\frac{2}{\epsilon_2}}\right)^\frac{\epsilon_2}{\epsilon_1}  +\left(\frac{z}{\scalez}\right)^{\frac{2}{\epsilon_1}}=1\ ,
\end{equation*}
and the euclidean radial distance to their surface can be computed in closed form, as shown in Eq.~\ref{eq:rad-dist}. In addition, superquadrics are also equipped with an explicit function, which can be used to sample points from their surface. Specifically, given the coordinates $(\eta,\omega)$ such that $\eta\in\left[-\frac{\pi}{2}, \frac{\pi}{2}\right]$ and $\omega\in\left[-\pi, \pi\right]$, the surface of a superquadric can be represented as:
\begin{align*}
   s(\eta,\omega)
    &=
    \begin{bmatrix}
        \scalex\cos{\eta}^{\epsilon_1}\cos\omega^{\epsilon_2}\\
        \scaley\cos{\eta}^{\epsilon_1}\sin\omega^{\epsilon_2}\\
        \scalez\sin{\eta}^{\epsilon_1}
    \end{bmatrix}.
\end{align*}
We refer to~\cite{jaklic2000segmentation} for additional details on superquadrics.
\section{Training Details}
In general, we train our model for $500$ epochs with $\lambda_{par}=0.1$ and then for other $500$ epochs with $\lambda_{par}=0.6$. For the results on Replica~\cite{Straub2019TheRD} and ScanNet++~\cite{Yeshwanth2023ScanNetAH} we trained our model on normalized objects (centered and rescaled to fit in a sphere of radius $0.5$). During training we apply rotation and translation augmentations. Specifically, we apply random rotations between $0^\circ$ and $180^\circ$ around the z axis and between $0^\circ$ and $7.5^\circ$ on the x and y axis. We also apply random translation with respect to the center in a radius of $0.05$. We optimize our network with Adam~\cite{loshchilov2017decoupled} a one-cycle learning rate schedule with a maximum learning rate of $4e-4$. We train on $4$ NVIDIA A100 with total batch size of $128$.

\section{Additional Results}

\begin{figure}
    \centering
    \includegraphics[width=0.75\linewidth]{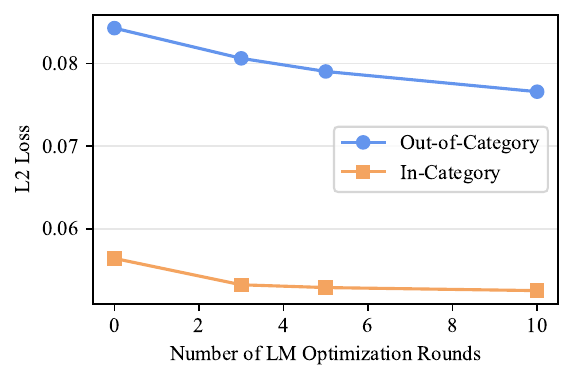}
    \caption{\textbf{LM optimization experiment.} We show how LM optimization improves results in terms of L2 Chamfer distance across a variable number of rounds. We report results both for in-category experiments and out-of-category ones.}
    \label{fig:lm}
\end{figure}

\paragraph{Does LM improve our final predictions?}
In our approach we use LM optimization as a post processing step. In this experiment (Fig.~\ref{fig:lm}) we want to assess how a different number of LM optimization rounds affects the final predictions in terms of L2 Chamfer Distance. In order to evaluate this aspect, we report L2 loss after different numbers of LM optimization steps, evaluating both in-category and out-of-category. From this experiment we can notice two main aspects. Firstly, we see that it leads to larger improvements in the out-of-category rather than in the in-category one. This is probably due to the less accurate initial predictions of our feedforward model in this setting and it shows that our optimization step can be used to decrease the gap between in-category and out-category. Secondly, we see that even if LM optimization improves our final predictions, it does not lead to substantial improvements. This suggests that the solutions predicted by our method are located in local minima and that a diverse type of optimization should be resorted to improve the predictions further.
\vspace{-10px}
\paragraph{Is the existence head needed?}
As explained in the main paper, \name{} predicts a set of $P$ superquadrics and a segmentation matrix. As in many cases less than $P$ superquadrics are in fact needed to represent the input point cloud, we also predict an existence parameter ($\alpha_j$) for each of them, which allows us to model a variable number of primitives. However, the existence of a superquadric can also be directly deducted from the segmentation matrix, by computing $\hat{\alpha_j}$. We made some experiments to assess the impact of using either $\alpha_j$ or $\hat{\alpha_j}$, both at training and at inference time and reported the results in Tab.~\ref{tab:ablation-existence}. In order to evaluate the impact of the prediction head, we conducted two additional experiments. In the first experiment, we evaluate the model trained as explained in Sec.~\ref{sec:object-level-method}. In the second experiment, we retrained our model without explicitly supervising $\alpha_j$ and using $\hat{\alpha_j}$ at inference time. 

\begin{table}[t]
\centering

\setlength{\tabcolsep}{7pt}
\resizebox{\columnwidth}{!}{
\begin{tabular}{l cc ccc }
\toprule
\textbf{Method}&\textbf{Training}&\textbf{Inference}&\textbf{L1} $\downarrow$ &
\textbf{L2} $\downarrow$ & 
\textbf{\#\,Prim.}$\downarrow$ \\
\midrule
\textcolor{black!30}{SQ}& \textcolor{black!30}{-}  & \textcolor{black!30}{-}  & \textcolor{black!30}{$3.668$}& \textcolor{black!30}{$0.279$} & \textcolor{black!30}{$10$}\\
\arrayrulecolor{black!10}\midrule\arrayrulecolor{black}
\name{} (Ours) &\cmark & \cmark& $\mathbf{1.698}$&  $0.047$ & $\mathbf{5.79}$  \\ 
\name{} (Ours) &\cmark & \xmark&$\mathbf{1.695}$&  $\mathbf{0.046}$ & $5.82$  \\ 
\name{} (Ours) &\xmark & \xmark&$1.711$&  $0.049$ & $5.84$  \\ 
\bottomrule
\end{tabular}
}
\caption{\textbf{Ablation study on existence head.} Evaluation on objects from ShapeNet~\cite{shapenet2015} dataset. The first two columns indicate whether we use the predicted $\alpha_j$ at training (left) and inference (right) time. Scores are scaled by $10^2$.} 
\label{tab:ablation-existence}
\end{table}

\begin{figure}[t]
    \centering
    \includegraphics[width=0.7\linewidth, trim={18px 20px 18px 5px}]{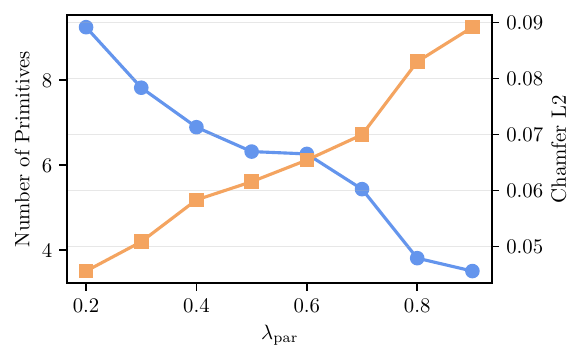}%
    \vspace{0.2cm}
    \caption{\textbf{Compactness vs. reconstruction accuracy tradeoff.} We run experiment for different values of the parsimony weight $\wpar$ (x-axis) and we visualize the {\color{Blue}resulting number of primitives} (y-axis, \textit{left}) and the {\color{BurntOrange}L2 Chamfer distance} (y-axis, \textit{right}).}
    \label{fig:tradeoff}
\end{figure}

\paragraph{Compactness–Accuracy Trade-off.}
The hyperparameter $\wpar$ controls the trade-off between reconstruction accuracy and representation compactness (see Eq.~\ref{eq:loss}).
We evaluate this trade-off quantitatively by first training the model with $\wpar=0.1$ for 500 epochs, followed by fine-tuning for 100 epochs with varying $\wpar$ values.
Fig.~\ref{fig:tradeoff} shows the impact of $\wpar$ on Chamfer distance and the average number of predicted primitives.
By adjusting $\wpar$, the model can smoothly balance compactness and accuracy,
allowing for easy fine-tuning to meet target reconstruction quality.
In our experiments, we use $\wpar=0.6$, which approximately corresponds to the intersection point of the two curves.

\vspace{-10px}
\paragraph{Is FPS needed?}
To understand whether point clouds need to be downsampled with Farthest Point Sampling (FPS) or points can just be randomly sampled, we compare the performances of \name{} on 3D scene datasets using the two approaches. We observe that the performances do not degenerate and that random sampling is enough for practical applications.

\begin{table}[t]
\centering

\setlength{\tabcolsep}{7pt}
\resizebox{\columnwidth}{!}{
\begin{tabular}{l cccc|ccc }
\toprule
&  &\multicolumn{3}{c}{\textbf{ScanNet++}~\cite{Yeshwanth2023ScanNetAH}} & \multicolumn{3}{c}{\textbf{Replica}~\cite{Straub2019TheRD}} \\
\midrule 
\textbf{Method}&\textbf{FPS}& \textbf{L1} $\downarrow$ &
\textbf{L2} $\downarrow$ & 
\textbf{\#\,Prim.}$\downarrow$ 
&\textbf{L1} $\downarrow$ &
\textbf{L2} $\downarrow$ & 
\textbf{\#\,Prim.}$\downarrow$\\
\midrule
\textcolor{black!30}{CSA}&\textcolor{black!30}{\cmark}  & \textcolor{black!30}{$2.91$}& \textcolor{black!30}{$0.41$} & \textcolor{black!30}{$11.64$} & \textcolor{black!30}{$3.68$}& \textcolor{black!30}{$0.70$} & \textcolor{black!30}{$9.63$}\\
\arrayrulecolor{black!10}\midrule\arrayrulecolor{black}
\name{} (Ours)&\cmark  & $\mathbf{1.70}$&  $\mathbf{0.11}$ & $\mathbf{5.18}$ &$\mathbf{1.79}$& $\mathbf{0.19}$ & $6.58$\\ 
\name{} (Ours)&\xmark  & $1.74$&  $0.13$ & $5.19$ &$1.81$& $0.20$ & $\mathbf{6.52}$\\ 
\bottomrule
\end{tabular}
}
\caption{\textbf{Ablation study on sampling technique.} Evaluation on objects from 3D scene datasets~\cite{Straub2019TheRD,Yeshwanth2023ScanNetAH}. The second column report whether Farthest Point Sampling (FPS) was used to sample the input points. Alternatively, points were randomly sampled. Scores are scaled by $10^2$.} 
\label{tab:eval-fps}
\end{table}

\section{Qualitative results on scenes}
\subsection{Replica}
\begin{figure*}
\centering
\begin{small}
\begin{tabular}{ccccc}
\hspace{0.01\linewidth} &\hspace{0.215\linewidth} & \hspace{0.215\linewidth} & \hspace{0.215\linewidth} & \hspace{0.215\linewidth} \\
&\textbf{Office 4} &\textbf{Room 0} &\textbf{Room 1} &\textbf{Room 2}
\end{tabular}\\
\vspace{1px}
\rotatebox{90}{\hspace{20px}\textbf{Point Cloud}}
\hspace{2px}
\includegraphics[width=0.238\linewidth,,clip]
{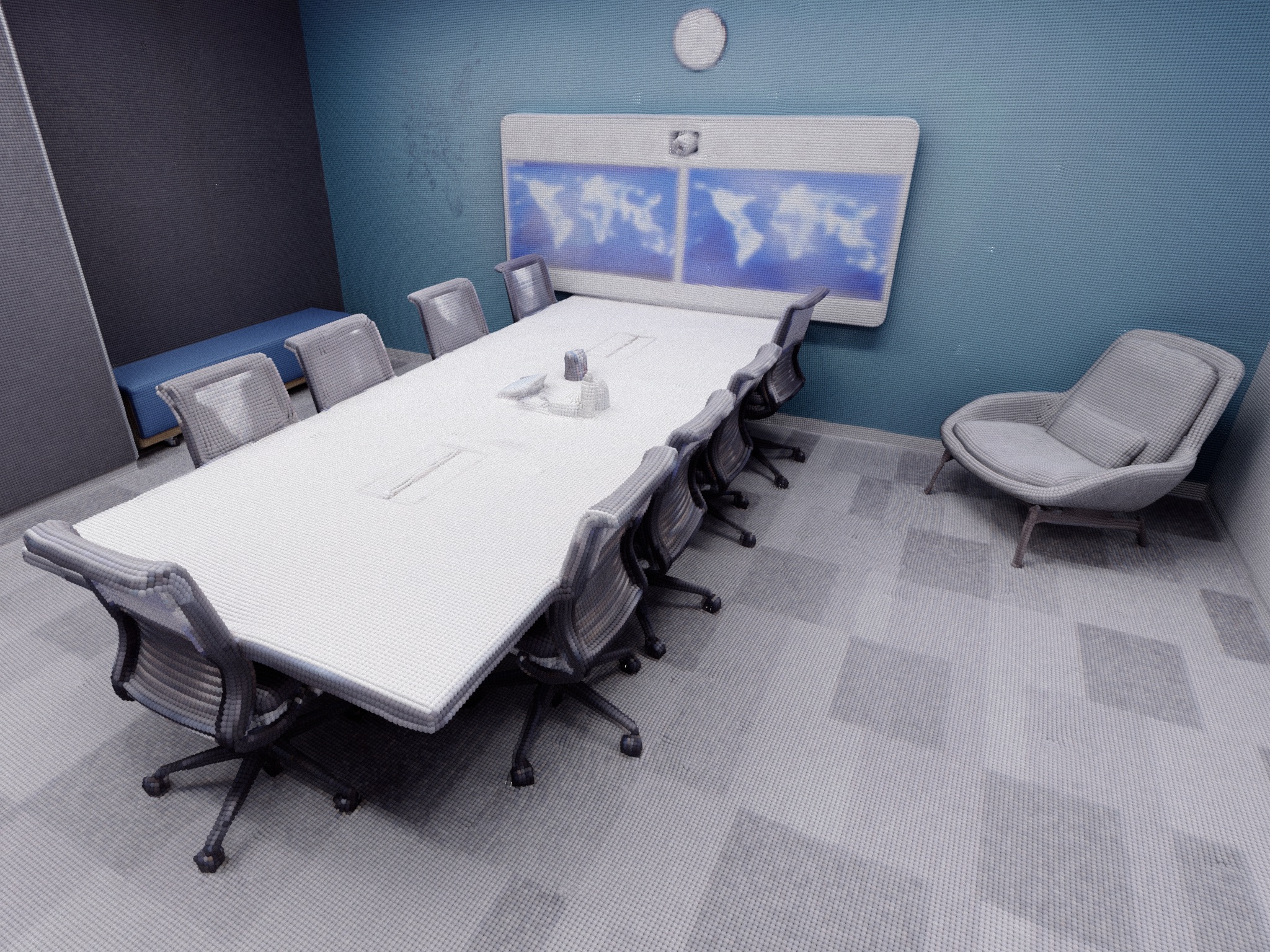}
\includegraphics[width=0.238\linewidth,clip]{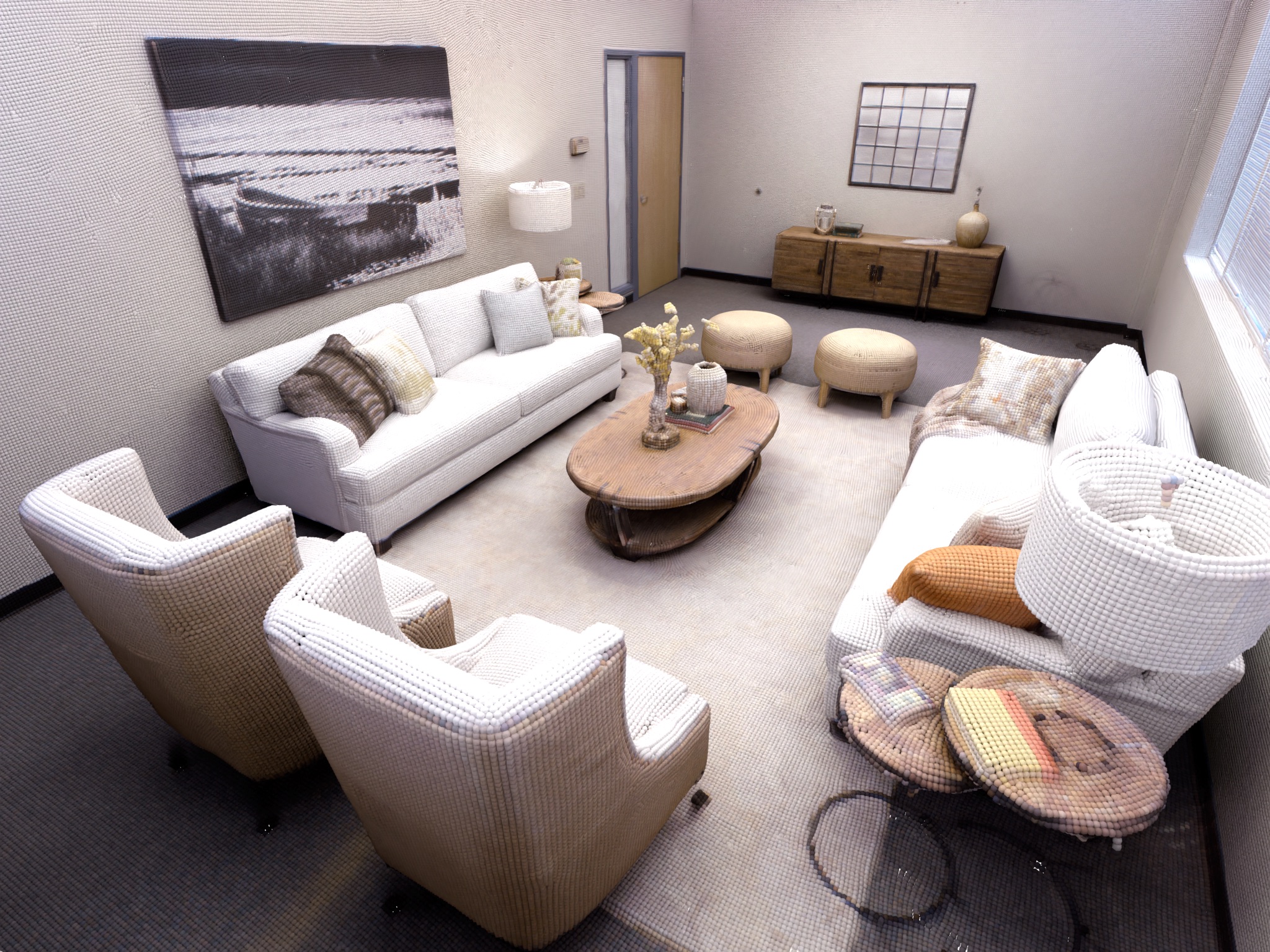}
\includegraphics[width=0.238\linewidth,,clip]
{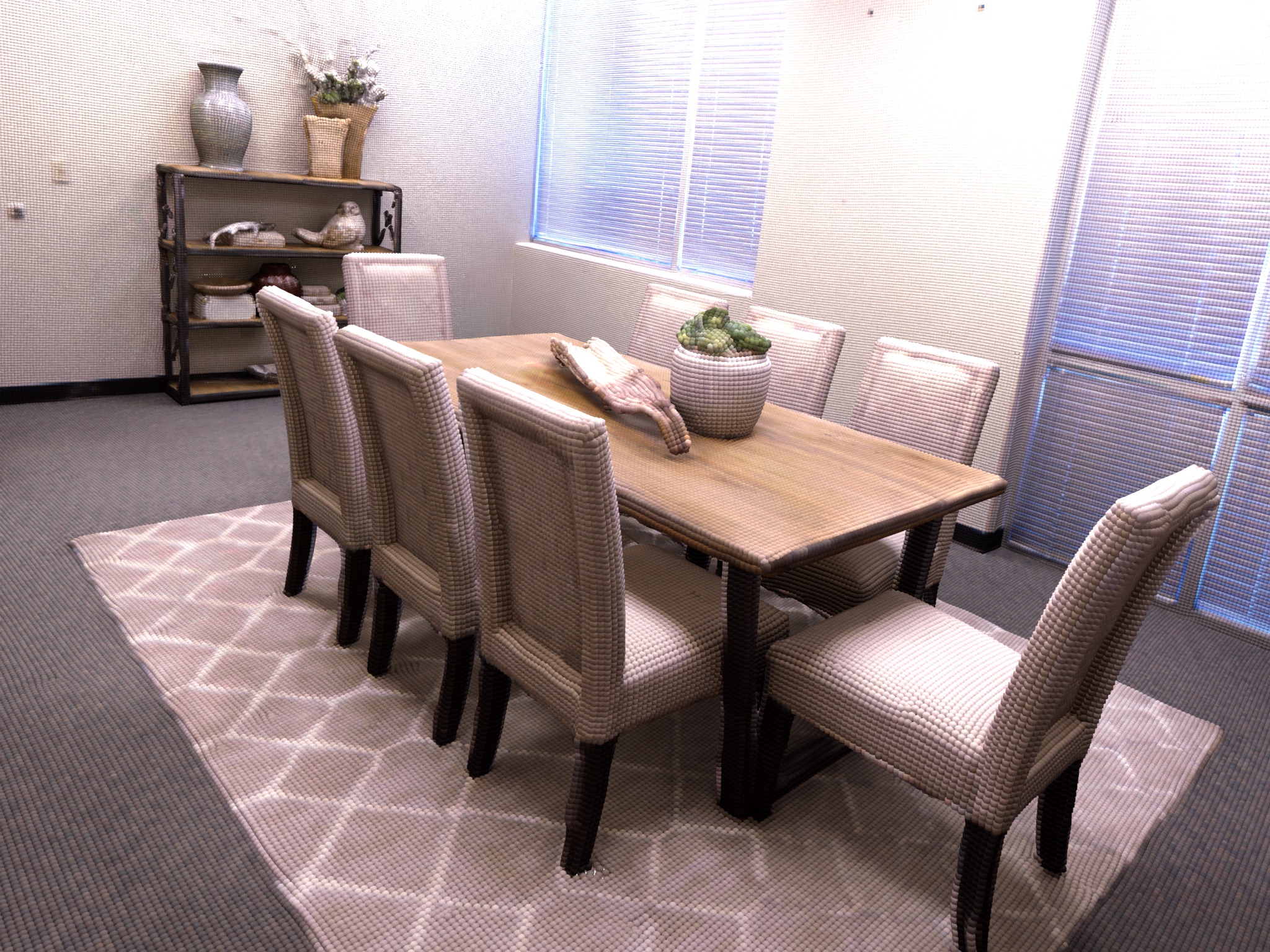}
\includegraphics[width=0.238\linewidth,clip]{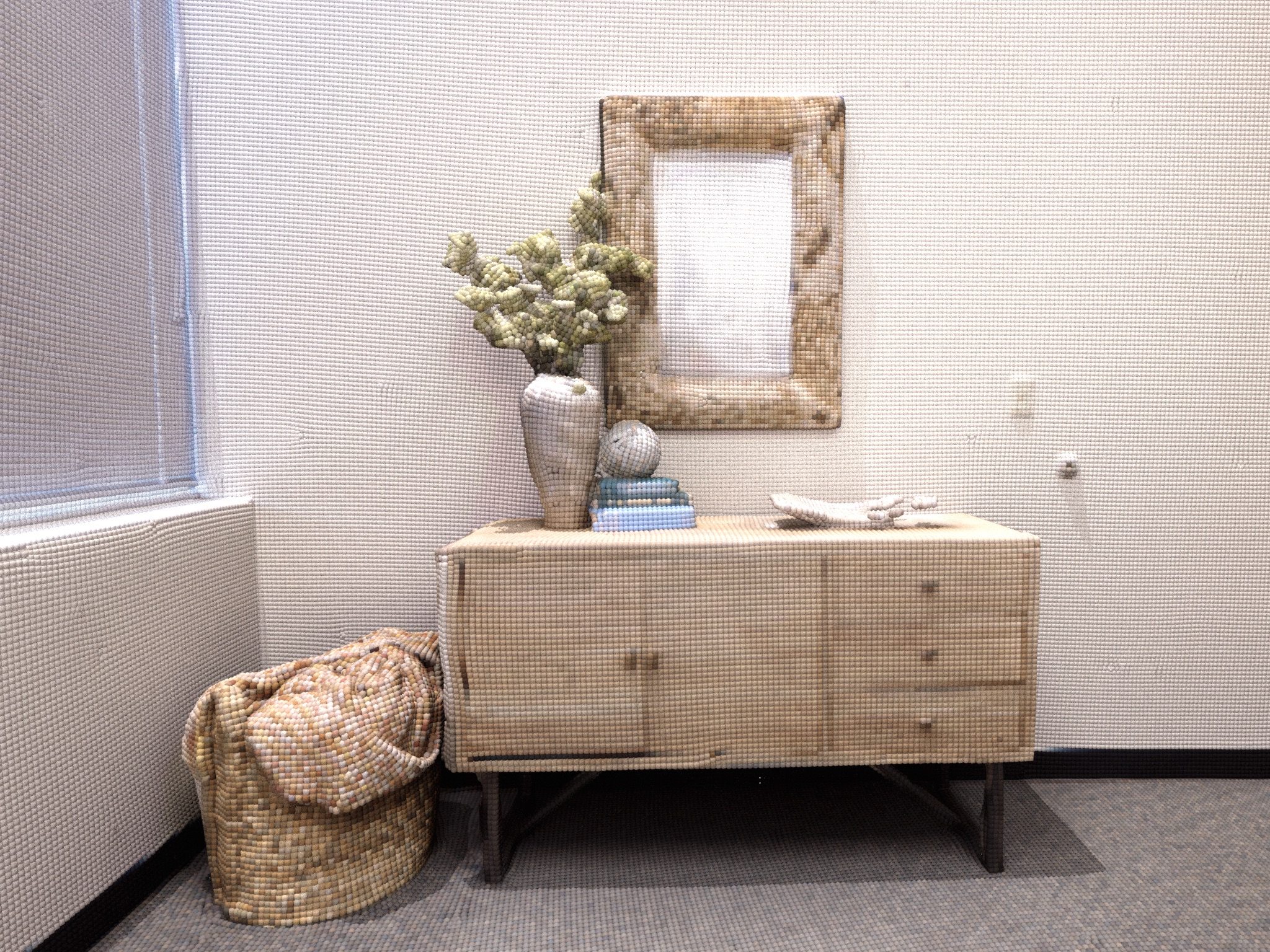}
\vspace{2px}\\

\rotatebox{90}{\hspace{16px}\textbf{Superquadrics}}
\hspace{1.3px}
\includegraphics[width=0.238\linewidth,,clip]{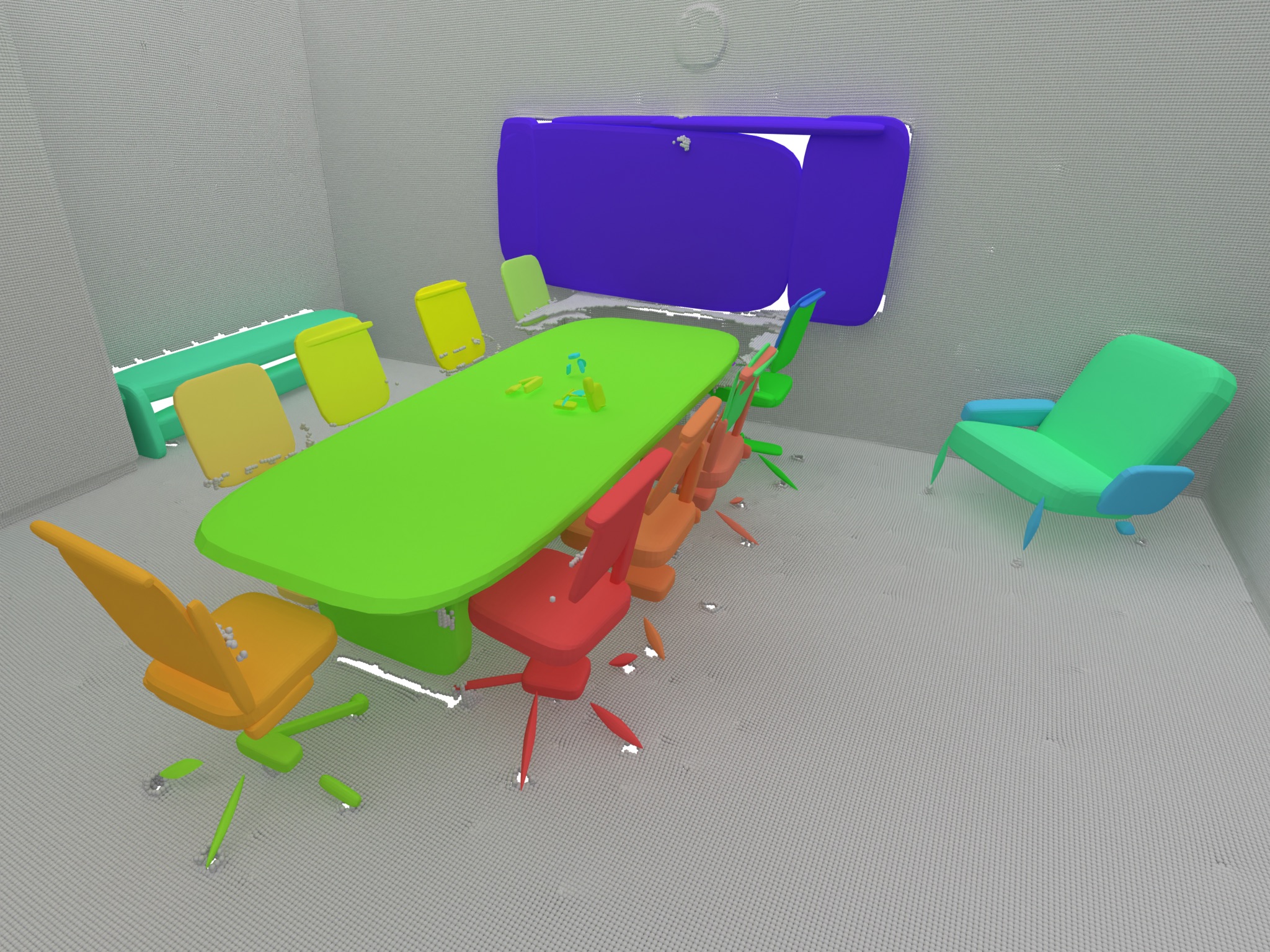}
\includegraphics[width=0.238\linewidth,clip]{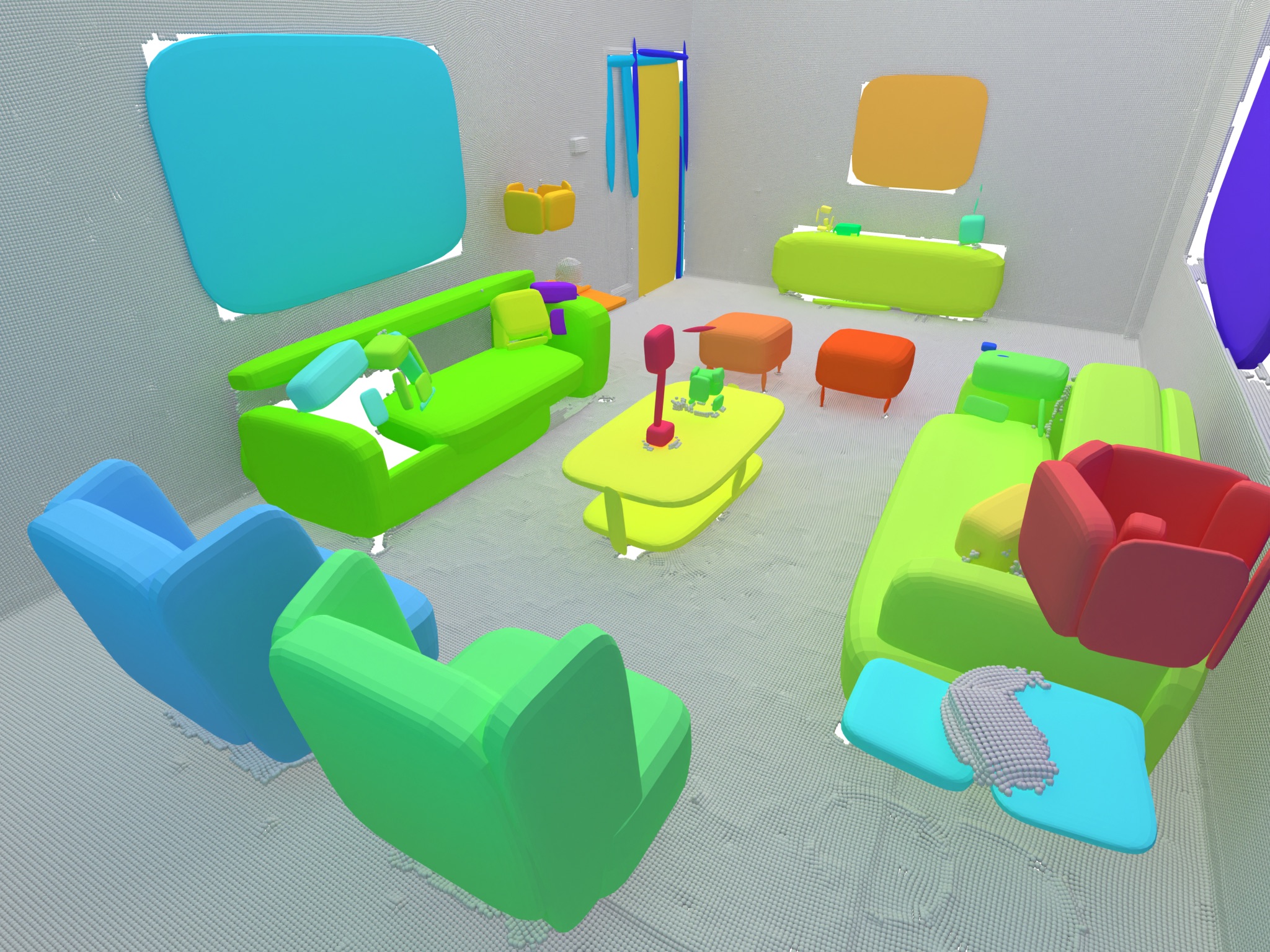}
\includegraphics[width=0.238\linewidth,,clip]{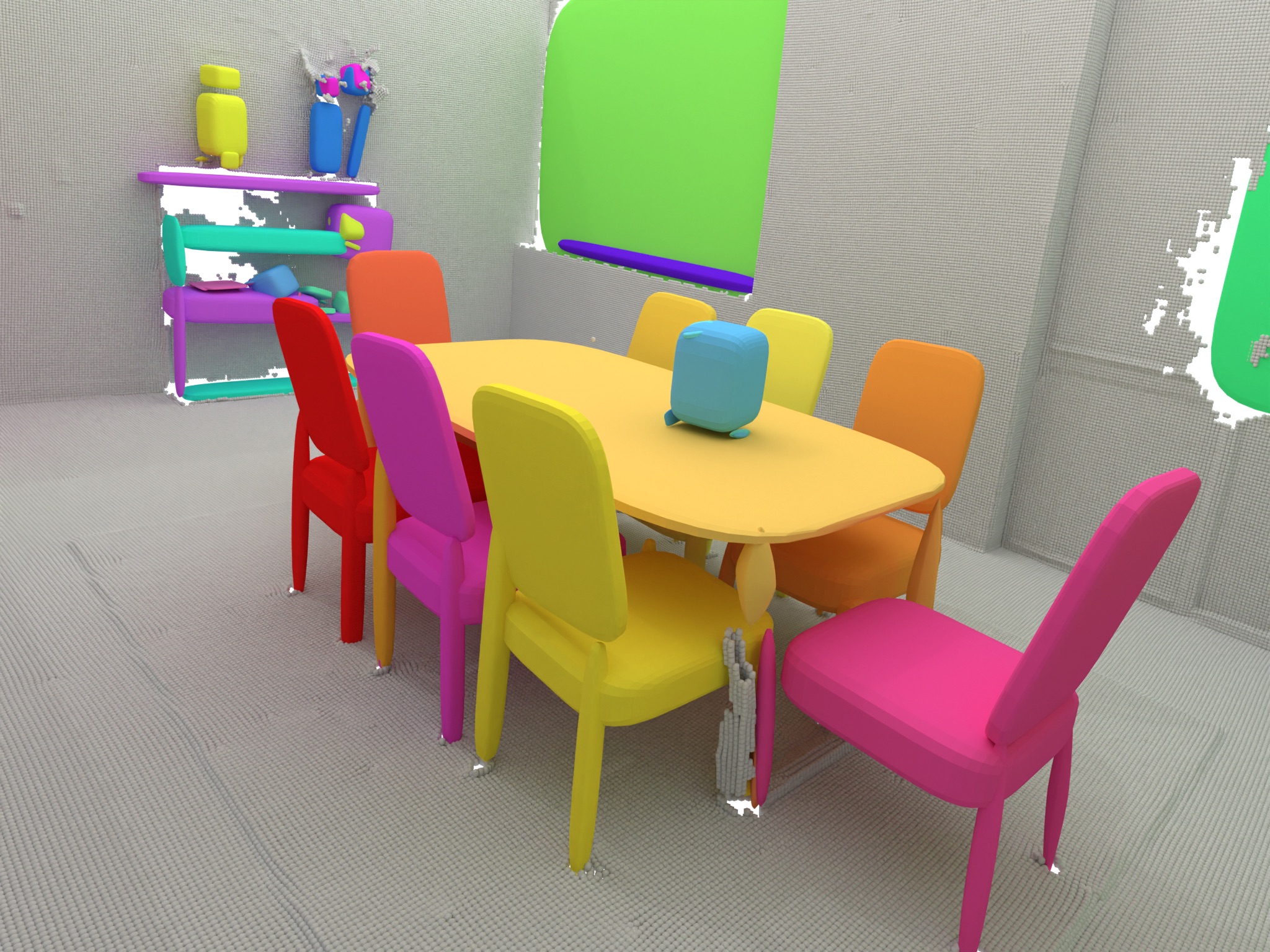}
\includegraphics[width=0.238\linewidth,clip]{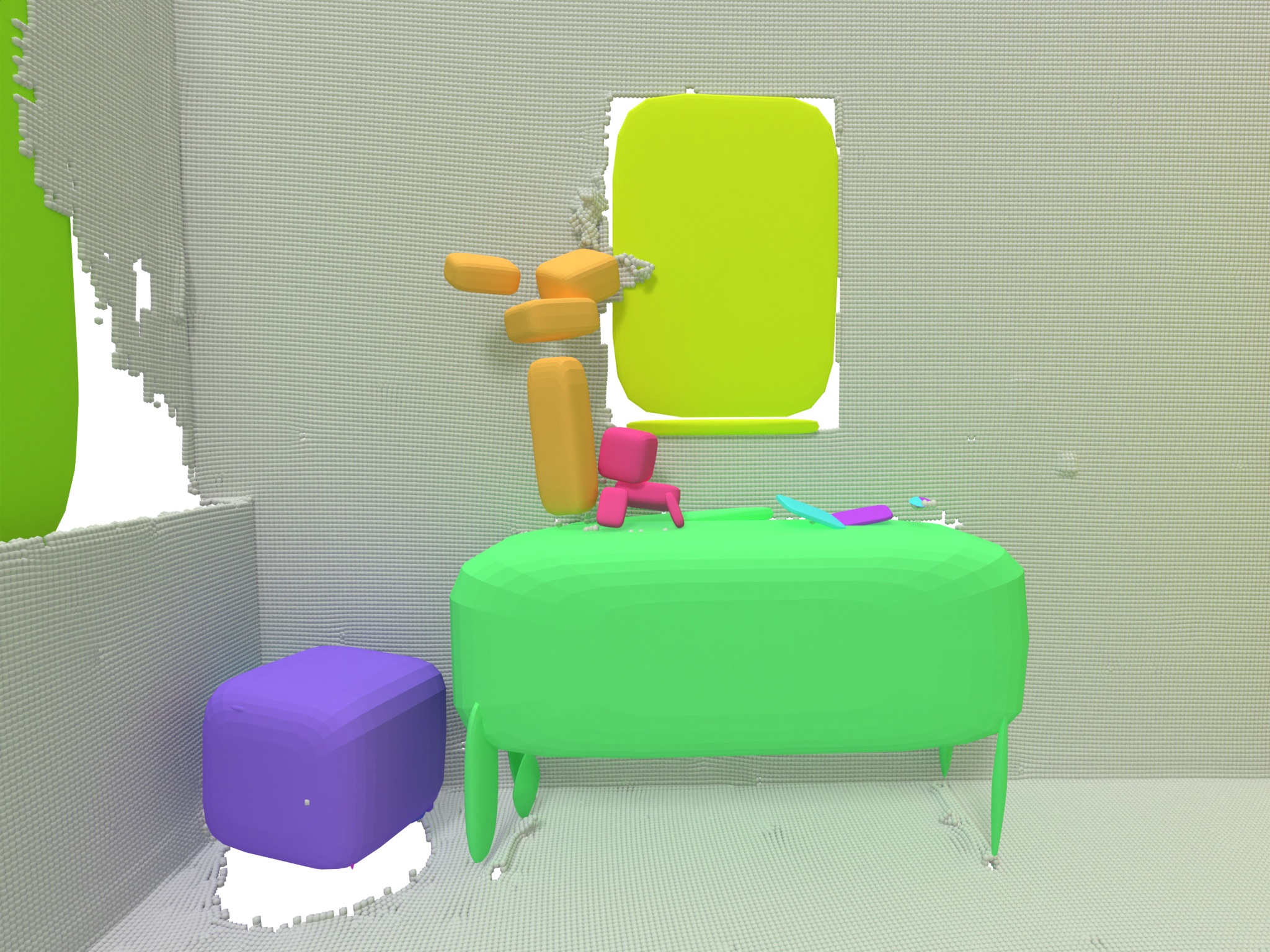}
\vspace{5px}\\

\end{small}
\caption{\textbf{Qualitative Results on Replica~\cite{Straub2019TheRD}.}
We show results on different scenes from Replica. In the first row we show a rendering of the input mesh, while on the second row we show the output superquadrics. We visualize different object instances with different colors.}
\label{fig:quali-replica}
\end{figure*}

In Fig.~\ref{fig:quali-replica} we visualize the superquadric decompositions for some scenes from Replica~\cite{Straub2019TheRD}. The instance masks are predicted using Mask3D~\cite{Schult2022Mask3DMT}. The points not belonging to any instance are visualized in gray.

\subsection{ScanNet++}
\begin{figure*}
\centering
\begin{small}
\begin{tabular}{ccccc}
\hspace{0.01\linewidth} &\hspace{0.215\linewidth} & \hspace{0.215\linewidth} & \hspace{0.215\linewidth} & \hspace{0.215\linewidth} \\
&\textbf{88f265fe25} &\textbf{95748dd597} &\textbf{2a1b555966} &\textbf{0b031f3119}
\end{tabular}\\
\vspace{1px}
\rotatebox{90}{\hspace{20px}\textbf{Point Cloud}}
\hspace{2px}
\includegraphics[width=0.238\linewidth,,clip]
{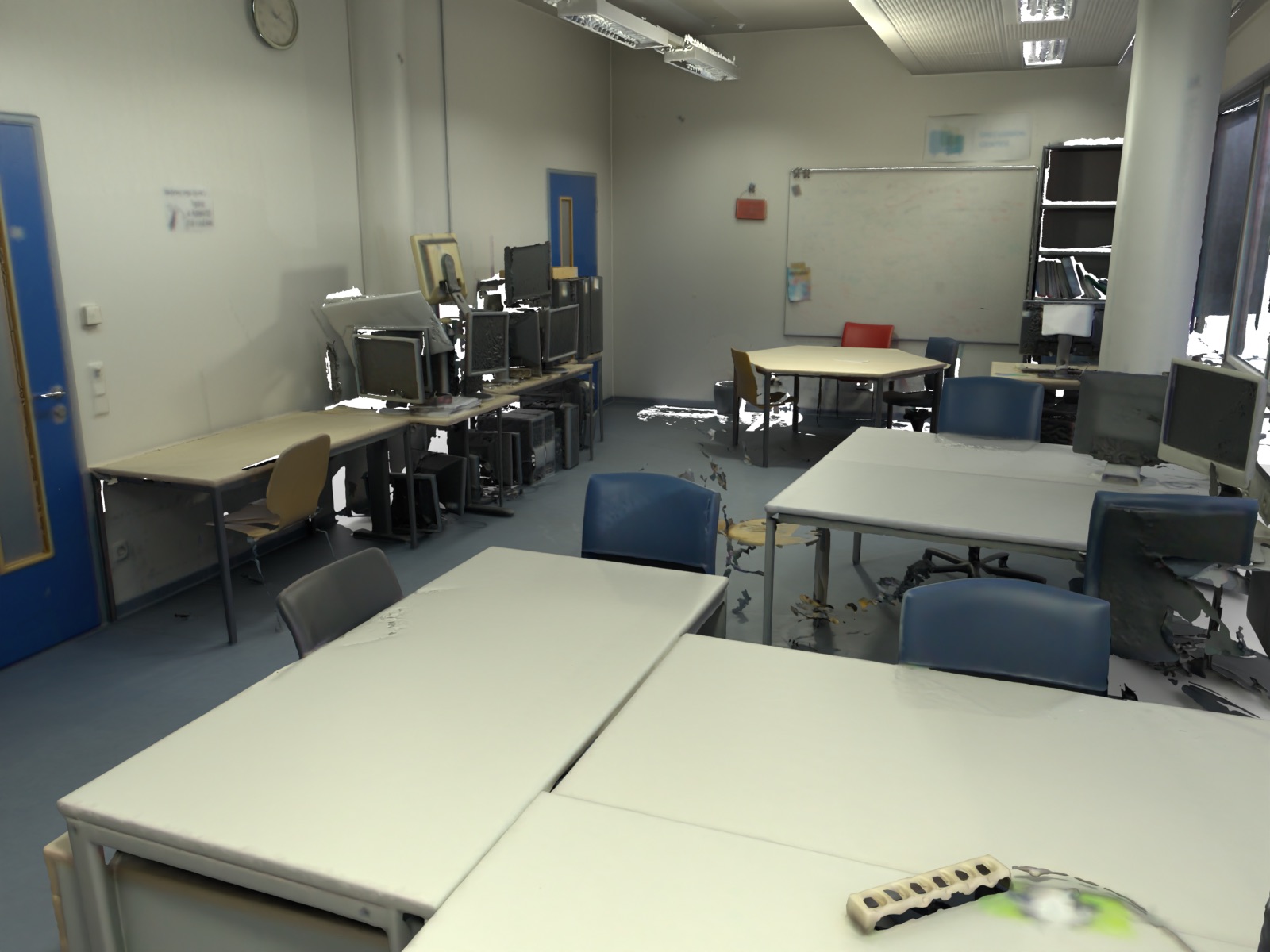}
\includegraphics[width=0.238\linewidth,clip]{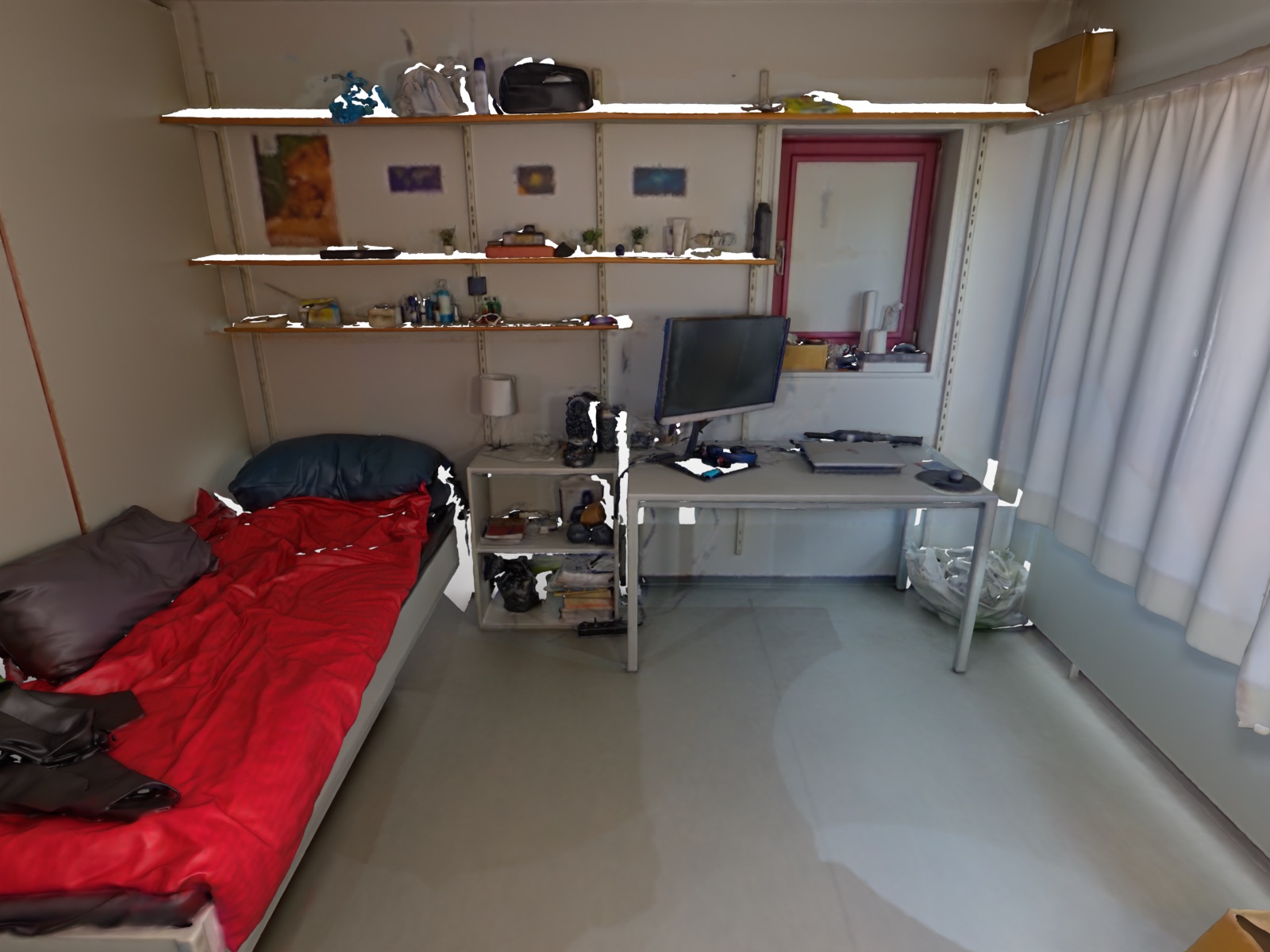}
\includegraphics[width=0.238\linewidth,,clip]
{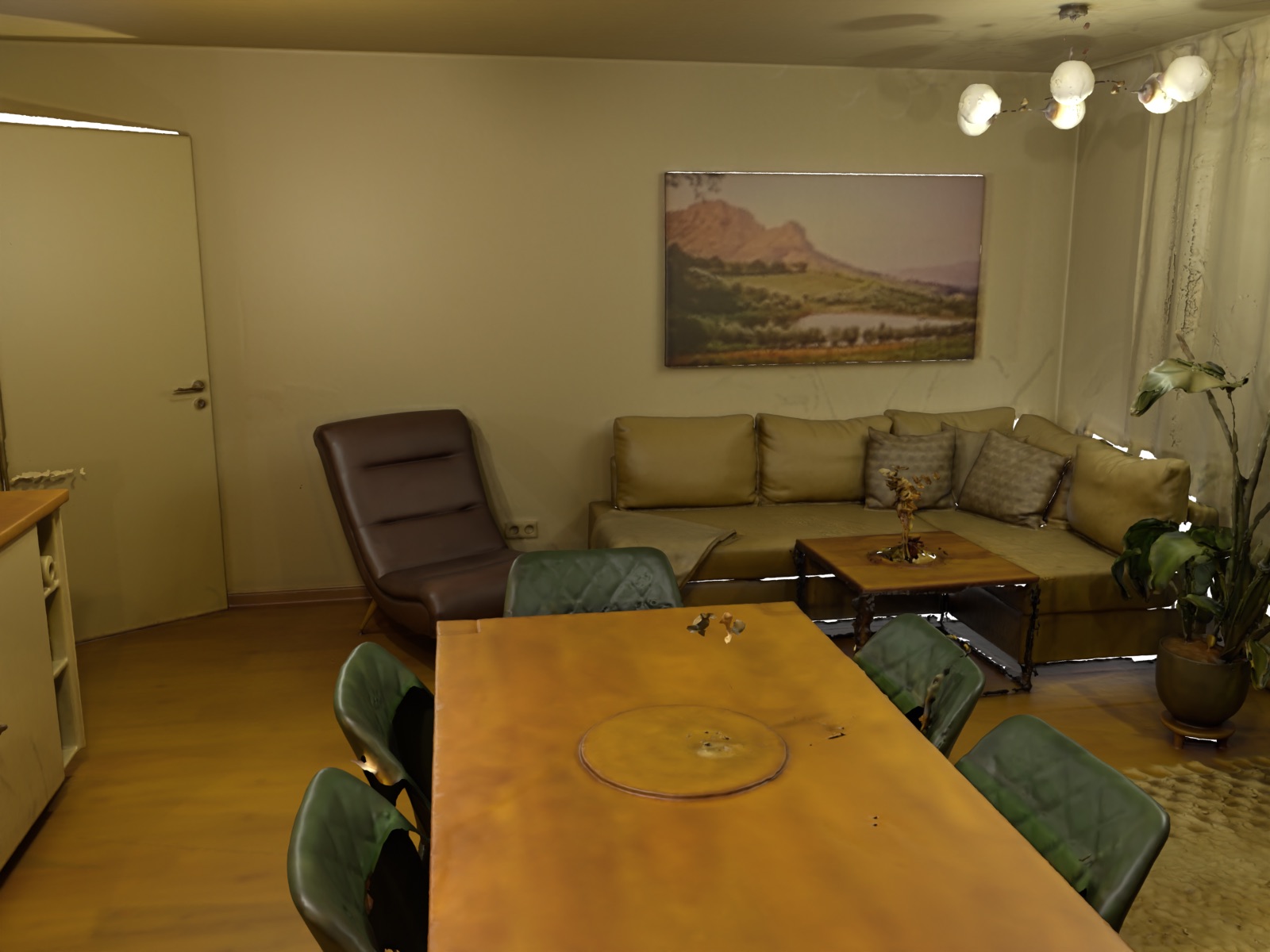}
\includegraphics[width=0.238\linewidth,clip]{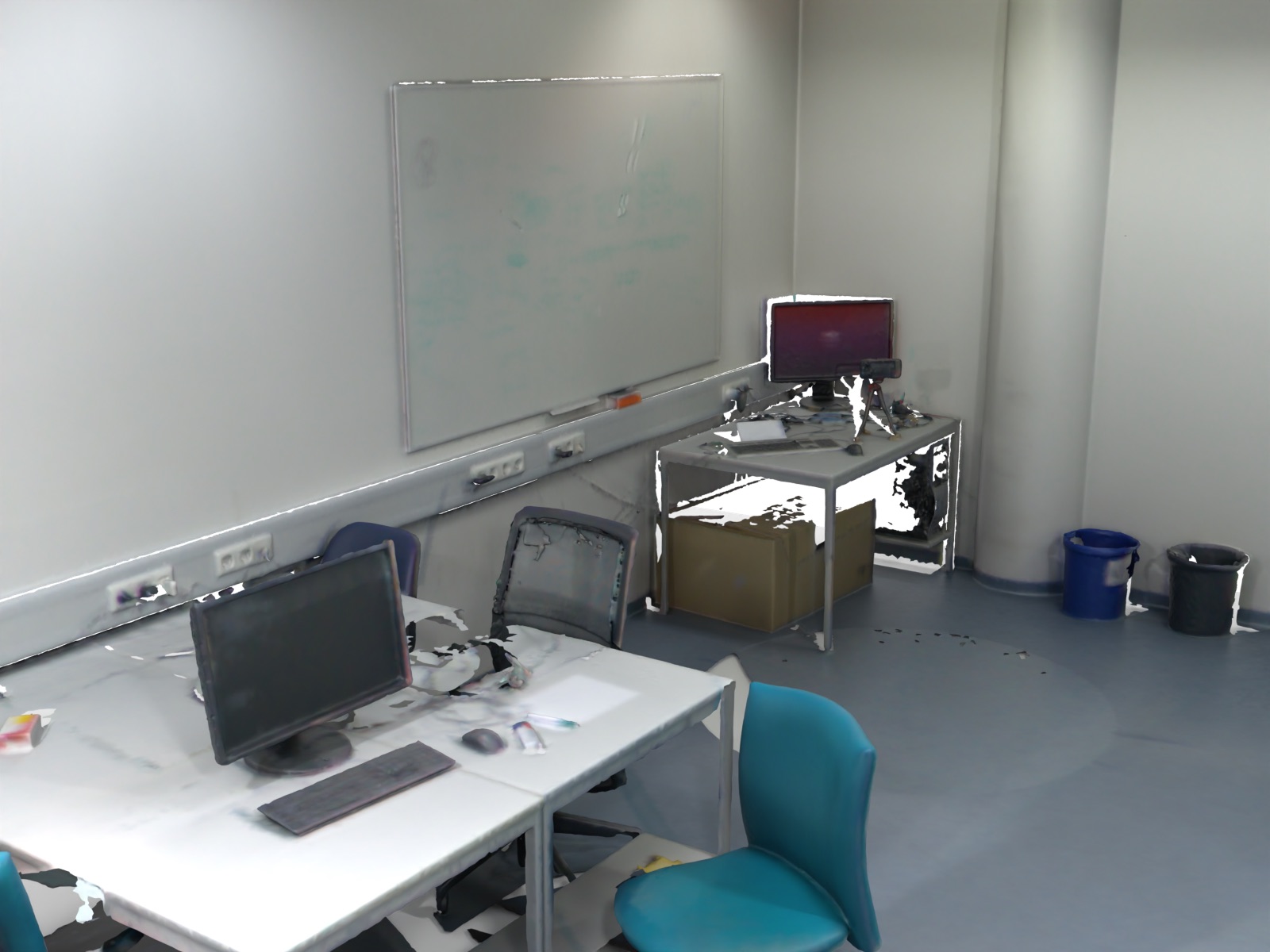}
\vspace{2px}\\

\rotatebox{90}{\hspace{16px}\textbf{Superquadrics}}
\hspace{1.3px}
\includegraphics[width=0.238\linewidth,,clip]{figures/scannetpp/88f265fe25_sq.jpeg}
\includegraphics[width=0.238\linewidth,clip]{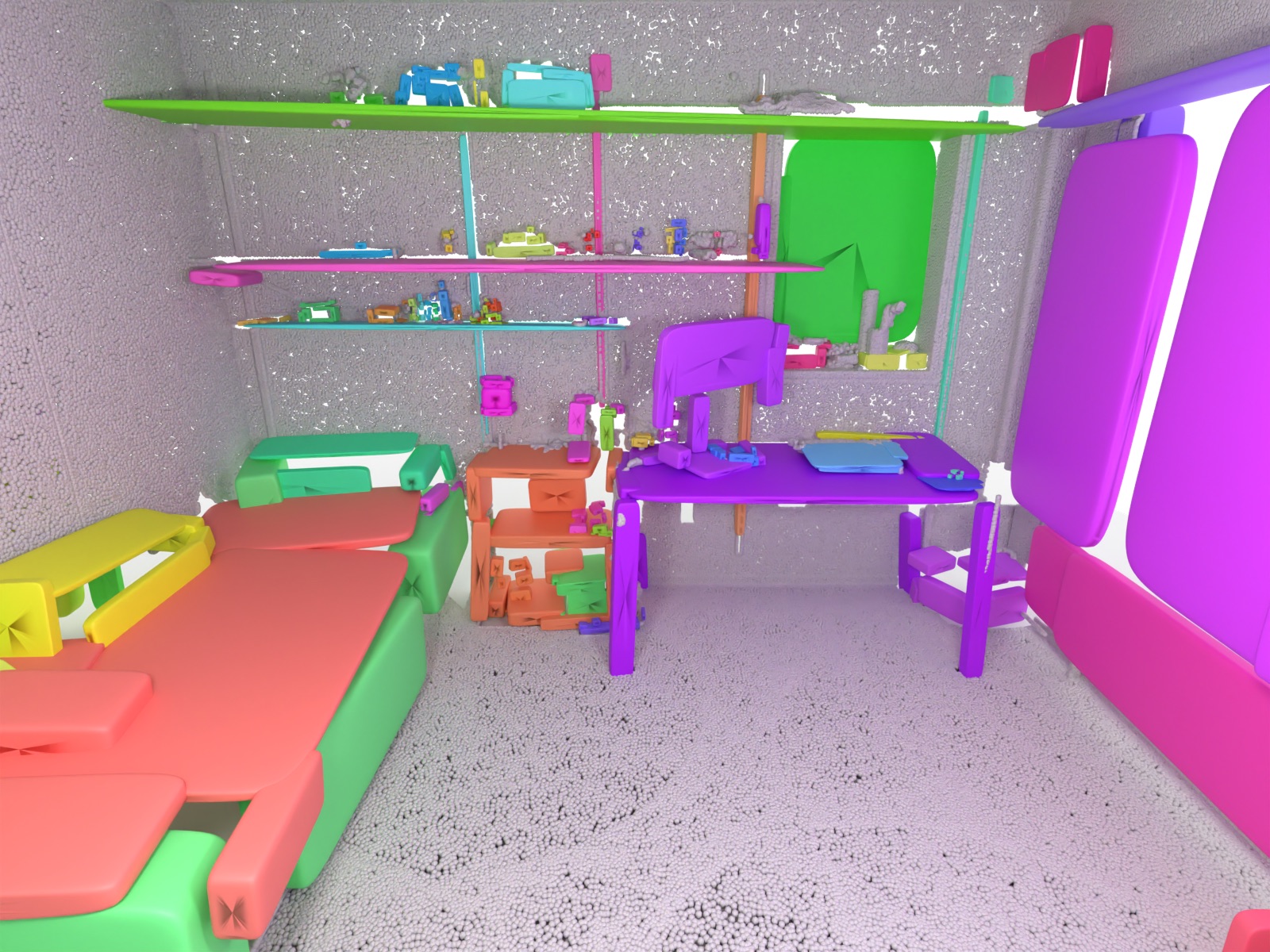}
\includegraphics[width=0.238\linewidth,,clip]{figures/scannetpp/2a1b555966_sq_bis.jpg}
\includegraphics[width=0.238\linewidth,clip]{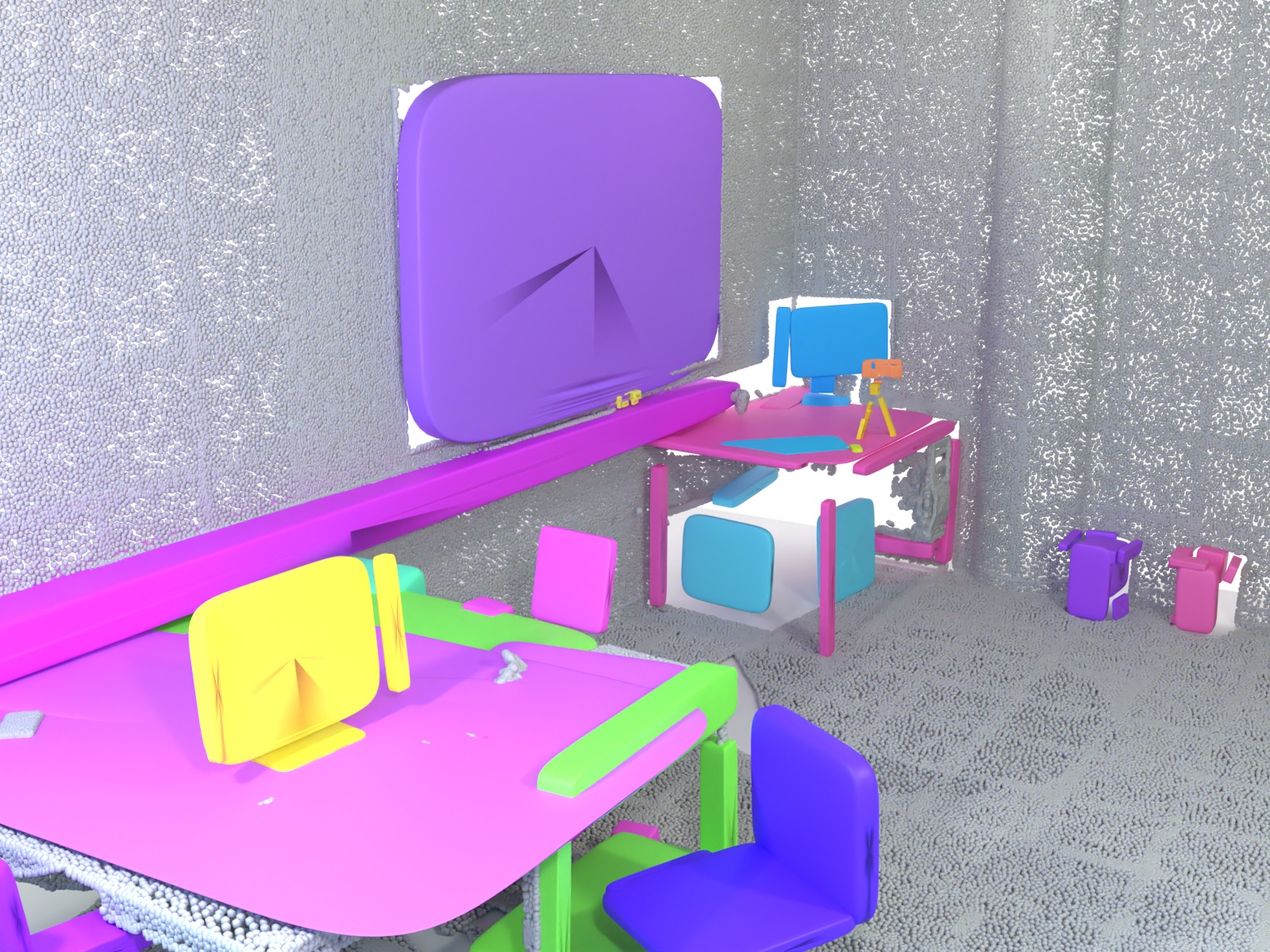}
\vspace{5px}\\

\end{small}
\caption{\textbf{Qualitative Results on ScanNet++~\cite{Yeshwanth2023ScanNetAH}.}
We show results on different scenes from ScanNet++. In the first row we show a rendering of the input mesh, while on the second row we show the output superquadrics. We visualize different object instances with different colors.}
\label{fig:quali-scannetpp}
\end{figure*}

In Fig.~\ref{fig:quali-scannetpp} we visualize the superquadric decompositions for some scenes from ScanNet++~\cite{Yeshwanth2023ScanNetAH}. The instance masks the ground truth instance masks. The points not belonging to any instance are visualized in gray.

\section{Robot Experiment}
In this section we introduce the key methods and parameters used in our robot experiments. We also present more detailed qualitative and quantitative evaluation results.
\subsection{Setup}
For \textit{path planning} in both ScanNet++~\cite{Yeshwanth2023ScanNetAH} and real-world scenarios, we use the Python binding of the Open Motion Planning Library (OMPL). The state space is defined as a \textit{3D RealVectorStateSpace}, with boundaries extracted from the 3D bounding box of the input point cloud. We employ a sampling-based planner (RRT*), setting a maximum planning time of 2 seconds per start-goal pair.

\noindent In ScanNet++ scenes, the occupancy grid and voxel grid are both set to a 10 cm resolution, with voxels generated from the original point cloud. The collision radius is 25 cm. For dense occupancy grid planning, we enforce an additional constraint in the validity checking to ensure that paths remain within 25 cm of free space, preventing them from extending outside the scene or penetrating walls. And the planned occupancy grid path serves as a reference for computing relative path optimality in our evaluation. Start and goal points are sampled within a 0.4m-0.6m height range in free space, as most furniture and objects are within this range. This allows for a fair evaluation of how different representations capture collisions for valid path planning. During evaluation, we further validate paths by interpolating them into 5 cm waypoint intervals. Each waypoint is checked against the occupancy grid to ensure that its nearest occupied grid is beyond 25 cm and its nearest free grid is within 25 cm. A path is considered unsuccessful if more than 10\% of waypoints fail this check. This soft constraint accounts for the sampling-based nature of RRT*, which does not enforce voxel-level validity but instead checks waypoints along the tree structure, leading to occasional minor violations. In the real-world path planning, we set the collision radius to 60 cm to approximate the size of the Boston Dynamics Spot robot. Spot follows the planned path using its Python API for execution.

\noindent For \textit{grasping} in real-world experiments, we use the \href{https://github.com/robotology/superquadric-lib/tree/master}{superquadric-library} to compute single-hand grasping poses based on superquadric parameters. The process begins by identifying the object of interest and its corresponding superquadric decomposition. One of the superquadrics is selected and fed into the grasping estimator. To execute the grasp, the robot first navigates to the object's location during the planning stage. Then, using its built-in inverse kinematics planner and controller, the robot moves its end-effector to the estimated grasping pose for object manipulation.

\subsection{Planning Results}
In Tab.~\ref{tab:exp_robot} we report the complete planning results on 15 Scannet++~\cite{Yeshwanth2023ScanNetAH} scenes.
\begin{table*}[t]
\centering
\resizebox{0.99\textwidth}{!}{
\begin{tabular}{lcccccccccccccccccccc}
\toprule
\multirow{2}{*}{\textbf{Method}} & \multicolumn{4}{c}{\textbf{0a76e06478}} & \multicolumn{4}{c}{\textbf{0c6c7145ba}} & \multicolumn{4}{c}{\textbf{0f0191b10b}} & \multicolumn{4}{c}{\textbf{1a8e0d78c0}} & \multicolumn{4}{c}{\textbf{1a130d092a}} \\
\cmidrule(lr){2-5} \cmidrule(lr){6-9} \cmidrule(lr){10-13} \cmidrule(lr){14-17} \cmidrule(lr){18-21}
& Time(ms) & Suc.(\%) & Opt. & Mem.  & Time(ms) & Suc.(\%) & Opt. & Mem.  & Time(ms) & Suc.(\%) & Opt. & Mem.  & Time(ms) & Suc.(\%) & Opt. & Mem.  & Time(ms) & Suc.(\%) & Opt. & Mem.  \\

\midrule
Occupancy  & 0.05 & 100 & 1.00 & 960KB &  0.06  & 100 & 1.00 & 667KB & 0.06 & 100 & 1.00 & 1031KB & 0.05 & 100 & 1.00 & 926KB & 0.05 & 100 & 1.00 & 803KB \\
PointCloud      & 0.07 & 86 & 0.98 & 18MB  & 0.09 & 91 & 0.99 & 12MB & 0.03 & 77 & 0.99 & 19MB & 0.05 & 91 & 0.99 & 18MB  & 0.05 & 89 & 0.98 & 18MB  \\
Voxels          & 0.03 & 100 & 0.97 & 91KB & 0.03 & 100 & 1.00 & 65KB & 0.03 & 100 & 0.99 & 99KB & 0.03 & 100 & 1.01 & 91KB & 0.03 & 100 & 1.09 & 99KB \\
Cuboids~\cite{ramamonjisoa2022monteboxfinder}     & 0.11 & 32 & 0.98 & 22KB & 0.10 & 18 & 1.02 & 19KB & 0.14 & 85 & 1.03 & 34KB & 0.10 & 50 & 1.06 & 21KB & 0.12 & 79 & 1.00 & 27KB \\
\name{}          & 0.17 & 100 & 0.99 & 52KB   & 0.16 & 100 & 0.97 & 48KB & 0.17 & 92 & 0.94 & 51KB & 0.14 & 91 & 0.99 & 39KB  & 0.13 & 100 & 0.98 & 35KB\\
\midrule

\multirow{2}{*}{\textbf{Method}} & \multicolumn{4}{c}{\textbf{0a76e06478}} & \multicolumn{4}{c}{\textbf{0b031f3119}} & \multicolumn{4}{c}{\textbf{0dce89ab21}} & \multicolumn{4}{c}{\textbf{0e350246d4}} & \multicolumn{4}{c}{\textbf{0eba3981c9}} \\
\cmidrule(lr){2-5} \cmidrule(lr){6-9} \cmidrule(lr){10-13} \cmidrule(lr){14-17} \cmidrule(lr){18-21}
& Time(ms) & Suc.(\%) & Opt. & Mem.  & Time(ms) & Suc.(\%) & Opt. & Mem.  & Time(ms) & Suc.(\%) & Opt. & Mem.  & Time(ms) & Suc.(\%) & Opt. & Mem.  & Time(ms) & Suc.(\%) & Opt. & Mem.  \\

\midrule
Occupancy  & 0.05 & 100 & 1.00 & 916KB &  0.06  & 100 & 1.00 & 1760KB & 0.05 & 100 & 1.00 & 1070KB & 0.06 & 100 & 1.00 & 366KB & 0.06 & 100 & 1.00 & 473KB \\
PointCloud      & 0.05 & 86 & 1.02 & 18MB  & 0.06 & 96 & 1.04 & 25MB & 0.05 & 84 & 1.13 & 19MB & 0.06 & 88 & 1.22 & 10MB  & 0.14 & 80 & 0.98 & 45MB  \\
Voxels          & 0.03 & 100 & 1.01 & 99KB & 0.03 & 100 & 1.00 & 160KB & 0.03 & 100 & 1.19 & 104KB & 0.03 & 100 & 1.00 & 51KB & 0.03 & 100 & 1.12 & 199KB \\
Cuboid\cite{ramamonjisoa2022monteboxfinder}     & 0.14 & 71 & 1.12 & 32KB & 0.11 & 78 & 1.03 & 24KB & 0.11 & 35 & 1.00 & 23KB & 0.09 & 62 & 1.00 & 15KB & 0.17 & 87 & 1.17 & 41KB \\
SuperDec          & 0.16 & 86 & 1.17 & 46KB   & 0.16 & 93 & 0.98 & 46KB & 0.13 & 100 & 1.07 & 33KB & 0.15 & 88 & 1.22 & 40KB  & 0.19 & 57 & 1.10 & 58KB\\

\midrule

\multirow{2}{*}{\textbf{Method}} & \multicolumn{4}{c}{\textbf{7cd2ac43b4}} & \multicolumn{4}{c}{\textbf{1841a0b525}} & \multicolumn{4}{c}{\textbf{25927bb04c}} & \multicolumn{4}{c}{\textbf{e0abd740ba}} & \multicolumn{4}{c}{\textbf{0f25f24a4f}} \\
\cmidrule(lr){2-5} \cmidrule(lr){6-9} \cmidrule(lr){10-13} \cmidrule(lr){14-17} \cmidrule(lr){18-21}
& Time(ms) & Suc.(\%) & Opt. & Mem.  & Time(ms) & Suc.(\%) & Opt. & Mem.  & Time(ms) & Suc.(\%) & Opt. & Mem.  & Time(ms) & Suc.(\%) & Opt. & Mem.  & Time(ms) & Suc.(\%) & Opt. & Mem.  \\

\midrule
Occupancy  & 0.06 & 100 & 1.00 & 1241KB &  0.05  & 100 & 1.00 & 1053KB & 0.06 & 100 & 1.00 & 407KB & 0.06 & 100 & 1.00 & 554KB & 0.05 & 100 & 1 & 7MB \\
PointCloud      & 0.05 & 100 & 1.09 & 25MB  & 0.04 & 89 & 0.98 & 16MB & 0.06 & 100 & 1.01 & 11MB & 0.06 & 97 & 0.93 & 16MB & 0.07 & 61 & 0.97 & 99MB  \\
Voxels          & 0.03 & 100 & 1.00 & 137KB & 0.03 & 100 & 0.98 & 82KB & 0.03 & 83 & 1.04 & 51KB & 0.03 & 100 & 1.04 & 83KB & 0.03 & 96 & 0.96 & 617KB \\
Cuboid\cite{ramamonjisoa2022monteboxfinder}     & 0.21 & 80 & 1.04 & 57KB & x & x & x & 15KB & 0.09 & 87 & 0.96 & 17KB & 0.07 & 52 & 1.04 & 11KB & x & x & x & x \\
SuperDec          & 0.15 & 100 & 1.05 & 45KB   & 0.10 & 94 & 0.87 & 18KB & 0.17 & 83 & 1.30 & 53KB & 0.12 & 100 & 0.87 & 27KB  & 0.21 & 57 & 0.82 & 71KB\\

\bottomrule
\end{tabular}
}
\caption{\textbf{Path Planning Results. } We show results of path planning for different ScanNet++~\cite{Yeshwanth2023ScanNetAH} scenes, whose ids are reported on the top. PointCloud method uses dense point clouds from ScanNet++, all other methods process the same input point cloud. Time refers to average execution time of the validity-check function during the sampling stage of planning. Success rate (Suc.) is calculated after excluding trials where no representation could generate valid path due to randomness of start and goal sampling. The Cuboid method encounters an out-of-memory failure when fitting scene \textit{0f25f24a4f} due to its large scale, and fails to find any valid path in scene \textit{1841a0b525}.}

\label{tab:exp_robot}
\end{table*}

\end{document}